\documentclass[12pt]{article}
\usepackage{amsmath}
\usepackage{graphicx,subcaption,mathtools}
\usepackage[round]{natbib}  
\usepackage[colorlinks,allcolors=blue]{hyperref}
\usepackage{amsmath,amsfonts,amssymb,amsthm,booktabs,color,epsfig,graphicx,hyperref,url,setspace}

\usepackage{import,comment,xr,apptools,titletoc}
\usepackage{stackengine}

\usepackage{bbm,bm}
\usepackage[T1]{fontenc} % For accents in words
\usepackage{enumitem}

\theoremstyle{plain}
\newtheorem{theorem}{Theorem}
\newtheorem{proposition}[theorem]{Proposition}
\newtheorem{lemma}[theorem]{Lemma}

\theoremstyle{definition}

\theoremstyle{remark}
\newtheorem{remark}{Remark}
\newtheorem{example}[theorem]{Example}

\DeclareMathOperator*{\argmin}{arg\,min}

\newcommand{\norm}[1]{\left|\left| #1 \right|\right|}

\newcommand{\Haus}{\mathrm{Haus}}
\newcommand{\Exp}{\mathrm{Exp}}
\newcommand{\Diag}{\mathrm{Diag}}

\newcommand{\grad}{\mathrm{grad}\,}
\renewcommand{\hat}{\widehat}
\renewcommand{\tilde}{\widetilde}

\theoremstyle{plain}
\newtheorem{innercustomthm}{Theorem}
\newenvironment{customthm}[1]
{\renewcommand\theinnercustomthm{#1}\innercustomthm}
{\endinnercustomthm}

\newtheorem{innercustomprop}{Proposition}
\newenvironment{customprop}[1]
{\renewcommand\theinnercustomprop{#1}\innercustomprop}
{\endinnercustomprop}

\AtAppendix{\counterwithin{theorem}{section}}
\AtAppendix{\counterwithin{remark}{section}}
\AtAppendix{\counterwithin{figure}{section}}
\AtAppendix{\counterwithin{equation}{section}}

\allowdisplaybreaks

\newcommand{\blind}{0}

\begin{document}

\def\spacingset#1{\renewcommand{\baselinestretch}%
{#1}\small\normalsize} \spacingset{1}

%%%%%%%%%%%%%%%%%%%%%%%%%%%%%%%%%%%%%%%%%%%%%%%%%%%%%%%%%%%%%%%%%%%%%%%%%%%%%%

\if0\blind
{
  \title{\bf Mode and Ridge Estimation in Euclidean and Directional Product Spaces: A Mean Shift Approach}
  \author{Yikun Zhang\thanks{
  Email: yikun@uw.edu. YZ is supported in part by YC's NSF grants DMS-2141808 and DMS-2310578.}\hspace{.2cm}\\
    Department of Statistics, University of Washington\\
    and \\
    Yen-Chi Chen\thanks{Email: yenchic@uw.edu. YC is supported by NSF grants DMS-2112907, DMS-2141808, DMS-2310578, and NIH U24-AG072122.} \\
    Department of Statistics, University of Washington}
  \maketitle
} \fi

\if1\blind
{
  \bigskip
  \bigskip
  \bigskip
  \begin{center}
    {\LARGE\bf Mode and Ridge Estimation in Euclidean and Directional Product Spaces: A Mean Shift Approach}
\end{center}
  \medskip
} \fi

\bigskip
\begin{abstract}
The set of local modes and density ridge lines are important summary characteristics of the data-generating distribution. In this work, we focus on estimating local modes and density ridges from point cloud data in a product space combining two or more Euclidean and/or directional metric spaces. 
Specifically, our approach extends the (subspace constrained) mean shift algorithm to such product spaces, addressing potential challenges in the generalization process. We establish the algorithmic convergence of the proposed methods, along with practical implementation guidelines. Experiments on simulated and real-world datasets demonstrate the effectiveness of our proposed methods. 
% Supplementary materials for this article are available online.
\end{abstract}

\noindent%
{\it Keywords:} Mean shift, Directional data, Mode clustering, Ridge estimation, Optimization on Product Manifolds.
\vfill

\newpage
\spacingset{1.75} % DON'T change the spacing!

\section{Introduction}

The rapid growth of machine learning applications has been accompanied by an increasing prevalence of large-scale data with complex structures. Extracting meaningful low-dimensional summaries from such data can provide critical insights into the underlying scientific problems \citep{izenman2012introduction,chazal2021introduction}. 
While many methods exist for characterizing low-dimensional structures \citep{fefferman2016testing}, this work focuses on estimating local modes and density ridges (also known as principal curves/surfaces in \citealt{Principal_curve_surf2011}) from point cloud data because of two reasons. Statistically, local modes and density ridges effectively capture the intrinsic low-dimensional structure of data and can be consistently estimated using kernel density estimator (KDE) \citep{romano1988weak,mokkadem2003law,Non_ridge_est2014}. Practically, the mean shift algorithm \citep{MS1975,MS1995,MS2002} and its subspace constrained variant \citep{Principal_curve_surf2011,SCMS_conv2013} are well-suited for identifying these features in Euclidean spaces.

However, some practical data of interest do not reside on a flat Euclidean space but a nonlinear $q$-dimensional sphere (or any topological space that is homeomorphic to a sphere). For instance, dihedral angles in protein structures are periodic and naturally represented on a unit circle ($q=1$) \citep{zimmermann2006support}. Moreover, data may consist of mixture components defined on metric spaces with different geometries. A concrete example arises in astronomy, where each observed object is recorded as a three-element tuple: (right ascension, declination, redshift) \citep{dawson2016sdss,brown2018gaia}. The first two elements encode its position on the celestial sphere ($q=2$), while the redshift value reflects its linear distance from Earth. In a broader sense, such data are known as spatio-temporal data, combining spatial components (Euclidean or directional) with the temporal elements that may be continual or periodic \citep{hall2006real}. For these complicated data structures, standard KDE and mean shift algorithm are no longer applicable.

\begin{figure}[t]
	\vspace{-20mm}
	\captionsetup[subfigure]{justification=centering}
	\centering
	\begin{subfigure}[t]{.48\linewidth}
		\centering
		\includegraphics[width=1\linewidth]{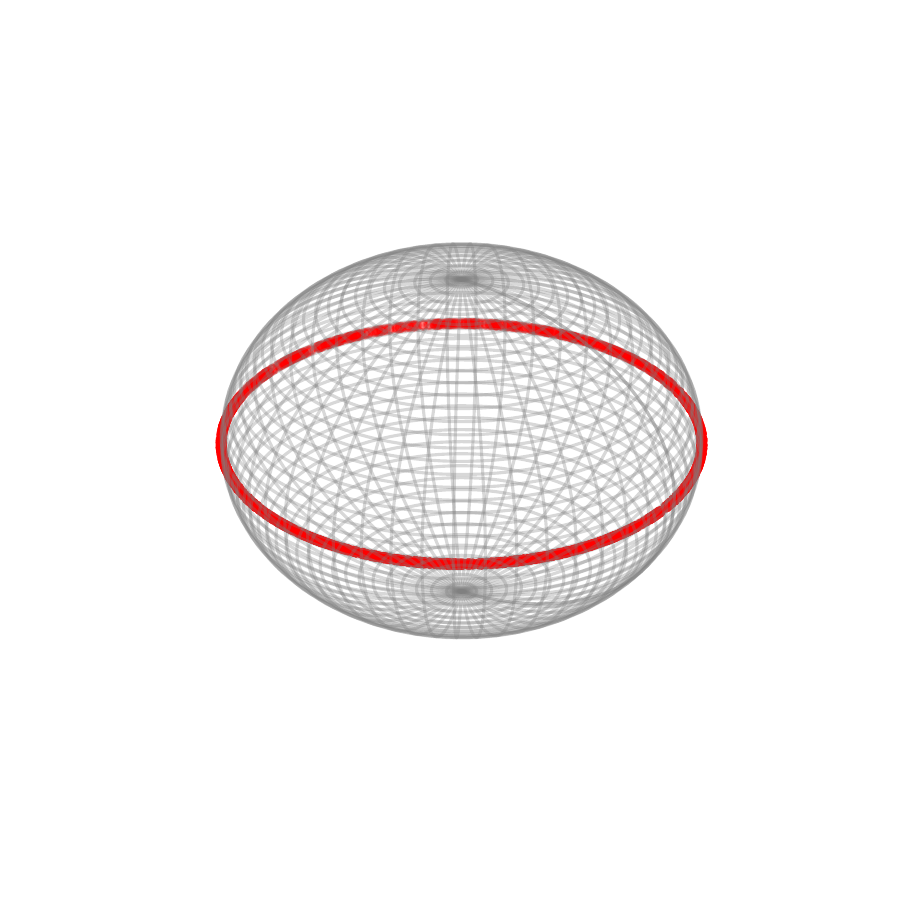}
	\end{subfigure}
	\hfil
	\begin{subfigure}[t]{.49\linewidth}
		\centering
		\includegraphics[width=1\linewidth]{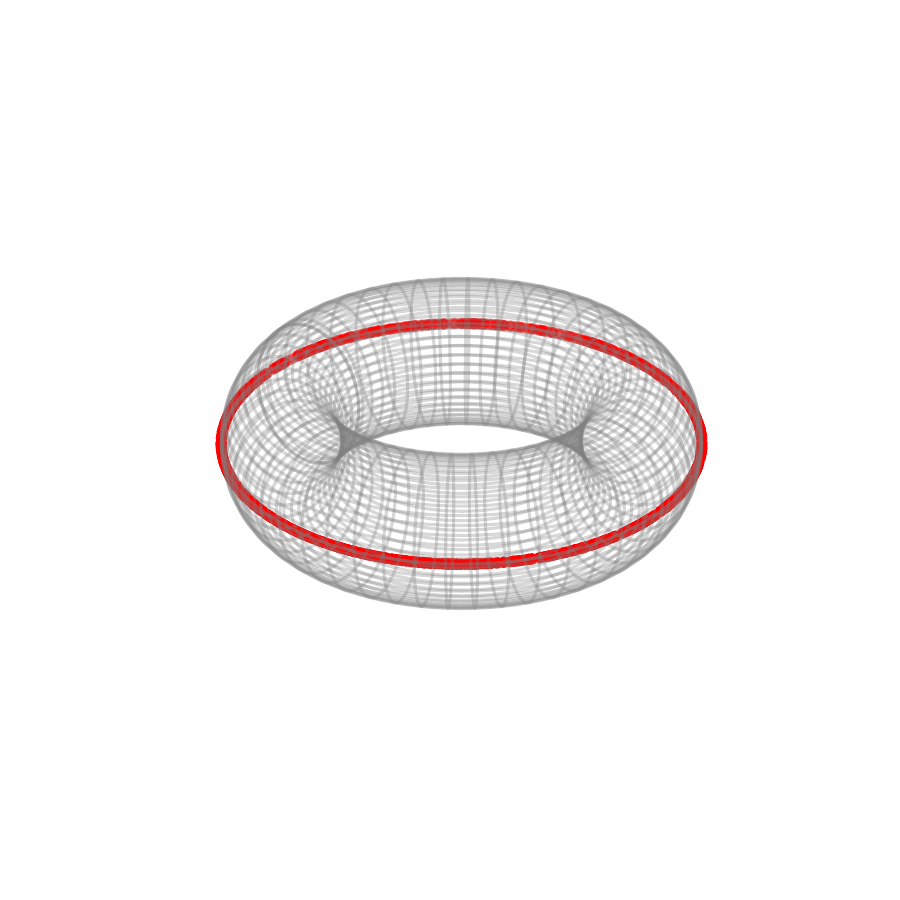}
	\end{subfigure}
	\vspace{-18mm}
	\caption{Simulated dataset $\{(\theta_i,\phi_i)\}_{i=1}^{1000}$ on $\Omega_2$ and $\Omega_1\times \Omega_1$. 
		While the two coordinates $(\theta_i,\phi_i)$ of the data in both panels are periodic, their supports have different topology, so the corresponding mode-seeking and ridge-finding algorithms will not be the same.}
	%Each observation $(\theta_i,\phi_i)$ is sampled uniformly from $\left[2p_1\pi, 2(p_1+1)\pi\right)\times \{2p_2\pi\}$ for some integers $p_1,p_2$.}
\label{fig:space_comp}
\end{figure}

In this paper, we propose methods for estimating local modes and density ridges of a data-generating density function supported on a Cartesian product space $\mathcal{S}_1 \times \mathcal{S}_2$, using KDE and (subspace constrained) mean shift algorithms. Here, $\mathcal{S}_1,\mathcal{S}_2$ represent either Euclidean/linear spaces $\mathbb{R}^D$, or directional spaces 
$\Omega_q=\left\{\bm{x}\in \mathbb{R}^{q+1}:\norm{\bm{x}}_2=1 \right\}$, where $\norm{\cdot}_2$ is the usual Euclidean norm. 
% One example is to find the largest singular value of a matrix $A\in \mathbb{R}^{(q_1+1)\times (q_2+1)}$, which is the maximum of $f(\bm{x},\bm{y})=\bm{x}^{\top}A \bm{y}$ on $\Omega_{q_1}\times \Omega_{q_2}$; see Section 2.2 in \cite{boumal2023intromanifolds}.
Although we focus on product spaces with two (topological) factors, our methodology can be readily extended to product spaces with finitely many factors. Furthermore, either or both factors can be topological spaces homeomorphic to $\Omega_q$, provided that the homeomorphism is known.
While mode-seeking and ridge-finding problems have been studied independently in $\mathbb{R}^D$ and $\Omega_q$ \citep{chacon2020modal,Non_ridge_est2014,DirSCMS2021}, the generalizations to $\mathcal{S}_1\times \mathcal{S}_2$ introduce new challenges. First, the topology of $\mathcal{S}_1\times \mathcal{S}_2$  differs fundamentally from that of $\mathbb{R}^D$ or $\Omega_q$ when one or both factors $\mathcal{S}_j, j=1,2$ are directional. Consider a random sample $\{(\theta_i,\phi_i)\}_{i=1}^{1000}$ with $\theta_i\in \left[0, 2\pi\right)$ and $\phi_i=0$. Depending on the problem context, this dataset may be viewed as either points from a circle on the sphere $\Omega_2$ (with $(\theta_i,\phi_i)$ as longitude and latitude) or on the torus $\Omega_1\times \Omega_1$. The periodicity constraints of $\Omega_2$ and $\Omega_1\times \Omega_1$ are different, and these two spaces are not homeomorphic; see \autoref{fig:space_comp}. Second, directly applying existing mode-seeking or ridge-finding methods to $\mathcal{S}_1,\mathcal{S}_2$ independently leads to an \emph{unidentifiability} issue on $\mathcal{S}_1\times\mathcal{S}_2$. For example, consider two densities $f_1,f_2$, on $\Omega_1\times \Omega_1$ with the sets of local modes as $\mathcal{M}=\left\{(0,0),(0,\frac{3\pi}{4}), (\frac{\pi}{2},0), (\frac{\pi}{2},\frac{3\pi}{4}) \right\}$ and $\mathcal{M}'=\left\{(0,\frac{3\pi}{4}), (\frac{\pi}{2},0) \right\}$, respectively; see also \autoref{subsec:sim2}. While $\mathcal{M}$ and $\mathcal{M}'$ differ, the marginal local modes, which are $0,\frac{\pi}{2}$ on the first coordinate and $0,\frac{3\pi}{4}$ on the second coordinate, are identical. Applying existing mode-seeking methods coordinatewise would fail to distinguish $f_1$ from $f_2$. To resolve this unidentifiability issue, we propose mean shift-based approaches that identify modes and ridges jointly on $\mathcal{S}_1\times\mathcal{S}_2$.

\begin{figure*}[t]
	\captionsetup[subfigure]{justification=centering}
	\centering
	\begin{subfigure}[t]{.24\linewidth}
		\centering
		\includegraphics[width=1\linewidth]{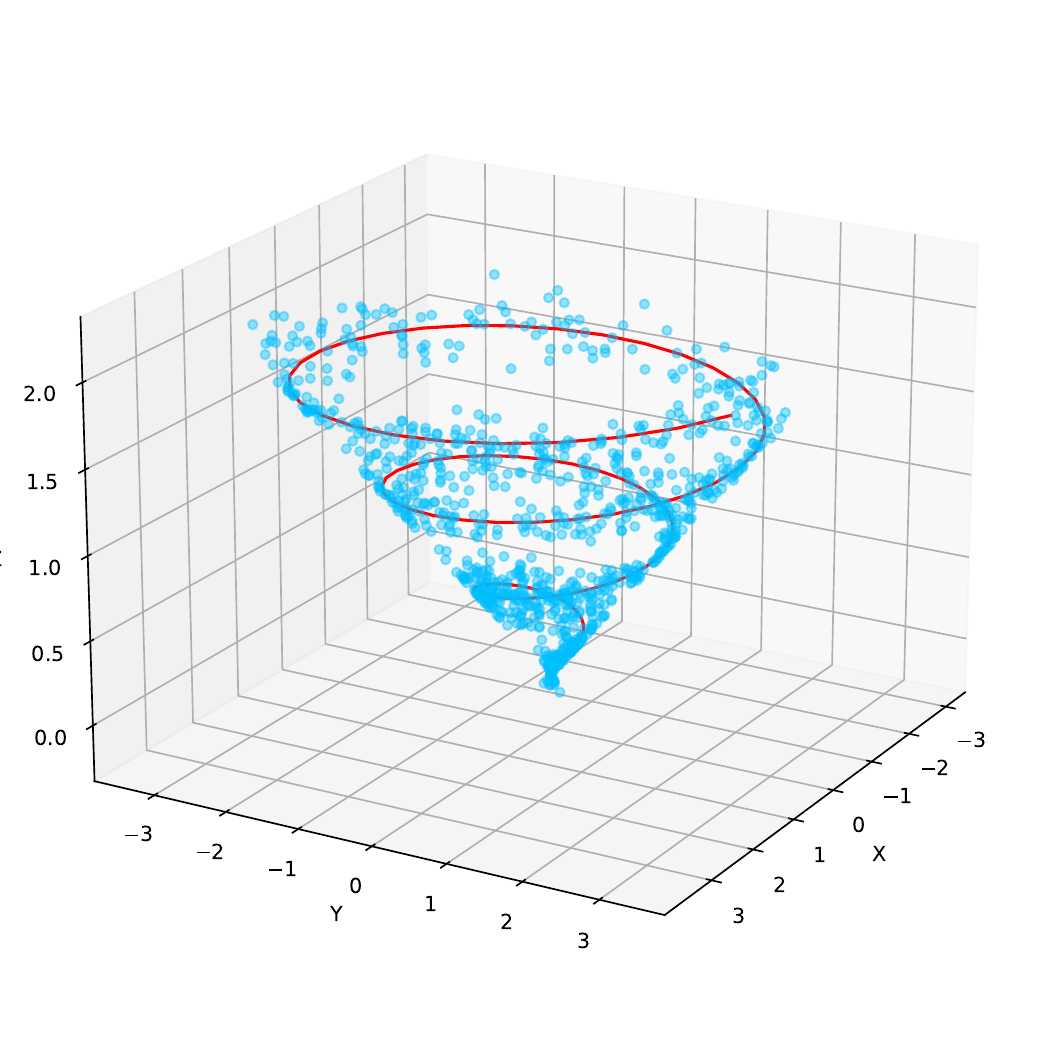}
		\caption{Simulated noisy observations around a spiral curve on $\Omega_2 \times \mathbb{R}$.}
	\end{subfigure}
	\hfil
	\begin{subfigure}[t]{.24\linewidth}
		\centering
		\includegraphics[width=1\linewidth]{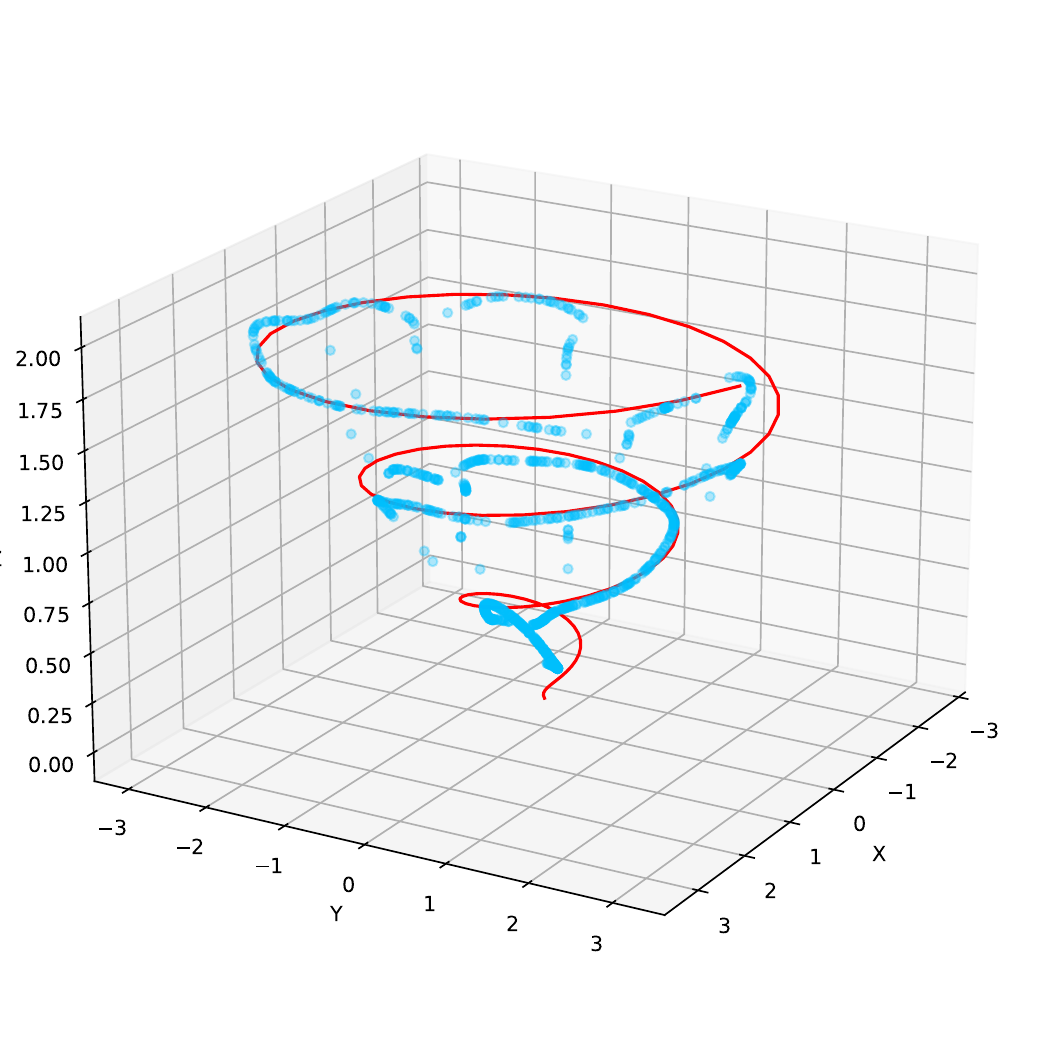}
		\caption{Regular SCMS on Cartesian coordinates in $\mathbb{R}^3$.}
	\end{subfigure}
	\hfil
	\begin{subfigure}[t]{.24\linewidth}
		\centering
		\includegraphics[width=1\linewidth]{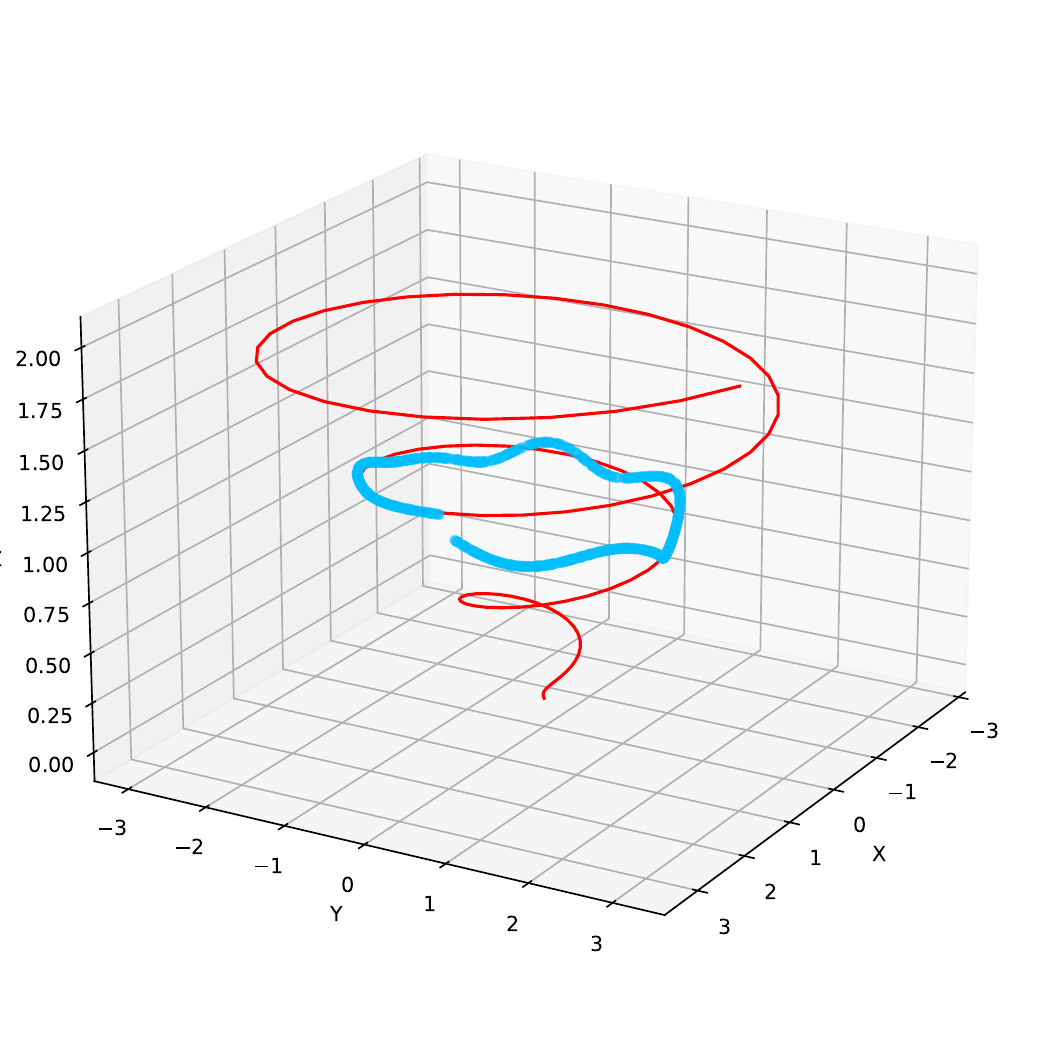}
		\caption{Regular SCMS on angular-linear coordinates in $\Omega_2 \times \mathbb{R}$.}
	\end{subfigure}
	\hfil
	\begin{subfigure}[t]{.24\linewidth}
		\centering
		\includegraphics[width=1\linewidth]{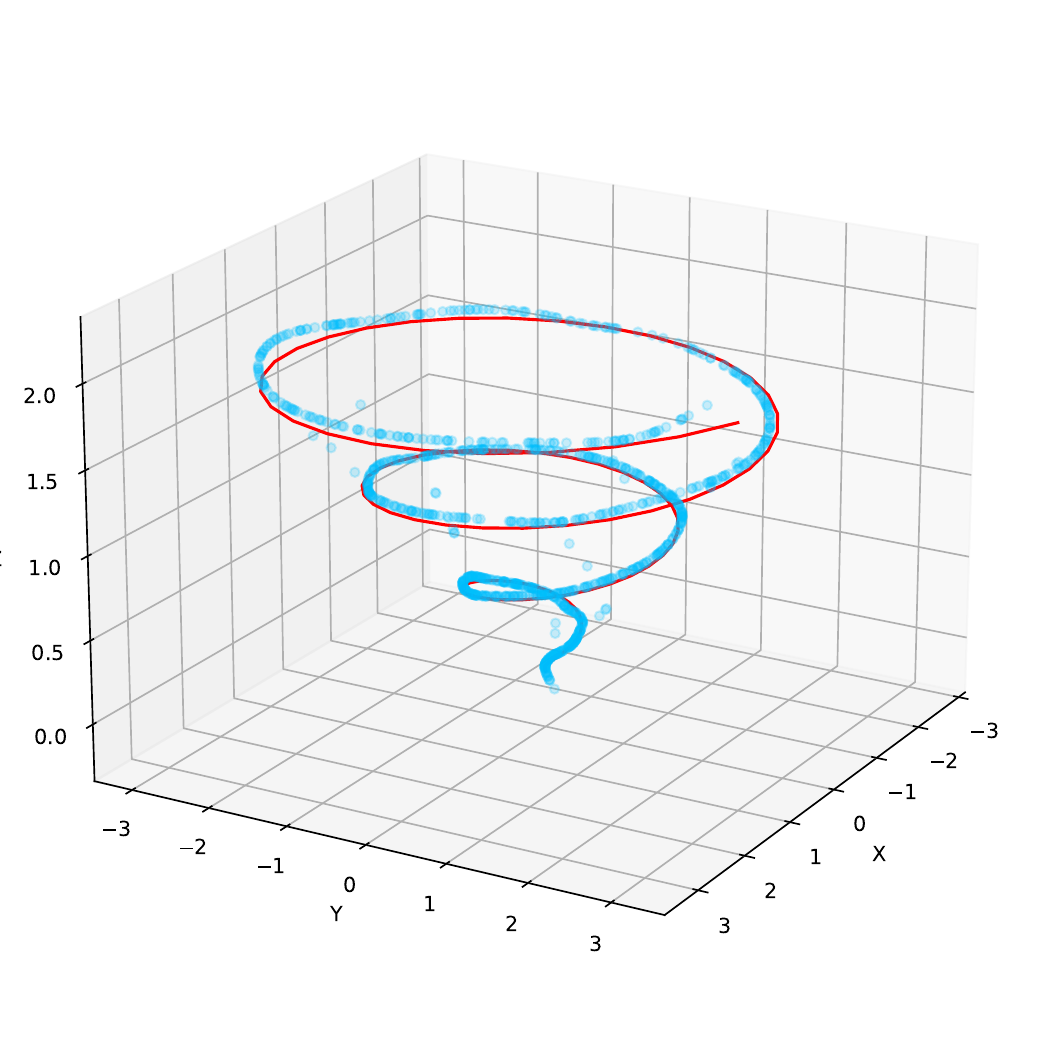}
		\caption{{\bf Our proposed SCMS algorithm in $\Omega_2 \times \mathbb{R}$.}}
	\end{subfigure}
	\caption{Estimated ridges from various SCMS algorithms on spiral curve data. In each panel, the red curve represents the hidden manifold, while blue dots show simulated noisy observations or the final convergent points of the SCMS algorithms.}
	\label{fig:spiral_cur}
\end{figure*}

{\bf Main Results}. $\bullet$ For the mode-seeking problem, we derive two versions of the mean shift algorithm with coordinate-adaptive bandwidths on $\mathcal{S}_1\times \mathcal{S}_2$ and establish their (linear) convergence properties; see \autoref{Sec:MS_prod}.

$\bullet$ While the formulation of the mean shift algorithm on $\mathcal{S}_1\times \mathcal{S}_2$ is natural and intuitive, its extension to subspace-constrained settings is nontrivial and demands careful consideration. We introduce our subspace constrained mean shift (SCMS) algorithm on $\mathcal{S}_1\times \mathcal{S}_2$ and rigorously prove its (linear) convergence. Furthermore, we provide a theoretically and empirically valid guideline for tuning the step size parameter in the algorithm; see \autoref{Sec:SCMS_prod}.

$\bullet$ The effectiveness of our proposed mean shift and SCMS algorithms on $\mathcal{S}_1\times \mathcal{S}_2$ is demonstrated through simulation studies and real-world applications; see \autoref{fig:spiral_cur} with more details in \autoref{Sec:Experiments} and \autoref{Sec:sim_setup}. 
%The code for our experiments can be found at \url{https://github.com/zhangyk8/ProdSCMS}.

{\bf Related Work}. Kernel density estimator (KDE) has been widely applied across various scientific fields; see \cite{Silverman1986,Scott2015,KDE_t} for comprehensive reviews. Relevant extensions of KDE include those for directional-linear (DirLin) or directional-directional (DirDir) data \citep{KDE_torus2011,Dir_Linear2013,garcia2015central}, which utilized directional KDE proposed by \cite{KDE_Sphe1987,KDE_direct1988} with product kernels for density estimation, independence testing, and goodness-of-fit testing. Additional researches on bump-hunting with regular KDE include \cite{parzen1962estimation,chacon2013data,chacon2023bump}. Building on these works, we extend the usual KDE to address mode-seeking and ridge-finding problems on directional and/or linear product spaces. After the first version of our paper, \cite{garcia2023hippocampus,garcia2024kernel} also studied KDE and ridge-finding algorithms for data on the polysphere to analyze hippocampus shape variation. While their approaches are similar to our methods when $\mathcal{S}_1,\mathcal{S}_2$ are directional, our framework also accommodates cases where one of $\mathcal{S}_1,\mathcal{S}_2$ is linear.

The mean shift algorithm has also been adapted to directional \citep{Multi_Clu_Gene2005,DMS_topology2010,DirMS2020} and manifold data \citep{subbarao2006nonlinear,subbarao2009nonlinear,cetingul2009intrinsic}; see \cite{carreira2015review} for a comprehensive review. 
It efficiently identifies local modes of KDE and can be modified into a subspace constrained version to detect density ridges \citep{Non_ridge_est2014,chen2014generalized}. Convergence properties of the (directional) subspace constrained mean shift (SCMS) algorithm and its extensions have been studied in \cite{SCMS_conv2013,qiao2021algorithms,DirSCMS2021}, exhibiting linear convergence to estimated ridges. We will demonstrate that our extended SCMS algorithm on $\mathcal{S}_1\times \mathcal{S}_2$ enriches these approaches.

\subsection{Setup and Notation} 

Our data consist of independent and identically distributed observations $\left\{\bm{Z}_i\right\}_{i=1}^n = \left\{(\bm{X}_i,\bm{Y}_i) \right\}_{i=1}^n$ sampled from a distribution on $\mathcal{S}_1\times \mathcal{S}_2$ with density $f$, where $\int_{\mathcal{S}_1\times \mathcal{S}_2} f(\bm{x},\bm{y}) \,\omega(d\bm{x},d\bm{y})=1$ with $\omega(d\bm{x},d\bm{y})$ inducing a volume form on $\mathcal{S}_1\times \mathcal{S}_2$ under associated Riemannian metrics \citep{pennec2006intrinsic}. If $\mathcal{S}_j=\mathbb{R}^{D_j}$, we assume that the marginal density of $f$ on $\mathcal{S}_j$ has a compact support. Note that $D_j$ is the intrinsic or manifold dimension of $\mathcal{S}_j$, and the ambient space of $\mathcal{S}_j$ is $\mathbb{R}^{D_j+\mathbbm{1}_{\{\mathcal{S}_j=\Omega_{D_j}\}}}$ for $j=1,2$, where $\mathbbm{1}_{\{\mathcal{S}_j=\Omega_{D_j}\}}$ indicates that the directional data space $\Omega_{D_j}$ has its ambient space as $\mathbb{R}^{D_j+1}$. Whenever $\mathcal{S}_1$ and/or $\mathcal{S}_2$ are directional, we extend the (density) function $f$ from $\mathcal{S}_1\times \mathcal{S}_2$ to its ambient space $\mathcal{S} \setminus N_A$ as:
\begin{equation}
\label{den_ext}
\begin{cases}
	f(\bm{x},\cdot) \equiv  f\left(\frac{\bm{x}}{\norm{\bm{x}}_2},\cdot \right) & \text{ if } \quad \mathcal{S}_1 =\Omega_{D_1},\\
	f(\cdot,\bm{y}) \equiv  f\left(\cdot, \frac{\bm{y}}{\norm{\bm{y}}_2} \right) & \text{ if } \quad \mathcal{S}_2 =\Omega_{D_2},
\end{cases}
\end{equation}
where $\mathcal{S}=\mathbb{R}^{D_1+\mathbbm{1}_{\{\mathcal{S}_1= \Omega_{D_1}\}}} \times \mathbb{R}^{D_2 + \mathbbm{1}_{\{\mathcal{S}_2= \Omega_{D_2}\}}}$ and 
$N_A = \left\{(\bm{x},\bm{y}) \in \mathcal{S}: \left(\bm{x}\cdot \mathbbm{1}_{\{\mathcal{S}_1= \Omega_{D_1}\}},\,\bm{y} \cdot \mathbbm{1}_{\{\mathcal{S}_2= \Omega_{D_2}\}} \right) =\bm{0} \right\}$. 
The indicators $\mathbbm{1}_{\{\bm{x}\in \Omega_{D_j}\}}$ with $j=1,2$ are introduced to accommodate the directional data space $\Omega_{D_j}$, whose ambient space is $\mathbb{R}^{D_j+1}$ for $j=1,2$. The set $N_A$ is nonempty only when any of $\mathcal{S}_j, j=1,2$ is directional. The extension \eqref{den_ext} is standard for handling directional densities \citep{Exact_Risk_bw2013,Dir_Linear2013,garcia2015central,DirMS2020,DirSCMS2021}. Under some differentiability assumption (see \autoref{Sec:Assum_Cons} in the supplement), the \emph{total} gradient and Hessian of $f$ in the ambient space of $\mathcal{S}_1\times \mathcal{S}_2$ are denoted by
\begin{align}
\begin{split}
	\label{tot_grad_hess}
	\nabla f(\bm{z})
	&= 
	\begin{pmatrix}
		\nabla_{\bm{x}} f\\
		\nabla_{\bm{y}} f
	\end{pmatrix}(\bm{z}),\,\,
	\nabla^2 f(\bm{z})
	= \begin{pmatrix}
		\nabla_{\bm{x}}^2 f & \nabla_{\bm{y}} \nabla_{\bm{x}} f\\
		\nabla_{\bm{x}}\nabla_{\bm{y}} f & \nabla_{\bm{y}}^2 f
	\end{pmatrix}(\bm{z}),
\end{split}
\end{align}
where $\nabla_{\bm{x}} f(\bm{z}) = \left(\frac{\partial f(\bm{z})}{\partial x_1},..., \frac{\partial f(\bm{z})}{\partial x_D}\right)^{\top}$ with $\bm{z}=(\bm{x},\bm{y})$ denotes the total gradient inside the (smallest) ambient Euclidean space $\mathbb{R}^D$ containing $\mathcal{S}_1$, and the same applies to $\nabla_{\bm{y}} f(\bm{z})$. That is, $D=D_1+1$ when $\mathcal{S}_1=\Omega_{D_1}$ (directional space) while $D=D_1$ when $\mathcal{S}_1 =\mathbb{R}^{D_1}$ (Euclidean space). 
We use the big-O notation $h(x)=O(g(x))$ or $h(x) \lesssim g(x)$ if the absolute value of $h(x)$ is upper bounded by a positive constant multiple of $g(x)$ for all sufficiently large $x$. If $h(x)\gtrsim g(x)$ and $h(x) \lesssim g(x)$, then $h(x),g(x)$ are asymptotically equal and it is denoted by $h(x)\asymp g(x)$. In contrast, $h(x)=o(g(x))$ when $\lim_{x\to\infty} |h(x)|/g(x)=0$. For random vectors, the notation $o_P(1)$ is short for a sequence of random vectors that converges to zero in probability, and $O_P(1)$ denotes a sequence that is bounded in probability.

\section{Background}

% This section reviews the construction of KDE in a Euclidean and/or directional product space $\mathcal{S}_1\times \mathcal{S}_2$ and derives the Riemannian gradient and Hessian within this space.

\subsection{Kernel Density Estimator on $\mathcal{S}_1\times \mathcal{S}_2$}

Given the observed data $\left\{\bm{Z}_i\right\}_{i=1}^n = \left\{(\bm{X}_i,\bm{Y}_i) \right\}_{i=1}^n$ on the product space $\mathcal{S}_1 \times \mathcal{S}_2$, we estimate the underlying density $f$ via KDE with product kernels as \citep{hall2006real,Dir_Linear2013,garcia2015central}:
\begin{align}
\label{KDE_prod}
\begin{split}
	\hat{f}_{\bm{h}}(\bm{x},\bm{y}) = \frac{1}{n} \sum_{i=1}^n K_1\left(\frac{\bm{x}-\bm{X}_i}{h_1} \right) K_2\left(\frac{\bm{y}-\bm{Y}_i}{h_2} \right),
\end{split}
\end{align}
where $K_j: \mathcal{S}_j \to \mathbb{R}$ is the kernel function for $j=1,2$, and each element of $\bm{h}=(h_1,h_2)$ is a bandwidth parameter. Depending on the geometry of $\mathcal{S}_j$, the radially symmetric kernel function $K_j$ takes the form:
\begin{align}
\label{kernel_fun}
\begin{split}
	K_j(\bm{u}) &= C_{k_j,D_j}(h_j) \cdot k_j\left(\norm{\bm{u}}_2^2 \right) =
	\begin{cases}
		\frac{C_{k,D_j}}{h_i^{D_j}}\cdot  k\left(\norm{\bm{u}}_2^2 \right) & \text{ if } \mathcal{S}_j =\mathbb{R}^{D_j},\\
		C_{L,D_j}(h_j) \cdot L\left(\frac{\norm{\bm{u}}_2^2}{2} \right) & \text{ if } \mathcal{S}_j =\Omega_{D_j},
	\end{cases}
\end{split}
\end{align}
for $j=1,2$, where $k$ and $L$ are the profiles of linear and directional kernels respectively while $C_{k,D_j}, C_{L,D_j}(h_j)$ are normalizing constants. A key distinction between linear and directional kernels in \eqref{kernel_fun} is whether the normalizing constants are separable from the bandwidth parameters. In particular, if $\mathcal{S}_1$ is directional, then $K_1\left(\frac{\bm{x}-\bm{X}_i}{h_1}\right) = C_{L,D_1}(h_1) \cdot L\left(\frac{\norm{\bm{x}-\bm{X}_i}_2^2}{2h_1^2} \right) = C_{L,D_1}(h_1) \cdot L\left(\frac{1-\bm{x}^T\bm{X}_i}{h_1^2} \right)$, which aligns with the usual KDE formulation for directional data \citep{KDE_Sphe1987,KDE_direct1988}. Furthermore, $\hat{f}_{\bm{h}}$ is defined not just on $\mathcal{S}_1\times\mathcal{S}_2$ but also its ambient space. Finally, if $\mathcal{S}_1,\mathcal{S}_2$ are both directional or Euclidean, $\hat{f}_{\bm{h}}$ can also be defined via a spherically symmetric kernel \citep{garcia2024kernel}.
% While it is intuitive to see that the normalizing constant has asymptotic rate $O\left(1/h_j^{D_j} \right)$ under the linear kernel, it is less obvious to obtain the similar rate $C_{L,D_j}(h_j)^{-1} = O\left(h_j^{D_j} \right)$ for the directional kernel; see Section 2 in \cite{Exact_Risk_bw2013}. 

The commonly used linear and directional kernel profiles are $k(s) = e^{-s/2}$ and $L(r)=e^{-r}$, corresponding to the Gaussian kernel and von Mises kernel, respectively. The von Mises kernel derives its name from the classical von Mises-Fisher (vMF) distribution \citep{Mardia2000directional} on $\Omega_q$, which has the density $f_{\text{vMF}}(\bm{x};\bm{\mu},\kappa) = C_q(\kappa) \cdot \exp\left(\kappa \bm{\mu}^{\top}\bm{x} \right)$, where $C_q(\kappa)=\frac{\kappa^{\frac{q-1}{2}}}{(2\pi)^{\frac{q+1}{2}} \mathcal{I}_{\frac{q-1}{2}}(\kappa)}$ is a normalizing constant, $\bm{\mu} \in \Omega_q$ is the directional mean of the vMF distribution, $\kappa \geq 0$ is the concentration parameter, and $\mathcal{I}_{\alpha}(\kappa)$ is the modified Bessel function of the first kind at order $\alpha$.
Applying Gaussian and/or von Mises kernels to $K_1$ and $K_2$ in \eqref{KDE_prod}, the KDE simplifies to
\begin{align}
\label{KDE_Exp}
\begin{split}
	&\hat{f}_{\bm{h}}(\bm{z}) = \frac{C(\bm{H})}{n} \sum_{i=1}^n \exp\left[-\frac{(\bm{z}-\bm{Z}_i)^{\top} \bm{H}^{-1} (\bm{z}-\bm{Z}_i)}{2} \right] \\ 
	&\text{ with } \quad \bm{H} =\Diag\left(h_1^2\bm{I}_{D_1+\mathbbm{1}_{\{\mathcal{S}_1=\Omega_{D_1}\}}}, h_2^2\bm{I}_{D_2 + \mathbbm{1}_{\{\mathcal{S}_2=\Omega_{D_2}\}}} \right),
\end{split}
\end{align}
where $\bm{z}=(\bm{x},\bm{y}) \in \mathcal{S}_1\times \mathcal{S}_2$, $\bm{I}_D$ is the identity matrix in $\mathbb{R}^{D\times D}$, and $C(\bm{H}) \equiv \prod_{j=1}^2 C_{k_j,D_j}(h_j)$ is the normalizing constant from \eqref{KDE_prod} and \eqref{kernel_fun}. For simplicity, we consider a (block) diagonal bandwidth matrix $\bm{H}$ \citep{wand1993comparison}, though our theory also applies to the general positive definite bandwidth matrix. Additionally, when $\mathcal{S}_j=\Omega_{D_j}$, it is more appropriate to apply a single bandwidth parameter $h_j$ to each coordinate on $\Omega_{D_j}$ due to the isotropic geometry. While the (subspace constrained) mean shift algorithms can be derived for general kernels in Sections~\ref{Sec:MS_prod} and \ref{Sec:SCMS_prod}, we only implement the algorithms with Gaussian and/or von Mises kernels in practice, given their smoothness properties.

\subsection{Gradient and Hessian System on $\mathcal{S}_1\times \mathcal{S}_2$}
\label{Sec:Grad_Hess}

Given that each component of the product space $\mathcal{S}_1\times \mathcal{S}_2$ may be a nonlinear manifold $\Omega_q$, we study the Riemannian gradient and Hessian of a density function $f$ defined on the tangent spaces of $\mathcal{S}_1\times \mathcal{S}_2$ and their connections to the total gradient and Hessian in the ambient (Euclidean) space.
We assume that the function $f$ is at least twice continuously differentiable in an open neighborhood of the ambient space containing $\mathcal{S}_1\times \mathcal{S}_2$; see the full assumptions in \autoref{Sec:Assum_Cons} of the supplement. 

$\bullet$ {\bf Riemannian Gradient}. The differential of $f$ at $\bm{z}=(\bm{x},\bm{y})\in \mathcal{S}_1\times \mathcal{S}_2$ is a linear map $df_{\bm{z}}: T_{\bm{z}} \equiv T_{\bm{x}}(\mathcal{S}_1)\times T_{\bm{y}}(\mathcal{S}_2) \to \mathbb{R}$ given by $df_{\bm{z}}(\bm{v}) = \frac{d}{dt} f\left(\gamma(t)\right) \big|_{t=0}$ for any smooth curve $\gamma:[0,1]\to \mathcal{S}_1\times \mathcal{S}_2$ and tangent vector $\bm{v}=(\bm{v}_{\bm{x}}, \bm{v}_{\bm{y}}) \in T_{\bm{z}} \equiv T_{\bm{x}}(\mathcal{S}_1)\times T_{\bm{y}}(\mathcal{S}_2)$. Then, the \emph{Riemannian gradient} $\grad f(\bm{z})$ is a vector field in the tangent space $T_{\bm{z}}$ defined as:
\begin{align}
\begin{split}
	\label{Riem_grad_def}
	df_{\bm{z}}(\bm{v}) &= \langle \grad f(\bm{z}), (\bm{v}_{\bm{x}}, \bm{v}_{\bm{y}}) \rangle = \grad f(\bm{z})^{\top} \bm{v},
\end{split}
\end{align}
where the inner product $\langle\cdot,\cdot\rangle$ is the usual one in the smallest ambient Euclidean space containing $\mathcal{S}_1\times\mathcal{S}_2$. From \eqref{Riem_grad_def}, we obtain a vector form of $\grad f(\bm{x},\bm{y})$ as:
\begin{align}
\begin{split}
	\label{Riem_grad}
	&\grad f(\bm{z}) = \mathcal{P}_{\bm{z}} \nabla f(\bm{z}) \; \text{ with } \; \mathcal{P}_{\bm{z}} = \begin{pmatrix}
		\mathcal{P}_1 & \bm{0}\\
		\bm{0} & \mathcal{P}_2
	\end{pmatrix} \text{ and } \begin{cases}
		\mathcal{P}_1 = \bm{I}_{D_1+\mathbbm{1}_{\{\mathcal{S}_1= \Omega_{D_1}\}}}-\bm{x}\bm{x}^{\top}\cdot \mathbbm{1}_{\{\mathcal{S}_1= \Omega_{D_1}\}},\\
		\mathcal{P}_2 = \bm{I}_{D_2+\mathbbm{1}_{\{\mathcal{S}_2= \Omega_{D_2}\}}}-\bm{y}\bm{y}^{\top}\cdot \mathbbm{1}_{\{\mathcal{S}_2= \Omega_{D_2}\}}.
	\end{cases}
\end{split}
\end{align}
Here, $\mathcal{P}_{\bm{z}}$ is the projection matrix onto the tangent space $T_{\bm{z}}$ and $\bm{I}_D$ denotes an identity matrix in $\mathbb{R}^{D\times D}$. Specifically, $\mathcal{P}_{\bm{z}}$ reduces to the identity matrix when $\mathcal{S}_1\times \mathcal{S}_2$ is Euclidean, while it acts as a projection operator when either of $\mathcal{S}_j,j=1,2$ is directional.

$\bullet$ {\bf Riemannian Hessian}. The \emph{Riemannian Hessian} of $f$ is a symmetric linear map $\mathcal{H}f(\bm{z})$ of the tangent space $T_{\bm{z}}\equiv T_{\bm{x}}(\mathcal{S}_1)\times T_{\bm{y}}(\mathcal{S}_2)$ into itself defined by
\begin{equation}
	\label{Riem_Hess_def}
	\mathcal{H} f(\bm{z})[\bm{v}] = \bm{\bar{\nabla}}_{\bm{v}} \grad f(\bm{z})
\end{equation}
for any tangent vector $\bm{v}=(\bm{v}_{\bm{x}}, \bm{v}_{\bm{y}})\in T_{\bm{z}}$, where $\bm{\bar{\nabla}}_{\bm{v}}$ is the Riemannian connection on $\mathcal{S}_1\times \mathcal{S}_2$; see Section 5.5 in \cite{Op_algo_mat_manifolds2009}. 
As $\mathcal{S}_1\times \mathcal{S}_2$ can be viewed as a (nonlinear) submanifold in its ambient Euclidean space, Proposition 5.3.2 in \cite{Op_algo_mat_manifolds2009} suggests that the Riemannian Hessian $\mathcal{H} f(\bm{z})$ can be written as:
\begin{align}
\label{Riem_Hess_def2}
\begin{split}
	\mathcal{H} f(\bm{z})[\bm{v}] &= 
	\mathcal{P}_{\bm{z}}
	\Big(
	\nabla_{\bm{x}}\grad f(\bm{z}),\, 
	\nabla_{\bm{y}} \grad f(\bm{z})
	\Big)[\bm{v}]\\
	&= \mathcal{P}_{\bm{z}} \Big[\nabla^2 f(\bm{z}) - \Diag\left(\mathcal{A}_1, \mathcal{A}_2\right) \Big][\bm{v}] \; \text{ with } \;
	\begin{cases}
		\mathcal{A}_1 = \bm{x}^{\top}\nabla_{\bm{x}} f(\bm{z}) \bm{I}_{D_1+1} \cdot \mathbbm{1}_{\{\mathcal{S}_1= \Omega_{D_1}\}},\\
		\mathcal{A}_2= \bm{y}^{\top}\nabla_{\bm{y}} f(\bm{z}) \bm{I}_{D_2+1} \cdot \mathbbm{1}_{\{\mathcal{S}_2= \Omega_{D_2}\}},
	\end{cases}
\end{split}
\end{align}
where $\Diag\left(\mathcal{A}_1, \mathcal{A}_2 \right)$ is a (block) diagonal matrix, and we plug in the formula \eqref{Riem_grad} of $\grad f(\bm{z})$ in the second equality. 
$\mathcal{A}_j$ is the ``corrections'' when the space $\mathcal{S}_j$ is directional/nonlinear for $j=1,2$. 
When the space $\mathcal{S}_j$ is Euclidean, $\mathcal{A}_j$ becomes a matrix $\bm{0}$.
Using $\bm{v} \in T_{\bm{z}}$ and the symmetric property of $\mathcal{H} f(\bm{z})$, we express the Hessian operator $\mathcal{H} f(\bm{z})$ as a matrix $\mathcal{H} f(\bm{z}) = \mathcal{P}_{\bm{z}} \Big[\nabla^2 f(\bm{z}) - \Diag\left(\mathcal{A}_1, \mathcal{A}_2 \right) \Big] \mathcal{P}_{\bm{z}}.$
Furthermore, both $\grad f(\bm{z})$ and $\mathcal{H}f(\bm{z})$ reduce to the total gradient $\nabla f(\bm{z})$ and Hessian $\nabla^2 f(\bm{z})$ when $\mathcal{S}_1,\mathcal{S}_2$ are flat Euclidean spaces.

\section{Mean Shift Algorithm on $\mathcal{S}_1\times \mathcal{S}_2$}
\label{Sec:MS_prod}

% This section introduces the definition of local modes of $f$ on a Euclidean and/or directional product space $\mathcal{S}_1\times \mathcal{S}_2$ and presents our proposed mean shift algorithm on this space.

\subsection{Mode Estimation on $\mathcal{S}_1\times \mathcal{S}_2$} 
\label{subsec:mode_est}

The Riemannian gradient $\grad f(\bm{z})$ and Hessian $\mathcal{H}f(\bm{z})$ in \autoref{Sec:Grad_Hess} induce the definition of the set of local modes $\mathcal{M}$ of $f$ on $\mathcal{S}_1\times \mathcal{S}_2$ as:
\begin{equation}
\label{ProdDen_Mode}
\mathcal{M} = \left\{\bm{z} \in \mathcal{S}_1\times \mathcal{S}_2: \grad f(\bm{z})=\bm{0}, \lambda_1(\bm{z})<0 \right\},
\end{equation}
where $\lambda_1(\bm{z})$ is the largest eigenvalue of $\mathcal{H}f(\bm{z})$ within the tangent space $T_{\bm{z}}$. Given the KDE $\hat{f}_{\bm{h}}$ in \eqref{KDE_prod}, a plug-in estimator of $\mathcal{M}$ is given by
\begin{equation}
\label{ProdDen_Mode_est}
\hat{\mathcal{M}} = \left\{\bm{z} \in \mathcal{S}_1\times \mathcal{S}_2: \grad \hat{f}_{\bm{h}}(\bm{z})=\bm{0}, \hat{\lambda}_1(\bm{z})<0 \right\}.
\end{equation}
Under some regularity conditions on kernel functions and the density $f$, $\hat{\mathcal{M}}$ is a consistent estimator of $\mathcal{M}$; see \autoref{Sec:Mode_cons} of the supplement.

\subsection{Derivations of the Mean Shift Algorithm on $\mathcal{S}_1\times \mathcal{S}_2$}

We assume that the linear and directional kernel profiles $k$ and $L$ in \eqref{kernel_fun} are continuously differentiable. 
% The specialization with Gaussian and/or von Mises kernels will also be elucidated because of its efficacy in practice. 
Under \eqref{KDE_prod} and \eqref{kernel_fun}, the total gradient of $\hat{f}_{\bm{h}}(\bm{z})$ becomes
\begin{align}
\label{KDE_tot_grad}
\begin{split}
	&\nabla \hat{f}_{\bm{h}}(\bm{z}) = \begin{pmatrix}
		\nabla_{\bm{x}} \hat{f}_{\bm{h}}(\bm{z})\\
		\nabla_{\bm{y}} \hat{f}_{\bm{h}}(\bm{z})
	\end{pmatrix} = \begin{pmatrix}
		G_{\bm{x}} \left[\frac{\sum\limits_{i=1}^n \bm{X}_i k_1'\left(\norm{\frac{\bm{x} -\bm{X}_i}{h_1}}_2^2 \right)  K_2\left(\frac{\bm{y}-\bm{Y}_i}{h_2} \right) }{\sum\limits_{i=1}^n k_1'\left(\norm{\frac{\bm{x} -\bm{X}_i}{h_1}}_2^2 \right) K_2\left(\frac{\bm{y}-\bm{Y}_i}{h_2} \right)} -\bm{x} \right]\\
		G_{\bm{y}} \left[\frac{\sum\limits_{i=1}^n \bm{Y}_i K_1\left(\frac{\bm{x} -\bm{X}_i}{h_1} \right)  k_2'\left(\norm{\frac{\bm{y}-\bm{Y}_i}{h_2}}_2^2 \right) }{\sum\limits_{i=1}^n K_1\left(\frac{\bm{x} -\bm{X}_i}{h_1} \right)  k_2'\left(\norm{\frac{\bm{y}-\bm{Y}_i}{h_2}}_2^2 \right)} -\bm{y} \right]
	\end{pmatrix},
\end{split}
\end{align}
where, with $C(\bm{H}) \equiv \prod_{j=1}^2 C_{k_j,D_j}(h_j)$, both factors 
\begin{align}
\label{MS_factor}
\begin{split}
	G_{\bm{x}} &= -\frac{2 C(\bm{H})}{nh_1^2} \sum_{i=1}^n k_1'\left(\norm{\frac{\bm{x}-\bm{X}_i}{h_1}}_2^2\right)K_2\left(\frac{\bm{y}-\bm{Y}_i}{h_2}\right),\\ 
	G_{\bm{y}} &= -\frac{2 C(\bm{H})}{nh_2^2} \sum_{i=1}^n K_1\left(\frac{\bm{x}-\bm{X}_i}{h_1}\right)k_2'\left(\norm{\frac{\bm{y}-\bm{Y}_i}{h_2}}_2^2\right)
\end{split}
\end{align}
are non-negative and $\bm{z}=(\bm{x},\bm{y})\in \mathcal{S}_1\times \mathcal{S}_2$. Thus, the mean shift vector $\Xi(\bm{z})$ in $\mathcal{S}_1 \times \mathcal{S}_2$ is defined as:
\begin{align}
\label{MS_vec}
\begin{split}
	\Xi(\bm{z}) &= \begin{pmatrix}
		\Xi_{\bm{x}}(\bm{x},\bm{y})\\
		\Xi_{\bm{y}}(\bm{x},\bm{y})
	\end{pmatrix} = \begin{pmatrix}
		\frac{\sum\limits_{i=1}^n \bm{X}_i k_1'\left(\norm{\frac{\bm{x} -\bm{X}_i}{h_1}}_2^2 \right)  K_2\left(\frac{\bm{y}-\bm{Y}_i}{h_2} \right) }{\sum\limits_{i=1}^n k_1'\left(\norm{\frac{\bm{x} -\bm{X}_i}{h_1}}_2^2 \right) K_2\left(\frac{\bm{y}-\bm{Y}_i}{h_2} \right)} -\bm{x} \\
		\frac{\sum\limits_{i=1}^n \bm{Y}_i K_1\left(\frac{\bm{x} -\bm{X}_i}{h_1} \right)  k_2'\left(\norm{\frac{\bm{y}-\bm{Y}_i}{h_2}}_2^2 \right) }{\sum\limits_{i=1}^n K_1\left(\frac{\bm{x} -\bm{X}_i}{h_1} \right)  k_2'\left(\norm{\frac{\bm{y}-\bm{Y}_i}{h_2}}_2^2 \right)} -\bm{y}
	\end{pmatrix}.
\end{split}
\end{align}
Unlike the standard mean shift vector in the Euclidean space \citep{MS2002}, $\Xi(\bm{z})$ \emph{does not} align with the total gradient estimator $\nabla\hat{f}_{\bm{h}}(\bm{z})$ in \eqref{KDE_tot_grad}, because $G_{\bm{x}}\neq G_{\bm{y}}$ in general. 
%One exception occurs when the bandwidths $h_1=h_2$ and the Gaussian and/or von Mises kernels are applied, under which the mean shift vector \eqref{MS_vec} becomes
%\begin{equation}
%\label{MS_vec_exp}
%\Xi(\bm{z}) = \frac{\sum\limits_{i=1}^n \bm{Z}_i \exp\left[-\frac{1}{2}(\bm{z}-\bm{Z}_i)^{\top}\bm{H}^{-1} (\bm{z}-\bm{Z}_i)\right] }{\sum\limits_{i=1}^n \exp\left[-\frac{1}{2}(\bm{z}-\bm{Z}_i)^{\top}\bm{H}^{-1} (\bm{z}-\bm{Z}_i)\right]} -\bm{z}.
%\end{equation}
%Yet, we will not require $h_1=h_2$ in the following discussion because $\mathcal{S}_1, \mathcal{S}_2$ may have different domains. 
Nevertheless, $\Xi(\bm{z})$ and $\nabla\hat{f}_{\bm{h}}(\bm{z})$ share the same set of roots within the tangent space $T_{\bm{z}}$.
Specifically, the solutions to $\mathcal{P}_{\bm{z}}\Xi(\bm{z})=\bm{0}$ coincide with the set of estimated local modes $\hat{\mathcal{M}}$, satisfying $\grad\hat{f}_{\bm{h}}(\bm{z})=\bm{0}$. Hence,
$\hat{\mathcal{M}}^{(T)} := \left\{\bm{z}\in \mathcal{S}_1\times \mathcal{S}_2: \mathcal{P}_{\bm{z}} \cdot \Xi(\bm{z})=\bm{0} \right\} = \hat{\mathcal{M}}$.
This equivalence arises because, from \eqref{KDE_tot_grad} and \eqref{MS_vec}, the mean shift vector $\Xi(\bm{z})$ is equal to $\tilde{D}(\bm{z})^{-1}\nabla\hat{f}_{\bm{h}}(\bm{z})$, which is a transformed estimated total gradient with
\begin{align}
\label{trans_tot_grad_mat}
\begin{split}
	\tilde{D}(\bm{z})= \Diag\left(G_{\bm{x}} \bm{I}_{D_1+\mathbbm{1}_{\{\mathcal{S}_1= \Omega_{D_1}\}}}, G_{\bm{y}} \bm{I}_{D_2+\mathbbm{1}_{\{\mathcal{S}_2= \Omega_{D_2}\}}} \right).
\end{split}
\end{align}
Since $\tilde{D}(\bm{z})$ is nonsingular for any strictly decreasing kernel (\emph{e.g.}, Gaussian or von Mises kernels), the projection of $\tilde{D}(\bm{z})^{-1}\nabla\hat{f}_{\bm{h}}(\bm{z})=\Xi(\bm{z})$ onto $T_{\bm{z}}$ is zero if and only if the Riemannian gradient estimator $\grad\hat{f}_{\bm{h}}(\bm{z})$ is zero.
Therefore, $\Xi(\bm{z})$ remains valid for identifying the local modes of $\hat{f}_{\bm{h}}(\bm{z})$. 
Here, we introduce two different versions of the mean shift algorithm on $\mathcal{S}_1\times \mathcal{S}_2$ based on $\Xi(\bm{z})$ and establish their (linear) convergence in Theorems~\ref{Thm:MS_conv} and \ref{Thm:MS_lin_conv}.

$\bullet$ {\bf Version A: Simultaneous Mean Shift}. This version updates all the components $\bm{z}^{(t)}=\big(\bm{x}^{(t)},\bm{y}^{(t)} \big)^{\top} \in \mathcal{S}_1\times \mathcal{S}_2$ simultaneously, in which for $t=0,1,...$,
\begin{align}
\label{MS_iter_sim}
\begin{split}
	\bm{z}^{(t+1)}  &\gets \Xi(\bm{z}^{(t)}) + \bm{z}^{(t)} = \begin{pmatrix}
		\frac{\sum\limits_{i=1}^n \bm{X}_i k_1'\left(\norm{\frac{\bm{x}^{(t)} -\bm{X}_i}{h_1}}_2^2 \right)  K_2\left(\frac{\bm{y}^{(t)}-\bm{Y}_i}{h_2} \right) }{\sum\limits_{i=1}^n k_1'\left(\norm{\frac{\bm{x}^{(t)} -\bm{X}_i}{h_1}}_2^2 \right) K_2\left(\frac{\bm{y}^{(t)}-\bm{Y}_i}{h_2} \right)}\\ 
		\frac{\sum\limits_{i=1}^n \bm{Y}_i K_1\left(\frac{\bm{x}^{(t)} -\bm{X}_i}{h_1} \right)  k_2'\left(\norm{\frac{\bm{y}^{(t)}-\bm{Y}_i}{h_2}}_2^2 \right) }{\sum\limits_{i=1}^n K_1\left(\frac{\bm{x}^{(t)} -\bm{X}_i}{h_1} \right)  k_2'\left(\norm{\frac{\bm{y}^{(t)}-\bm{Y}_i}{h_2}}_2^2 \right)}
	\end{pmatrix}
\end{split}
\end{align}
with extra standardizations $\bm{x}^{(t+1)} \gets \frac{\bm{x}^{(t+1)}}{\norm{\bm{x}^{(t+1)}}_2}$ and/or $\bm{y}^{(t+1)} \gets \frac{\bm{y}^{(t+1)}}{\norm{\bm{y}^{(t+1)}}_2}$ if $\mathcal{S}_1$ and/or $\mathcal{S}_2$ are directional. 
This approach seamlessly integrates the Euclidean and directional mean shift algorithms into a unified iteration formula, as shown in \eqref{MS_iter_sim}. 

$\bullet$ {\bf Version B: Componentwise Mean Shift}. This version updates the sequence $\left\{\bm{z}^{(t)}\right\}_{t=0}^{\infty}=\left\{(\bm{x}^{(t)}, \bm{y}^{(t)})\right\}_{t=0}^{\infty}$ in a two-step manner as:
\begin{align}
\label{MS_iter_comp}
\begin{split}
	&\bm{x}^{(t+1)} \gets \Xi_{\bm{x}}(\bm{x}^{(t)}, \bm{y}^{(t)}) + \bm{x}^{(t)} \quad \text{ with } \quad \bm{x}^{(t+1)} \gets \frac{\bm{x}^{(t+1)}}{\norm{\bm{x}^{(t+1)}}_2^2}\quad \text{ if }\; \mathcal{S}_1 = \Omega_{D_1},\\
	&\bm{y}^{(t+1)} \gets \Xi_{\bm{y}}(\bm{x}^{(t+1)},\bm{y}^{(t)}) + \bm{y}^{(t)} \quad \text{ with } \quad \bm{y}^{(t+1)} \gets \frac{\bm{y}^{(t+1)}}{\norm{\bm{y}^{(t+1)}}_2^2}\quad \text{ if }\; \mathcal{S}_2 = \Omega_{D_2}
\end{split}
\end{align}
for $t=0,1,...$. The formula updates the two components, $\bm{x}^{(t)}$ and $\bm{y}^{(t)}$, in an alternating manner: first holding $\bm{y}^{(t)}$ constant to update $\bm{x}^{(t)}$ and then reversing the roles. This iteration draws inspiration from the coordinate ascent/descent algorithm \citep{wright2015coordinate}.

The above two versions can be readily specialized to mean shift algorithms using von Mises and/or Gaussian kernels, and the details are therefore omitted. Under mild conditions on the kernel profiles, we establish the ascending and convergence properties of the mean shift algorithm on $\mathcal{S}_1 \times \mathcal{S}_2$, with proofs provided in \autoref{Sec:Conv_pf}.

\begin{theorem}
\label{Thm:MS_conv}
Denote the sequence from the mean shift algorithm by $\left\{\bm{z}^{(t)}\right\}_{t=0}^{\infty}= \left\{(\bm{x}^{(t)},\bm{y}^{(t)}) \right\}\subset \mathcal{S}_1\times\mathcal{S}_2$. Assume that 

$\bullet$ {\bf (C1)} The kernel profiles $k_1,k_2$ (either linear $k$ or directional $L$) are strictly decreasing and differentiable on $[0,\infty)$ with $k_1(0), k_2(0) < \infty$.

$\bullet$ {\bf (Weak Condition)} Both $k_1$ and $k_2$ are convex.

$\bullet$ {\bf (Strong Condition)} The entire product kernel profile $K(r,s) = k_1(r)\cdot k_2(s)$ is convex.

\noindent Then, for any fixed bandwidths $\bm{h}$ and sample size $n$, we have that

(a) Under {\bf (C1)} and {\bf (Weak Condition)}, the sequence of density estimates $\left\{\hat{f}_{\bm{h}}(\bm{z}^{(t)}) \right\}_{t=0}^{\infty}$ yielded by Version B is non-decreasing and thus converges.

(b) Under {\bf (C1)} and {\bf (Strong Condition)}, the sequence of density estimates $\left\{\hat{f}_{\bm{h}}(\bm{z}^{(t)}) \right\}_{t=0}^{\infty}$ yielded by either Version A or B is non-decreasing and thus converges.

(c) Under the assumptions in (a) or (b), we have that  $\lim\limits_{t\to\infty} \norm{\bm{z}^{(t+1)}-\bm{z}^{(t)}}_2=0$.

(d) Assume the conditions in (a) or (b). If the local modes of $\hat{f}_{\bm{h}}$ are isolated, $\left\{\bm{z}^{(t)}\right\}_{t=0}^{\infty}$ converges to a local mode of $\hat{f}_{\bm{h}}$ when it is initialized within its small neighborhood.
\end{theorem}

\begin{remark}
When the entire product kernel profile $K(r,s) = k_1(r)\cdot k_2(s)$ is convex, its component functions $k_1$ and $k_2$ are also convex. Thus, the strong condition in Theorem~\ref{Thm:MS_conv} implies the weak condition, whereas the converse is not true in general. For Gaussian and/or von Mises kernel profiles $k(s) = L(s) = e^{-s}$, the conditions in (a,b,c) are satisfied. Lastly, the isolated mode condition of $\hat{f}_{\bm{h}}$ in (d) can be derived from regularity conditions on the original density function $f$ under the uniform consistency of $\hat{f}_{\bm{h}}$; see Theorem~\ref{Thm:Mode_cons} in \autoref{Sec:Cons_Theory} of the supplement.
\end{remark}

The result in (d) of Theorem~\ref{Thm:MS_conv} is known as the local convergence of the proposed mean shift algorithm.
For its global convergence (from almost every initial point in $\mathcal{S}_1\times \mathcal{S}_2$), one potential strategy is to interpret the algorithm as a (generalized) Expectation-Maximization (EM) algorithm \citep{MS_EM2007,DMS_EM2021} and apply the convergence theory of EM algorithm \citep{EM_Jeff,EM_ext2008}. Another feasible approach leverages the {\L}ojasiewicz property of $\hat{f}_{\bm{h}}$, as explored in \cite{yamasaki2023convergence}.
Finally, we establish the following linear convergence of our proposed mean shift algorithms \eqref{MS_iter_sim} and \eqref{MS_iter_comp} under the gradient ascent framework on $\mathcal{S}_1\times \mathcal{S}_2$ with their intrinsic step sizes depending on the bandwidths $h_1,h_2$; see \autoref{Sec:Lin_conv} for details and its proof.

\begin{theorem}
\label{Thm:MS_lin_conv}
Assume that the conditions of Theorem~\ref{Thm:Mode_cons} in \autoref{Sec:Mode_cons} of the supplement hold. Given the sequence $\left\{\bm{z}^{(t)}\right\}_{t=0}^{\infty}$ defined by our mean shift algorithm \eqref{MS_iter_sim} or \eqref{MS_iter_comp}, there exist constants $\tilde{r}_1 >0, \Upsilon_1 \in (0,1)$ such that
$$d_g(\bm{z}^{(t)}, \hat{\bm{m}}) \leq \Upsilon_1^t \cdot d_g(\bm{z}^{(0)}, \hat{\bm{m}}) \quad \text{ and } \quad
d_g(\bm{z}^{(t)}, \bm{m}) \leq \Upsilon_1^t \cdot d_g(\bm{z}^{(0)}, \bm{m}) + O(h^2) + O_P\left(\sqrt{\frac{|\log h|}{nh^{D_T}}} \right)$$%
when $\bm{z}^{(0)} \in \left\{\bm{z}\in \mathcal{S}_1\times\mathcal{S}_2: d_g(\bm{z},\bm{m}) \leq \tilde{r}_1 \right\}$ with $\bm{m}\in \mathcal{M}$, $\max\left\{h_1,h_2\right\} \lesssim h$ is sufficiently small, and the sample size $n$ is sufficiently large. Here, $D_T=D_1+D_2$ is the intrinsic dimension and $d_g(\cdot,\cdot)$ is the geodesic distance on $\mathcal{S}_1\times \mathcal{S}_2$.
\end{theorem}

\section{SCMS Algorithm on $\mathcal{S}_1\times \mathcal{S}_2$}
\label{Sec:SCMS_prod}

This section presents the formal definition of (density) ridges of $f$ on $\mathcal{S}_1\times \mathcal{S}_2$ and highlights the challenges associated with generalizing the standard SCMS algorithm to this setting, along with our proposed solution.

\subsection{Ridge Estimation on $\mathcal{S}_1\times \mathcal{S}_2$} 

Given the Riemannian gradient $\grad f(\bm{z})$ and Hessian $\mathcal{H}f(\bm{z})$ on $\mathcal{S}_1\times \mathcal{S}_2$, we generalize the definitions of order-$d$ Euclidean \citep{Eberly1996ridges,Non_ridge_est2014} and directional \citep{DirSCMS2021} density ridges to the product space as:
\begin{align}
\label{ProdDen_Ridge}
\begin{split}
	\mathcal{R}_d =& \big\{\bm{z}=(\bm{x},\bm{y}) \in \mathcal{S}_1\times \mathcal{S}_2: V_d(\bm{z}) V_d(\bm{z})^{\top} \nabla f(\bm{z}) =\bm{0}, \lambda_{d+1}(\bm{z}) < 0 \big\},
\end{split}
\end{align}
where $\lambda_1(\bm{z}) \geq \cdots \geq \lambda_{D_T}(\bm{z})$ are eigenvalues of $\mathcal{H} f(\bm{z})$ within the tangent space, $D_T=D_1+D_2$ is the intrinsic dimension of $\mathcal{S}_1\times \mathcal{S}_2$, and $V_d(\bm{z}) = \big(\bm{v}_{d+1}(\bm{z}),..., \bm{v}_{D_T}(\bm{z})\big)$ with its columns as unit eigenvectors associated with the $(D_T-d)$ smallest eigenvalues within the tangent space $T_{\bm{z}}$. When $\mathcal{S}_1=\Omega_{D_1}$, $(\bm{x},\bm{0})$ is an unit eigenvector of $\mathcal{H} f(\bm{z})$ associated with eigenvalue 0 and orthogonal to $T_{\bm{z}}$. The same applies to $(\bm{0},\bm{y})$ when $\mathcal{S}_2=\Omega_{D_2}$. 

As in \autoref{subsec:mode_est}, our KDE $\hat{f}_{\bm{h}}$ in \eqref{KDE_prod} provides a natural estimator of $\mathcal{R}_d$ as:
\begin{align}
\label{ProdDen_Ridge_est}
\begin{split}
	\hat{\mathcal{R}}_d =& \big\{\bm{z} \in \mathcal{S}_1\times \mathcal{S}_2: \hat{V}_d(\bm{z}) \hat{V}_d(\bm{z})^{\top} \nabla \hat{f}_h(\bm{z}) =\bm{0}, \hat{\lambda}_{d+1}(\bm{z}) < 0 \big\},
\end{split}
\end{align}
where $\hat{V}_d(\bm{z})$ and $\hat{\lambda}_{d+1}(\bm{z})$ are the counterparts of $V_d(\bm{z})$ and $\lambda_{d+1}(\bm{z})$ defined by $\mathcal{H}\hat{f}_{\bm{h}}(\bm{z})$. Under some regularity conditions, $\hat{\mathcal{R}}_d$ is a consistent estimator of $\mathcal{R}_d$; see \autoref{Sec:Ridge_cons} of the supplement for detailed discussions.

\subsection{SCMS Algorithm on $\mathcal{S}_1\times \mathcal{S}_2$: Pitfall and Solution}

% $\bullet$ {\bf SCMS Algorithm on $\mathcal{S}_1\times \mathcal{S}_2$}. 

To identify the estimated density ridge $\hat{\mathcal{R}}_d$ in practice, the SCMS algorithm is one of the most effective off-the-shelf methods. However, generalizing the mean shift algorithm in \autoref{Sec:MS_prod} with the standard technique in \cite{Principal_curve_surf2011,SCMS_conv2013} to the product space $\mathcal{S}_1\times \mathcal{S}_2$ results in an incorrect estimated ridge.

$\bullet$ {\bf Pitfall}. Naively, one may adapt the standard SCMS iteration by updating the sequence $\left\{\bm{z}^{(t)} \right\}_{t=0}^{\infty} \subset \mathcal{S}_1\times \mathcal{S}_2$ via the mean shift vector \eqref{MS_vec} as:
\begin{equation}
\label{Naive_SCMS}
\bm{z}^{(t+1)} \gets \bm{z}^{(t)} + \hat{V}_d(\bm{z}^{(t)}) \hat{V}_d(\bm{z}^{(t)})^{\top} \Xi(\bm{z}^{(t)})
\end{equation}
with an additional standardization whenever $\mathcal{S}_1$ and/or $\mathcal{S}_2$ are directional for $t=0,1,...$. Unfortunately, this naive SCMS iteration on $\mathcal{S}_1\times \mathcal{S}_2$ does not converge to the correct estimated ridge $\hat{\mathcal{R}}_d$ but a transformed estimated ridge $\hat{\mathcal{R}}_d^{(T)}$ defined as:
\begin{align}
\label{Biased_est_ridge}
\begin{split}
	\hat{\mathcal{R}}_d^{(T)} =& \big\{\bm{z} \in \mathcal{S}_1\times \mathcal{S}_2: \hat{V}_d(\bm{z}) \hat{V}_d(\bm{z})^{\top} \tilde{D}(\bm{z})^{-1} \nabla \hat{f}_{\bm{h}}(\bm{z}) =\bm{0}, \hat{\lambda}_{d+1}(\bm{z}) < 0 \big\},
\end{split}
\end{align}
where $\tilde{D}(\bm{z})$ is defined in \eqref{trans_tot_grad_mat}. This discrepancy arises because the mean shift vector \eqref{MS_vec} does not align with the total gradient estimator \eqref{KDE_tot_grad}, but instead corresponds to the transformed estimated gradient $\tilde{D}(\bm{z})^{-1} \nabla \hat{f}_{\bm{h}}(\bm{z})$. As a result, when the factors $G_{\bm{x}},G_{\bm{y}}$ in $\tilde{D}(\bm{z})$ differ, $\hat{\mathcal{R}}_d \neq \hat{\mathcal{R}}_d^{(T)}$. 
Asymptotically, $\hat{\mathcal{R}}_d^{(T)}$ approaches the biased quantity
\begin{align*}
\mathcal{R}_d^{(T)} &= \big\{\bm{z} \in \mathcal{S}_1\times \mathcal{S}_2: V_d(\bm{z}) V_d(\bm{z})^{\top} \bm{H}\tilde{F}(\bm{z})^{-1} \nabla f(\bm{z}) =\bm{0}, \lambda_{d+1}(\bm{z}) < 0 \big\}
\end{align*}
when $h_1,h_2 \lesssim h\to 0$ and the sample size $n\to \infty$; see the following proposition with its proof in \autoref{Sec:Conv_pf}. Consequently, the naive SCMS algorithm leads to an inconsistent estimator of the true ridge when $\mathcal{R}_d^{(T)} \neq \mathcal{R}_d$.

\begin{proposition}
	\label{Thm:trans_grad_conv}
	Assume conditions (A1-2) in \autoref{Sec:Assum_Cons} of the supplement. Then,
	\begin{align*}
		&\left[\bm{H}\tilde{D}(\bm{z})\right]^{-1} \grad \hat{f}_{\bm{h}}(\bm{z}) -\tilde{F}(\bm{z})^{-1}\grad f(\bm{z}) = O(h^2) + O_P\left(\sqrt{\frac{1}{nh^{D_1+D_2+2}}} \right)
	\end{align*}
	for any fixed $\bm{z}\in \mathcal{S}_1\times \mathcal{S}_2$ with $h_1,h_2\lesssim h \to 0$ and $nh^{D_1+D_2+2} \to \infty$, where $\tilde{F}(\bm{z})$ is a nonrandom function depending on $f(\bm{z})$ and kernels. See the full version in Proposition~\ref{Thm:trans_grad_conv_full}.
\end{proposition}

$\bullet$ {\bf Solution}. 
% When the commonly used Gaussian and/or von Mises kernels are applied, the difference between $G_{\bm{x}}$ and $G_{\bm{y}}$ is mitigated so that the transformed gradient $\tilde{D}(\bm{z}) \nabla \hat{f}_{\bm{h}}(\bm{z})$ reduces to $\bm{H} \nabla \hat{f}_{\bm{h}}(\bm{z}) \propto \Xi(\bm{z})$ with $\bm{H}$ defined in \eqref{KDE_Exp}.
Some simple algebra from \eqref{KDE_Exp} show that $\bm{H}^{-1} \Xi(\bm{z}) = \frac{\nabla\hat{f}_{\bm{h}}(\bm{z})}{\hat{f}_{\bm{h}}(\bm{z})}$
under Gaussian and/or von Mises kernels. Based on this observation,
we propose the SCMS algorithm on $\mathcal{S}_1\times \mathcal{S}_2$ with the following iterative update rule:
\begin{equation}
\label{scaled_SCMS}
\bm{z}^{(t+1)} \gets \bm{z}^{(t)} + \eta\cdot \hat{V}_d(\bm{z}^{(t)}) \hat{V}_d(\bm{z}^{(t)})^{\top} \bm{H}^{-1} \Xi(\bm{z}^{(t)}),
\end{equation}
where $\eta >0$ is the step size, and additional standardization is applied when $\mathcal{S}_1$ and/or $\mathcal{S}_2$ are directional for $t=0,1,\dots$.
Unlike the naive SCMS algorithm \eqref{Naive_SCMS}, this version introduces an extra tuning parameter, the step size $\eta$, which plays a crucial role in the convergence behavior. As discussed in \autoref{Sec:stepsize}, it is inappropriate to use a constant step size, say $\eta=1$, independent of the bandwidth parameters $\bm{h}$. 
Normally, iterating \eqref{scaled_SCMS} with smaller step sizes guarantees convergence but may result in slow rates, whereas larger step sizes can accelerate convergence, but may overshoot the targeted ridge $\hat{\mathcal{R}}_d$. 
As a guideline, we suggest adapting the step size to the bandwidth parameters as:
\begin{equation}
\label{ROT_stepsize}
\eta=\min\{\max(\bm{h}) \cdot \min(\bm{h}), 1 \} = \min\left\{h_1h_2, 1 \right\}
\end{equation}
so that when $h_1,h_2 \lesssim h$ are small, $\eta$ mimics the asymptotic rate $O(h^2)$ of adaptive step sizes in Euclidean/directional (subspace constrained) mean shift algorithms \citep{MS1995,Ery2016,DirSCMS2021}. The upper bound $1$ in \eqref{ROT_stepsize} prevents $\eta$ from being too large, thereby avoiding overshooting the targeted ridge $\hat{\mathcal{R}}_d$. Unlike the mean shift algorithm on $\mathcal{S}_1\times\mathcal{S}_2$ in \autoref{Sec:MS_prod}, we do not formulate the componentwise SCMS algorithm, because the eigenspace projector $\hat{V}_d(\bm{z}) \hat{V}_d(\bm{z})^{\top}$ must be recomputed in each sub-step of its iteration. Given the time complexity $O\left((D_1+D_2)^3 \right)$ for spectral decomposition on $\mathcal{H}\hat{f}_{\bm{h}}(\bm{z})$, the componentwise SCMS approach is less computationally efficient than the approach in \eqref{scaled_SCMS}. Under the step size \eqref{ROT_stepsize}, we establish the following (linear) convergence results for our proposed SCMS algorithm on $\mathcal{S}_1\times \mathcal{S}_2$.
Additionally, simulation studies in \autoref{Sec:stepsize} of the supplement demonstrates the effectiveness of our step size choice in \eqref{ROT_stepsize}.

\begin{theorem}
\label{Thm:SCMS_lin_conv}
Assume that the assumptions in Theorem~\ref{Thm:Ridge_cons} and conditions (A4-5) in \autoref{Sec:Lin_conv_SCMS} of the supplement hold. For a sequence $\left\{\bm{z}^{(t)} \right\}_{t=0}^{\infty}$ from our SCMS algorithm \eqref{scaled_SCMS} with step size $\eta$ dominated by $\max(\bm{h})$, there exist constants $\tilde{r}_2>0, A' \equiv A'(\bm{z}^{(0)}) >0$, and $\Upsilon_2 \in (0,1)$ such that
$$
\max\left\{d_g(\bm{z}^{(t)}, \hat{\mathcal{R}}_d),\, d_g(\bm{z}^{(t)}, \mathcal{R}_d) \right\} \leq A' \cdot \Upsilon_2^t
$$
when $\bm{z}^{(0)} \in \mathcal{R}_d \oplus \tilde{r}_2=\left\{\bm{z}\in \mathcal{S}_1\times \mathcal{S}_2: d_g(\bm{z},\mathcal{R}_d) \leq \tilde{r}_2 \right\}$, $\max(\bm{h})\lesssim h$ is sufficiently small, and the sample size $n$ is sufficiently large. Here, $d_g(\bm{z},\mathcal{R}_d) = \inf\left\{d_g(\bm{z},\bm{x}): \bm{x}\in \mathcal{R}_d \right\}$ with $d_g(\cdot,\cdot)$ being the geodesic distance on $\mathcal{S}_1\times \mathcal{S}_2$. See the full version in Theorem~\ref{Thm:SCMS_lin_conv_full}.
\end{theorem}

The core of proving Theorem~\ref{Thm:SCMS_lin_conv} is to show that our proposed SCMS algorithm \eqref{scaled_SCMS} aligns with the general subspace constrained gradient ascent framework on $\mathcal{S}_1\times \mathcal{S}_2$, with the intrinsic step size related to the bandwidths $h_1,h_2$; see \autoref{Sec:Lin_conv} of the supplement for details. Theorem~\ref{Thm:SCMS_lin_conv} indicates that our proposed rule of thumb for the step size $\eta$ ensures the algorithm's linear convergence when $h_1,h_2$ are small. 
Additionally, it implies that the time complexity of our proposed SCMS algorithm \eqref{scaled_SCMS}, applied to a dataset with size $n$ on $\mathcal{S}_1\times \mathcal{S}_2$, is $O\left(n\cdot (D_1+D_2)^3 \cdot \log\left(\frac{1}{\epsilon}\right) \right)$ for achieving an $\epsilon$-error.

\section{Experiments}
\label{Sec:Experiments}

In this section, we present two real-world applications of our proposed mode-seeking and ridge-finding methods with directional-linear and directional-directional data. The detailed setups of our experiments and simulation results can be found in \autoref{Sec:sim_setup}.

\subsection{Application I: Local Mode Estimation on Earthquake Data}
\label{subsec:local_mode_earthquake}

It is well-known in seismology \citep{schuster1897lunar,tavares2011influences} that solar and lunar periodicities influence earthquake occurrences. To showcase an application of our proposed mean shift algorithm in product spaces, we aim to jointly identify the times and locations where earthquakes occur more intensively. We analyze global earthquakes of magnitude 6 or higher from 2002-07-01 00:00:00 UTC to 2022-06-30 23:59:59 UTC through the Earthquake Catalog (\url{https://earthquake.usgs.gov/earthquakes/search/}) of the United States Geological Survey. This dataset has 3,066 events in total. The earthquake locations (longitudes and latitudes) are mapped to points $\left\{\bm{X}_i \right\}_{i=1}^{7602}$ on the unit sphere $\Omega_2$ with their occurrence times denoted by timestamps $\left\{t_i\right\}_{i=1}^{7602}$. The resulting dataset $\left\{(\bm{X}_i,t_i)\right\}_{i=1}^{7602} \subset \Omega_2 \times \mathbb{R}$ is directional-linear (or spatio-temporal). We determine bandwidth parameters $h_1 \approx 0.223$ (directional) and $h_2\approx 38062140.811$ (linear) via \eqref{bw_ROT_Dir} and \eqref{NR_rule} in \autoref{Sec:Add_exp_bw}, respectively. The initial points for the (simultaneous) mean shift algorithm are chosen as the original dataset. We implement the algorithm on the directional-linear product space $\Omega_2 \times \mathbb{R}$ using parallel programming in the Ray environment \citep{moritz2018ray}. The convergence criteria are set to $\norm{\left(\bm{X}_i^{(t+1)},t_i^{(t+1)} \right) -\left(\bm{X}_i^{(t)},t_i^{(t)} \right)}_2 \leq 10^{-7}$ or a maximum of $5,000$ iterations. Those non-convergent points after $5,000$ iterations are removed from the final set of estimated local modes.

\begin{figure*}[t]
	\captionsetup[subfigure]{justification=centering}
	\centering
	\begin{subfigure}[t]{.49\linewidth}
		\centering
		\includegraphics[width=1\linewidth]{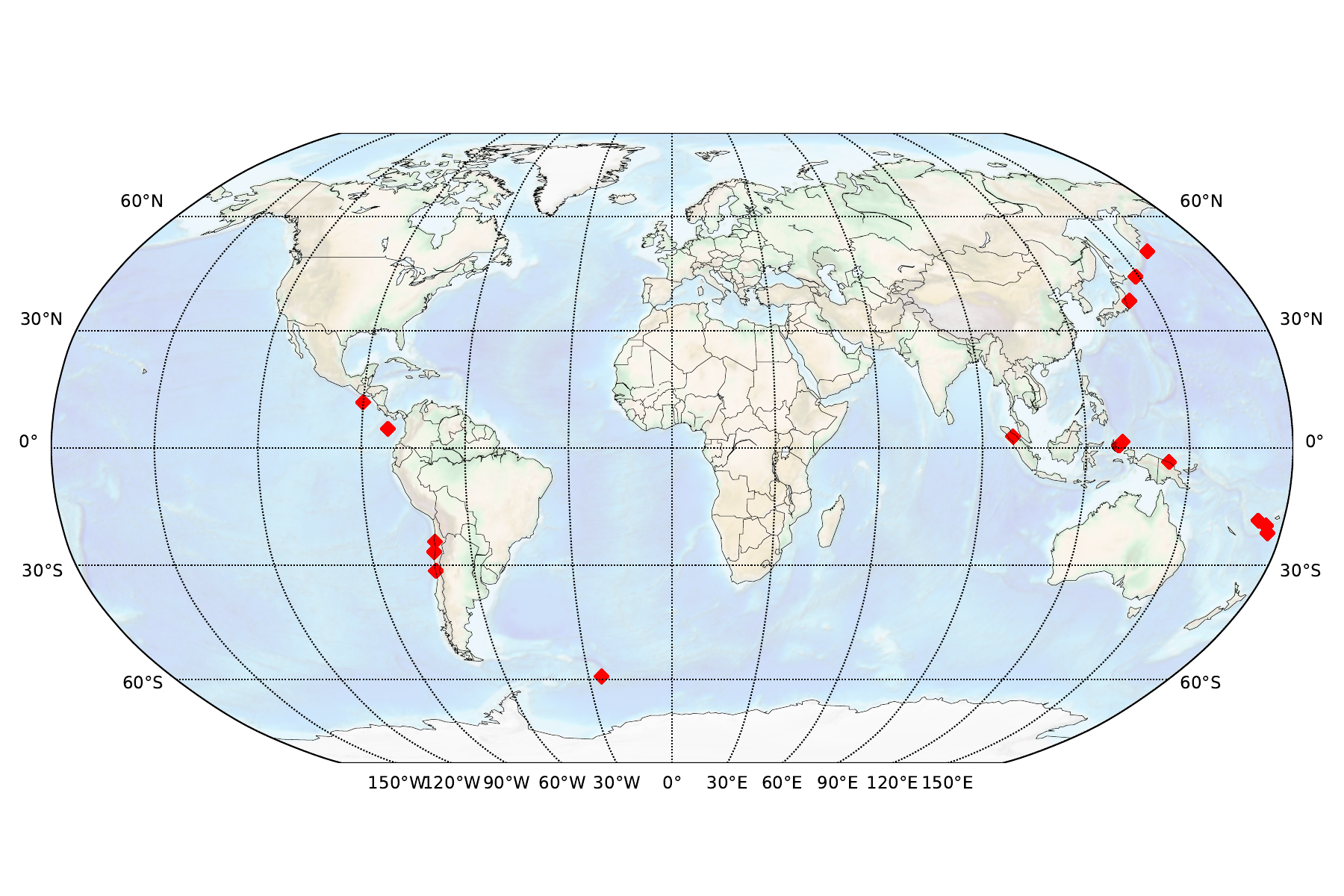}
		\caption{Earthquake modes projected to their directional/spatial components on $\Omega_2$.}
	\end{subfigure}
	\hfil
	\begin{subfigure}[t]{.49\linewidth}
		\centering
		\includegraphics[width=1\linewidth]{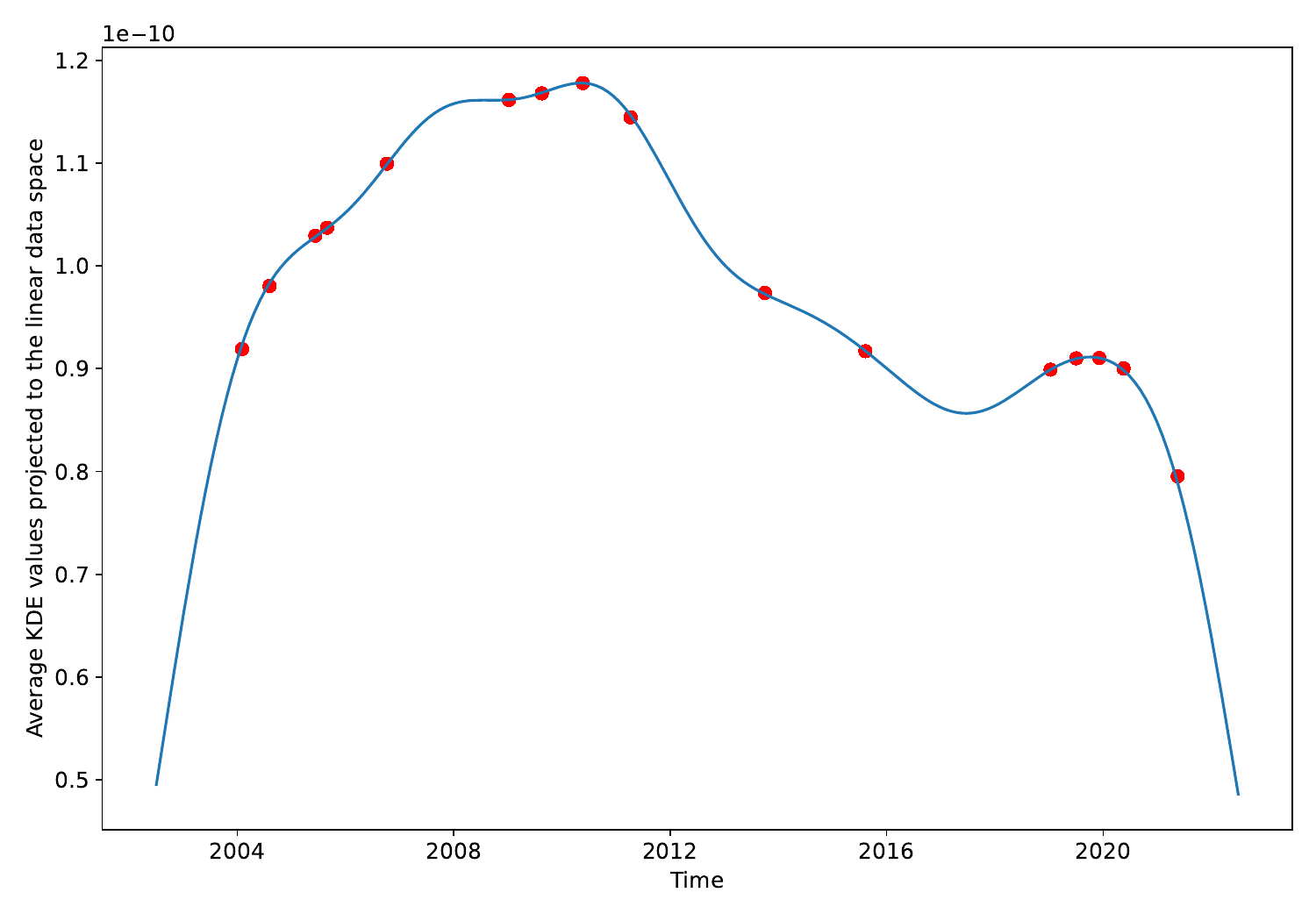}
		\caption{Earthquake modes projected to their linear/temporal components (lifted onto the scaled marginal density) in $\mathbb{R}$.}
	\end{subfigure}
	\caption{Estimated local modes from our proposed mean shift algorithm applied to the earthquake data. In each panel, red dots denote the estimated earthquake modes. In (b), the density estimates are obtained by averaging the directional-linear KDE over the directional (spatial) component and projecting the results onto the linear (temporal) axis. This density plot is the estimated (scaled) marginal density with respect to time $t$.}
	\label{fig:ModeSeek_EQ}
\end{figure*}

The estimated local modes from our proposed mean shift algorithm on $\Omega_2 \times \mathbb{R}$ are shown in \autoref{fig:ModeSeek_EQ}. When projected onto the directional (location) space and linear (time) space in Panels (a) and (b), respectively, the resulting modes exhibit greater spatial and temporal variability than those of the marginal estimated density due to the multimodal nature of the directional-linear density on $\Omega_2 \times \mathbb{R}$. Notably, the algorithm successfully identifies regions with high seismic activities, such as the West Coast of South America and areas near Indonesia in Panel (a) of \autoref{fig:ModeSeek_EQ}. Furthermore, except for the boundary effects, the estimated temporal density of earthquakes in Panel (b) is relatively flat, indicating the uniformity and unpredictability associated with the occurrence of large earthquakes. Finally, we also applied the standard Euclidean mean shift algorithm on this dataset. However, the resulting estimated local modes were scattered across the entire world map, failing to provide any meaningful insights.

% some patterns that earthquakes occur more frequently in the first half of each month than in the second half. It would be interesting to investigate whether this temporal pattern is linked to solar or lunar cycles in the future. Additionally, converting earthquake timestamps to solar or lunar calendars and applying our proposed mean shift algorithm could yield local modes with more interpretable connections to periodic celestial influences.

\subsection{Application II: Cosmic Filament Detection}
\label{subsec:cosmic_web}

Previous astronomical studies have shown that, on megaparsec scales, matter in the Universe forms a complicated large-scale network structure called the cosmic web \citep{de1986slice,springel2006large,cautun2014evolution}. Among its major characteristics, filaments are of particular interest due to their connections to the nature of dark matter \citep{zhang2009spin} and correlations with stellar properties of nearby galaxies \citep{zhang2013alignments,clampitt2016detection}. Here, we demonstrate the application of our proposed SCMS algorithm to detect cosmic filaments from astronomical survey data. 

For an illustrative purpose, we focus on a subset of galaxies in the Data Release 16 (DR16) of the Sloan Digital Sky Survey (SDSS-IV; \citealt{ahumada202016th}) with redshift value $0.05\leq z < 0.07$. Specifically, we obtain the data from the FIREFLY value-added catalog \citep{comparat2017stellar,wilkinson2017firefly} (\url{https://www.sdss.org/dr17/spectro/eboss-firefly-value-added-catalog}) and only incorporate galaxies with reliable and positive definite redshift values. Each galaxy in the dataset has its coordinate as $(\xi_i,\phi_i,z_i)$, where $\xi_i$ is its right ascension (\texttt{RA}), $\phi_i$ is its declination (\texttt{DEC}), and $z_i$ is its redshift value (\texttt{Z\_NOQSO}). We focus on the North Galactic Cap region ($100 < \mathtt{RA} <270$ and $-5< \mathtt{DEC} < 70$), yielding 97,435 galaxies with redshift value $0.05\leq z < 0.07$. To focus on denser regions, we retain 80\% of the galaxies with the highest density estimates, resulting in a sample size as $n=77,948$.

\begin{figure*}[t]
	\captionsetup[subfigure]{justification=centering}
	\centering
	\begin{subfigure}[t]{.49\linewidth}
		\centering
		\includegraphics[width=0.8\linewidth]{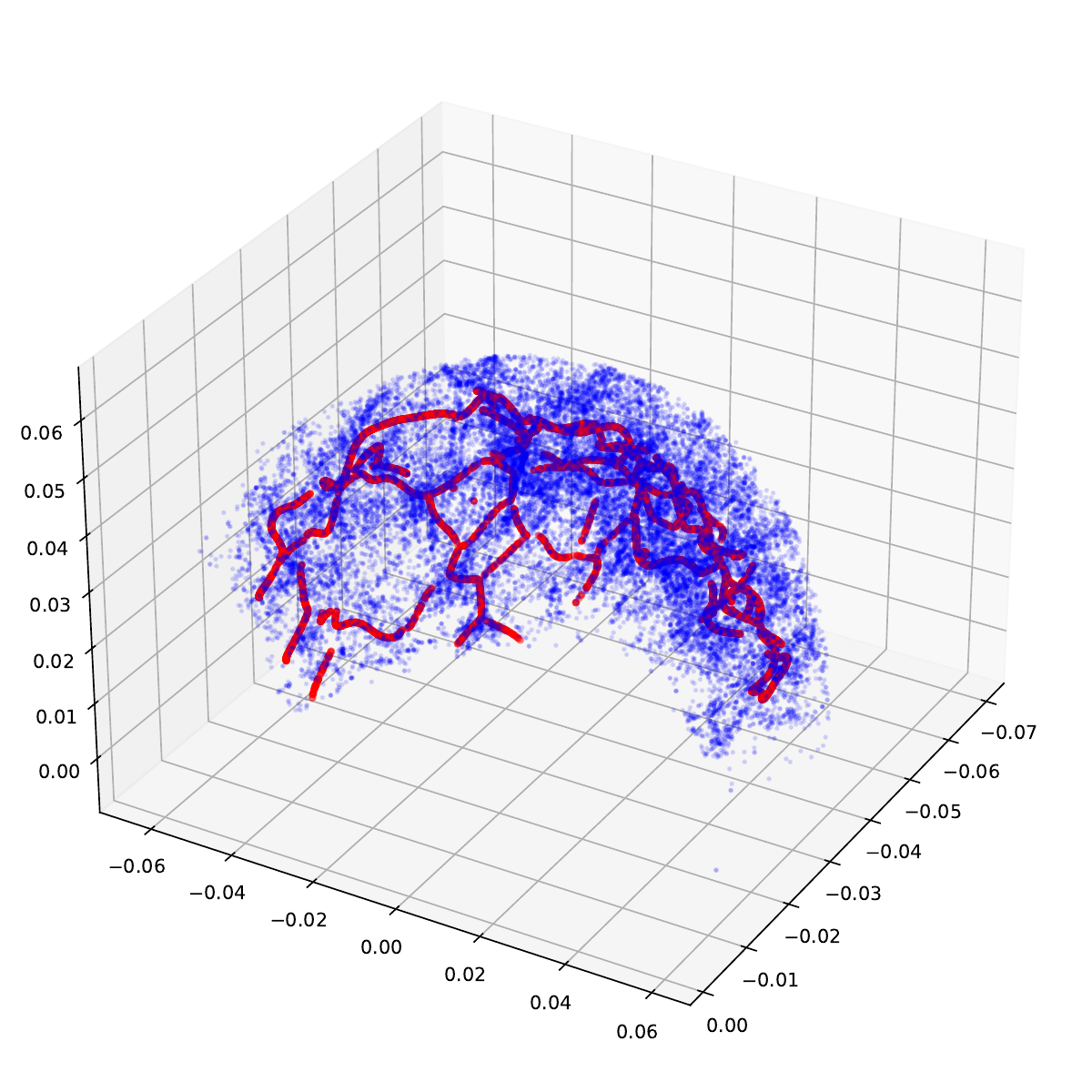}
		\caption{Observed galaxies and detected filaments under the Cartesian coordinate system in $\mathbb{R}^3$.}
	\end{subfigure}
	\hfil
	\begin{subfigure}[t]{.49\linewidth}
		\centering
		\includegraphics[width=0.9\linewidth]{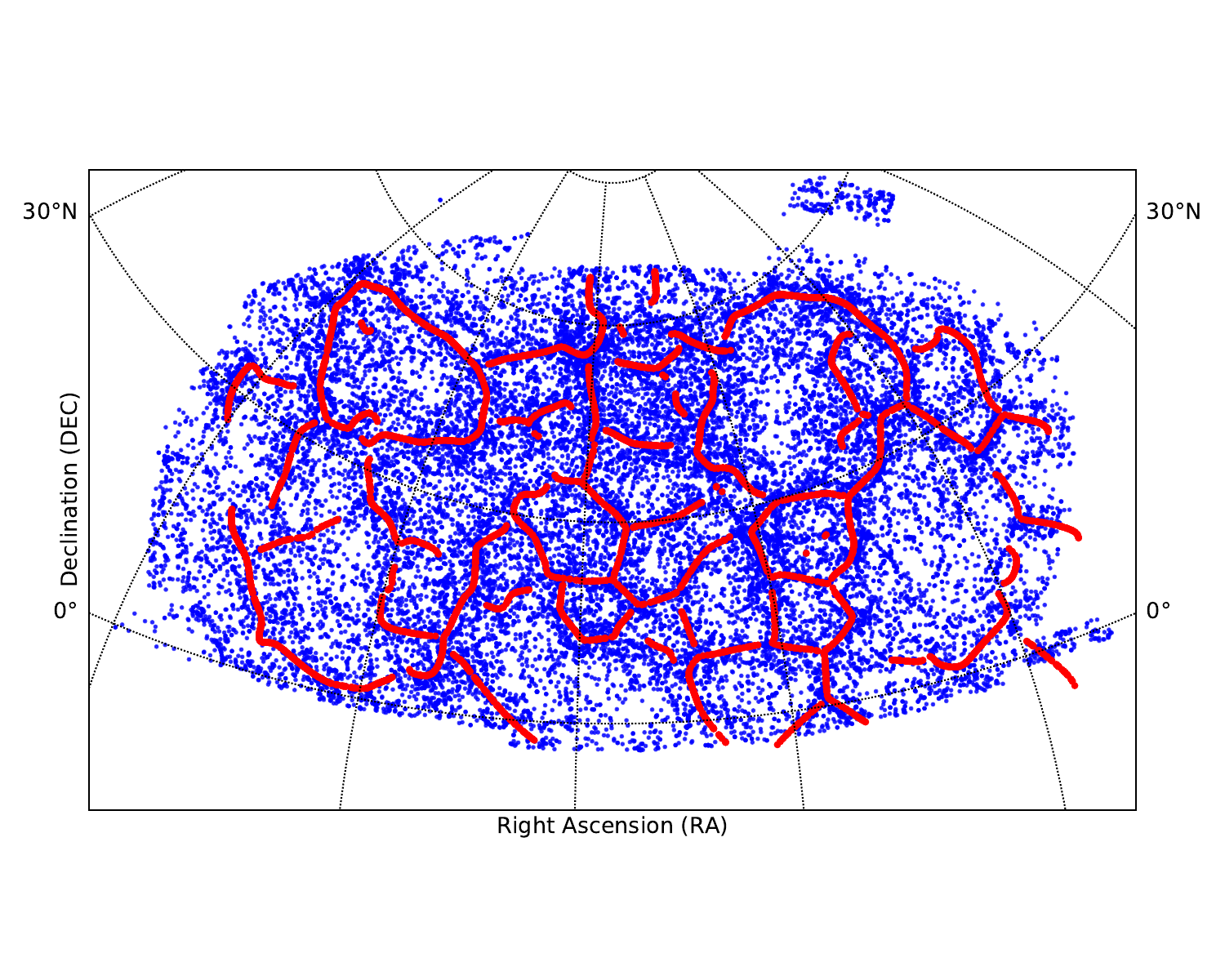}
		\caption{Observed galaxies and detected filaments projected onto the (RA, DEC) space on $\Omega_2$.}
	\end{subfigure}
	\caption{Visualization of a subset of galaxies (blue dots) and filament points (red dots) detected by our proposed SCMS algorithm. For clarity, we randomly sample 30,000 galaxies from the original dataset and plot them in Panels (a) and (b).}
	\label{fig:cosmic_fila}
\end{figure*}

Traditionally, the regular/Euclidean SCMS or other filament detection algorithms are employed in a 3D Cartesian space, where galaxy coordinates are given by $(X_i,Y_i,Z_i)=\left( g(z_i) \cos\xi_i \cos \phi_i,\, g(z_i) \sin\xi_i \cos \phi_i,\, g(z_i) \sin\phi_i \right)$
and $g(\cdot)$ is a distance transforming function \citep{tempel2014detecting}. To circumvent challenges in selecting $g(\cdot)$, we estimate the filamentary structure directly in the original directional-linear space with our proposed SCMS algorithm under the data representation $\left\{(\bm{X}_i,z_i) \right\}_{i=1}^n \subset \Omega_2\times \mathbb{R}$, where $\bm{X}_i$ is the Cartesian coordinate of $(\xi_i,\phi_i)$ for $i=1,...,n$.
Bandwidths for the directional and linear components are determined via the rule of thumb \eqref{bw_ROT_Dir} and normal reference rule \eqref{NR_rule}, calculated on a larger galaxy sample on the North Galactic Cap with redshift value $0\leq z<2$ to ensure stability. This yields $h_1\approx 0.0376$ for the directional component and $h_2\approx 0.0187$ for the linear component. 
% Computing the bandwidth parameters on a larger galaxy sample helps stabilize the detected filament structure.
Our proposed SCMS algorithm is again implemented using parallel programming in the Ray environment, with a tolerance level of $10^{-7}$ and a maximum of $5,000$ iterations. Those non-convergent points after $5,000$ iterations are excluded from the final filament estimates.

\autoref{fig:cosmic_fila} presents the observed galaxies and detected cosmic filament points in 3D Cartesian and 2D (RA, DEC) spaces. Notably, the filament structures identified by our proposed SCMS algorithm trace regions of high galaxy density, highlighting the algorithm;s capability. In the future work, we plan to apply our method to the entire SDSS-IV galaxy data and examine the relationships between the detected filaments and the stellar properties of observed galaxies; see also \cite{zhang2022sconce}.

\section{Conclusion and Future Work}

In this paper, we generalized the (subspace constrained) mean shift algorithms to Euclidean/directional product spaces, established their (linear) convergence properties, and provided practical guidelines for implementation. The utility of our proposed methods was demonstrated on both DirLin and DirDir datasets. In the future, we aim to extend the mean shift and SCMS algorithms to a product space with an infinite number of factors, such as some functional spaces, with the kernelized mean shift algorithm \citep{anand2013semi} serving as a potential starting point. We also plan to study mode-seeking and ridge-finding problems on product spaces comprising more general topological structures, such as matrix manifolds \citep{sasakimode2022} or more complicated (Riemannian) manifolds \citep{Prod_Manifold2021}. 
Additionally, developing data-adaptive bandwidth selection methods for these algorithms remains an important avenue for future research.

\section*{Acknowledgments}

We thank the Editor, Associate Editor, and three reviewers for their constructive comments.

Funding for the Sloan Digital Sky 
Survey IV has been provided by the 
Alfred P. Sloan Foundation, the U.S. 
Department of Energy Office of 
Science, and the Participating 
Institutions. SDSS-IV acknowledges support and 
resources from the Center for High 
Performance Computing  at the 
University of Utah. The SDSS 
website is \url{www.sdss.org}. 
SDSS-IV is managed by the 
Astrophysical Research Consortium 
for the Participating Institutions 
of the SDSS Collaboration including 
the Brazilian Participation Group, 
the Carnegie Institution for Science, 
Carnegie Mellon University, Center for 
Astrophysics | Harvard \& 
Smithsonian, the Chilean Participation 
Group, the French Participation Group, 
Instituto de Astrof\'isica de 
Canarias, The Johns Hopkins 
University, Kavli Institute for the 
Physics and Mathematics of the 
Universe (IPMU) / University of 
Tokyo, the Korean Participation Group, 
Lawrence Berkeley National Laboratory, 
Leibniz Institut f\"ur Astrophysik 
Potsdam (AIP),  Max-Planck-Institut 
f\"ur Astronomie (MPIA Heidelberg), 
Max-Planck-Institut f\"ur 
Astrophysik (MPA Garching), 
Max-Planck-Institut f\"ur 
Extraterrestrische Physik (MPE), 
National Astronomical Observatories of 
China, New Mexico State University, 
New York University, University of 
Notre Dame, Observat\'ario 
Nacional / MCTI, The Ohio State 
University, Pennsylvania State 
University, Shanghai 
Astronomical Observatory, United 
Kingdom Participation Group, 
Universidad Nacional Aut\'onoma 
de M\'exico, University of Arizona, 
University of Colorado Boulder, 
University of Oxford, University of 
Portsmouth, University of Utah, 
University of Virginia, University 
of Washington, University of 
Wisconsin, Vanderbilt University, 
and Yale University.

The authors report there are no competing interests to declare.

\bigskip
\begin{center}
{\large\bf SUPPLEMENTARY MATERIAL}
\end{center}

\begin{description}
	
	\item[Title:] Supplement to ``Mode and Ridge Estimation in Euclidean and Directional Product Spaces: A Mean Shift Approach'' (.pdf file)
	
	The supplementary materials contain some auxiliary results, assumptions, and proofs of theorems in the main paper.
	
	\item[Python code:] Python code for reproducing our experiments can be found at the GitHub repository (\url{https://github.com/zhangyk8/ProdSCMS}).
\end{description}

\bibliography{Bib_DLSCMS}

\newpage

	\def\spacingset#1{\renewcommand{\baselinestretch}%
		{#1}\small\normalsize} \spacingset{1}

	%%%%%%%%%%%%%%%%%%%%%%%%%%%%%%%%%%%%%%%%%%%%%%%%%%%%%%%%%%%%%%%%%%%%%%%%%%%%%%
	
	\if0\blind
	{
		\title{\bf Supplement to ``Mode and Ridge Estimation in Euclidean and Directional Product Spaces: A Mean Shift Approach''}
		\author{Yikun Zhang\thanks{
				Email: yikun@uw.edu.}\hspace{.2cm}\\
			Department of Statistics, University of Washington\\
			and \\
			Yen-Chi Chen\thanks{Email: yenchic@uw.edu.} \\
			Department of Statistics, University of Washington}
		\date{}
		\maketitle
	} \fi
	
	\if1\blind
	{
		\bigskip
		\bigskip
		\begin{center}
			{\LARGE\bf Supplement to ``Mode and Ridge Estimation in Euclidean and Directional Product Spaces: A Mean Shift Approach''}
		\end{center}
		\medskip
	} \fi
	
\end{comment}

\singlespacing

\begin{center}
	{\LARGE\bf Supplement to ``Mode and Ridge Estimation in Euclidean and Directional Product Spaces: A Mean Shift Approach''}
\end{center}

\setcounter{page}{1}

\appendix

The appendices contain some auxiliary assumptions, results, and proofs of theorems in the main paper, whose outline is stated as follows.
\begin{itemize}
	\item {\bf \autoref{Sec:sim_setup}: Experimental Setup and Additional Results.} We delineate the choices of tuning parameters in our experiments and provide additional results for our simulation studies.
	
	\item {\bf \autoref{Sec:Cons_Theory}: Statistical Consistency.}
	We state our assumptions and the statistical convergence rates of the KDE on $\mathcal{S}_1\times \mathcal{S}_2$ as well as the associated mode and ridge estimators.
	
	\item {\bf \autoref{Sec:Lin_conv}: Linear Convergence of the Mean Shift and SCMS Algorithms on $\mathcal{S}_1\times \mathcal{S}_2$.}
	We formulate the (subspace constrained) gradient ascent framework on a general product manifold. Further, we argue how the proposed mean shift and SCMS algorithms fit into this framework and consequently, obtain their linear convergence properties. 
	In addition, we present some empirical evidence from simulation studies that our suggested rule of thumb for the step size parameter $\eta$ in the proposed SCMS algorithm is effective.
	
	\item {\bf \autoref{Sec:Conv_pf}: Proofs of Theorem~\ref{Thm:MS_conv} and Proposition~\ref{Thm:trans_grad_conv}.}
	We provide the proofs of Theorem~\ref{Thm:MS_conv} and Proposition~\ref{Thm:trans_grad_conv} in the main paper. 
	
	\item {\bf \autoref{Sec:auxi_pf} Proof of Proposition~\ref{prop:bw_LSCV}.} We provide the proof of Proposition~\ref{prop:bw_LSCV} in the supplement.
\end{itemize}

\section{Experimental Setup and Additional Results}
\label{Sec:sim_setup}

Unless otherwise specified, the initial set of mesh points for each experiment is the original dataset, and the stopping criterion is set to a tolerance level of $10^{-7}$. We implement our SCMS algorithm using the logarithm of the KDE, as it offers faster convergence speeds \citep{SCMS_conv2013}.

\subsection{Bandwidth Selection}
\label{Sec:Add_exp_bw}

As studying the choices of bandwidth parameters for the directional and linear data is not the main focus of this paper, we mainly leverage some existing rule-of-thumb bandwidth selectors in the literature. Without loss of generality, we assume that the dataset $\left\{\bm{Z}_i \right\}_{i=1}^n=\left\{(\bm{X}_i,\bm{Y}_i) \right\}_{i=1}^n \subset \Omega_{D_1}\times \mathbb{R}^{D_2}$ is directional-linear. For the bandwidth parameter in the directional part $\left\{\bm{X}_i\right\}_{i=1}^n$ on $\Omega_{D_1}$, we utilize the rule of thumb in Proposition 2 of \cite{Exact_Risk_bw2013}, where the estimated concentration parameter is given by (4.4) in \cite{spherical_EM}. That is,
\begin{align}
	\label{bw_ROT_Dir}
	\begin{split}
		h_{\text{ROT}} = \left[\frac{4\pi^{\frac{1}{2}} \mathcal{I}_{\frac{D_1-1}{2}}(\hat{\kappa})^2}{\hat{\kappa}^{\frac{D_1+1}{2}}\left[2 D_1\cdot\mathcal{I}_{\frac{D_1+1}{2}}(2\hat{\kappa}) + (D_1+2)\hat{\kappa} \cdot \mathcal{I}_{\frac{D_1+3}{2}}(2\hat{\kappa}) \right]n} \right]^{\frac{1}{D_1+4}},
	\end{split}
\end{align}
with $\hat{\kappa} = \frac{\bar{R}(D_1+1-\bar{R})}{1-\bar{R}^2}$, where $\bar{R}=\frac{\norm{\sum_{i=1}^n \bm{X}_i}_2}{n}$ given the directional dataset $\left\{\bm{X}_i\right\}_{i=1}^n \subset \Omega_{D_1} \subset \mathbb{R}^{D_1+1}$ and $\mathcal{I}_{\alpha}(\kappa)$ is the modified Bessel function of the first kind of order $\kappa$. For the bandwidth parameter in the linear part $\left\{\bm{Y}_i\right\}_{i=1}^n$ on $\mathbb{R}^{D_2}$, we adopt the normal reference rule from Eq. (17) in \cite{Mode_clu2016} as:
\begin{equation}
	\label{NR_rule}
	h_{\text{NR}} = \bar{S}_n \times \left(\frac{4}{D_2+4} \right)^{\frac{1}{D_2+6}} n^{-\frac{1}{D_2+6}}
\end{equation}
with $\bar{S}_n=\frac{1}{D_2}\sum_{j=1}^{D_2} S_{n,j}$, where $S_{n,j}$ is the sample standard deviation along the $j$-th coordinates of $\left\{\bm{Y}_i\right\}_{i=1}^n$.

Another method for selecting the bandwidth $\bm{h}$ is through the Least Squares Cross-Validation (LSCV) technique. This approach aims to select a data-adaptive bandwidth by minimizing the mean integrated square error (MISE) $\int_{\mathcal{S}_1\times \mathcal{S}_2} \left[\hat{f}_{\bm{h}}(\bm{x},\bm{y}) - f(\bm{x},\bm{y})\right]^2 \omega(d\bm{x},d\bm{y})$. Practically, this is achieved through the following procedure:
\begin{equation}
\label{bw_LSCV}
\bm{h}_{\text{LSCV}} = \argmin_{\bm{h}} \text{LSCV}(\bm{h}) := \argmin_{\bm{h}} \int_{\mathcal{S}_1\times \mathcal{S}_2} \hat{f}_{\bm{h}}(\bm{x},\bm{y})^2 \omega(d\bm{x},d\bm{y}) - \frac{2}{n} \sum_{i=1}^n \hat{f}_{\bm{h},-i}(\bm{X}_i,\bm{Y}_i),
\end{equation}
where $\hat{f}_{\bm{h},-i}(\bm{x},\bm{y}) = \frac{1}{n-1} \sum_{j=1,j\neq i}^n K_1\left(\frac{\bm{x}-\bm{X}_j}{h_1} \right) K_2\left(\frac{\bm{y}-\bm{Y}_j}{h_2} \right)$ is the leave-one-out KDE of \eqref{KDE_prod}. Under von Mises and/or Gaussian kernels, we derive a closed-form expression of \eqref{bw_LSCV} in the following proposition, whose proof is in \autoref{Sec:auxi_pf}. In our published code and subsequent simulation studies, we also implemented the LSCV bandwidth selection method for our proposed algorithms. However, its performance was inferior to the rule-of-thumb bandwidths \eqref{bw_ROT_Dir} and \eqref{NR_rule}. Consequently, we opted not to report the results obtained using LSCV bandwidths.

\begin{proposition}[Explicit LSCV loss under von Mises and/or Gaussian kernels]
\label{prop:bw_LSCV}
Let $\hat{f}_{\bm{h}}(\bm{x},\bm{y})$ in \eqref{KDE_prod} be defined on the product space $\mathcal{S}_1\times \mathcal{S}_2$ with $\bm{h}=(h_1,h_2)$. If the von Mises kernel profile $L(r)=e^{-r}$ is used for directional components and Gaussian kernel profile $k(s) = e^{-s/2}$ is applied to Euclidean components, then we have the following results.

(a) When $\mathcal{S}_1\times \mathcal{S}_2 = \mathbb{R}^{D_1}\times \mathbb{R}^{D_2}$,
\begin{align*}
	\mathrm{LSCV}(\bm{h}) &= \frac{1}{2^{D_1+D_2} \pi^{\frac{D_1+D_2}{2}} n h^{D_1}h^{D_2}}\left[1+ \frac{2}{n}\sum\limits_{i=1}^{n-1} \sum\limits_{j=1,j>i}^n \exp\left(-\frac{\norm{\bm{X}_i-\bm{X}_j}_2^2}{4h_1^2} - \frac{\norm{\bm{Y}_i-\bm{Y}_j}_2^2}{4h_2^2}\right) \right] \\
	&\quad - \frac{4}{(2\pi)^{\frac{D_1+D_2}{2}} n(n-1)h_1^{D_1} h_2^{D_2}}\sum\limits_{i=1}^{n-1} \sum\limits_{j=1,j>i}^n \exp\left(-\frac{\norm{\bm{X}_i-\bm{X}_j}_2^2}{2h_1^2} - \frac{\norm{\bm{Y}_i-\bm{Y}_j}_2^2}{2h_2^2}\right).
\end{align*}

(b) When $\mathcal{S}_1\times \mathcal{S}_2 = \Omega_{D_1}\times \Omega_{D_2}$,
\begin{align*}
	\mathrm{LSCV}(\bm{h}) &= \frac{C_{D_1}\left(\frac{1}{h_1^2}\right)^2\cdot C_{D_2}\left(\frac{1}{h_2^2}\right)^2}{n\cdot  C_{D_1}\left(\frac{2}{h_1^2}\right)\cdot C_{D_2}\left(\frac{2}{h_2^2}\right)} + \frac{2}{n^2}\sum\limits_{i=1}^{n-1} \sum\limits_{j=1,j>i}^n \frac{C_{D_1}\left(\frac{1}{h_1^2}\right)^2\cdot C_{D_2}\left(\frac{1}{h_2^2}\right)^2}{C_{D_1}\left(\frac{\norm{\bm{X}_i + \bm{X}_j}_2}{h_1^2}\right)\cdot C_{D_2}\left(\frac{\norm{\bm{Y}_i + \bm{Y}_j}_2}{h_2^2}\right)} \\
	&\quad - \frac{4 C_{D_1}\left(\frac{1}{h_1^2}\right)\cdot C_{D_2}\left(\frac{1}{h_2^2}\right)}{n(n-1)} \sum\limits_{i=1}^{n-1} \sum\limits_{j=1,j>i}^n \exp\left(\frac{\bm{X}_i^T\bm{X}_j}{h_1^2} + \frac{\bm{Y}_i^T\bm{Y}_i}{h_2^2}\right),
\end{align*}
where $C_q(\kappa)>0$ is the normalizing constant of the vMF distribution.

(c) When $\mathcal{S}_1\times \mathcal{S}_2 = \Omega_{D_1}\times \mathbb{R}^{D_2}$,
\begin{align*}
	\mathrm{LSCV}(\bm{h}) &= \frac{C_{D_1}\left(\frac{1}{h_1^2}\right)^2}{2^{D_2} \pi^{\frac{D_2}{2}} n \cdot C_{D_1}\left(\frac{2}{h_1^2}\right)\cdot h^{D_2}} + \frac{C_{D_1}\left(\frac{1}{h_1^2}\right)^2}{2^{D_2-1}\pi^{\frac{D_2}{2}}n^2} \sum\limits_{i=1}^{n-1} \sum\limits_{j=1,j>i}^n \frac{\exp\left(-\frac{\norm{\bm{Y}_i-\bm{Y}_j}_2^2}{4h_2^2}\right)}{C_{D_1}\left(\frac{\norm{\bm{X}_i+\bm{X}_j}_2}{h_1^2}\right)}\\
	&\quad - \frac{4C_{D_1}\left(\frac{1}{h_1^2}\right)}{(2\pi)^{\frac{D_2}{2}} n(n-1) h_2^{D_2}}\sum\limits_{i=1}^{n-1} \sum\limits_{j=1,j>i}^n \exp\left(\frac{\bm{X}_i^T\bm{X}_j}{h_1^2} - \frac{\norm{\bm{Y}_i-\bm{Y}_j}_2^2}{2h_2^2}\right).
\end{align*}
\end{proposition}

\begin{remark}
The rule-of-thumb and LSCV bandwidth selection methods described above are primarily designed to minimize the MISE of the KDE \eqref{KDE_prod}. As a result, the resulting bandwidths may not be optimal for mode seeking and ridge estimation tasks \citep{casa2020modal}. Furthermore, when the underlying data-generating distribution is a mixture of directional-linear or directional-directional densities, it may not be appropriate to compute the bandwidth parameters separately on each component of the product space $\mathcal{S}_1\times \mathcal{S}_2$. In such cases, more refined bandwidth selection methods, as proposed in \cite{Exact_Risk_bw2013,portugues2014nonparametric}, are applicable for improved performance.
\end{remark}

\subsection{Manifold-Recovering Error Measure}
\label{Sec:Manifold_Rec_err}

Given noisy observations from a hidden manifold structure $\mathcal{C}$, such as the spiral curve example in \autoref{fig:spiral_cur}, a manifold learning method can learn a collection of discrete data points or a set of solutions to a system of equations that approximate $\mathcal{C}$. For our proposed SCMS algorithm, this set of data points is represented by a discrete sample from the estimated ridge $\hat{\mathcal{R}}_d$. To quantify the estimation error from $\hat{\mathcal{R}}_d$ to the true manifold structure $\mathcal{C}$, one \emph{cannot} simply rely on the (average) distances of points on $\hat{\mathcal{R}}_d$ to $\mathcal{C}$ as:
\begin{equation}
	\label{avg_err_dist}
	d_{\text{avg}}(\hat{\mathcal{R}}_d,\mathcal{C}) = \frac{1}{|\hat{\mathcal{R}}_d|}\sum_{\bm{z}\in \hat{\mathcal{R}}_d} d(\bm{z},\mathcal{C}),
\end{equation}
where $|\hat{\mathcal{R}}_d|$ is the cardinality of (a discrete sample from) $\hat{\mathcal{R}}_d$ and $d(\bm{z},\mathcal{C}) = \inf\left\{d(\bm{z},\bm{c}):\bm{c}\in \mathcal{C} \right\}$ with the distance measure $d(\cdot,\cdot)$ defined in the (ambient) metric space containing $\hat{\mathcal{R}}_d$ and $\mathcal{C}$. This is because the estimated ridge $\hat{\mathcal{R}}_d$ (or any other estimated manifold structures) may only approximate a small portion of the true manifold $\mathcal{C}$ even with uniform (but noisy) observations from $\mathcal{C}$ as inputs; see, for instance, Panel (c) of \autoref{fig:spiral_cur} in the main paper or Panel (c) of \autoref{fig:sph_cone} below. Thus, we borrow the idea from the definition of Hausdorff distances and define the manifold-recovering error measure as:
\begin{equation}
	\label{manifold_rec_err}
	d_H\left(\hat{\mathcal{R}}_d, \mathcal{C} \right) = \frac{1}{2}\left[\frac{1}{|\hat{\mathcal{R}}_d|}\sum_{\bm{z}\in \hat{\mathcal{R}}_d} d(\bm{z},\mathcal{C}) + \frac{1}{|\mathcal{D}^{\mathcal{C}}|} \sum_{\bm{z}\in \mathcal{D}^{\mathcal{C}}} d(\bm{z},\hat{\mathcal{R}}_d) \right],
\end{equation}
where $\mathcal{D}^{\mathcal{C}}$ is a set of uniformly sampling points from the true manifold structure $\mathcal{C}$. This error measure averages over the approximation error and coverage of $\mathcal{C}$ with the estimated ridge/manifold.

\subsection{Simulation 1: Mode-Seeking on $\Omega_1 \times \mathbb{R}$} 

\begin{figure*}[t]
	\captionsetup[subfigure]{justification=centering}
	\centering
	\begin{subfigure}[t]{.32\linewidth}
		\centering
		\includegraphics[width=1\linewidth]{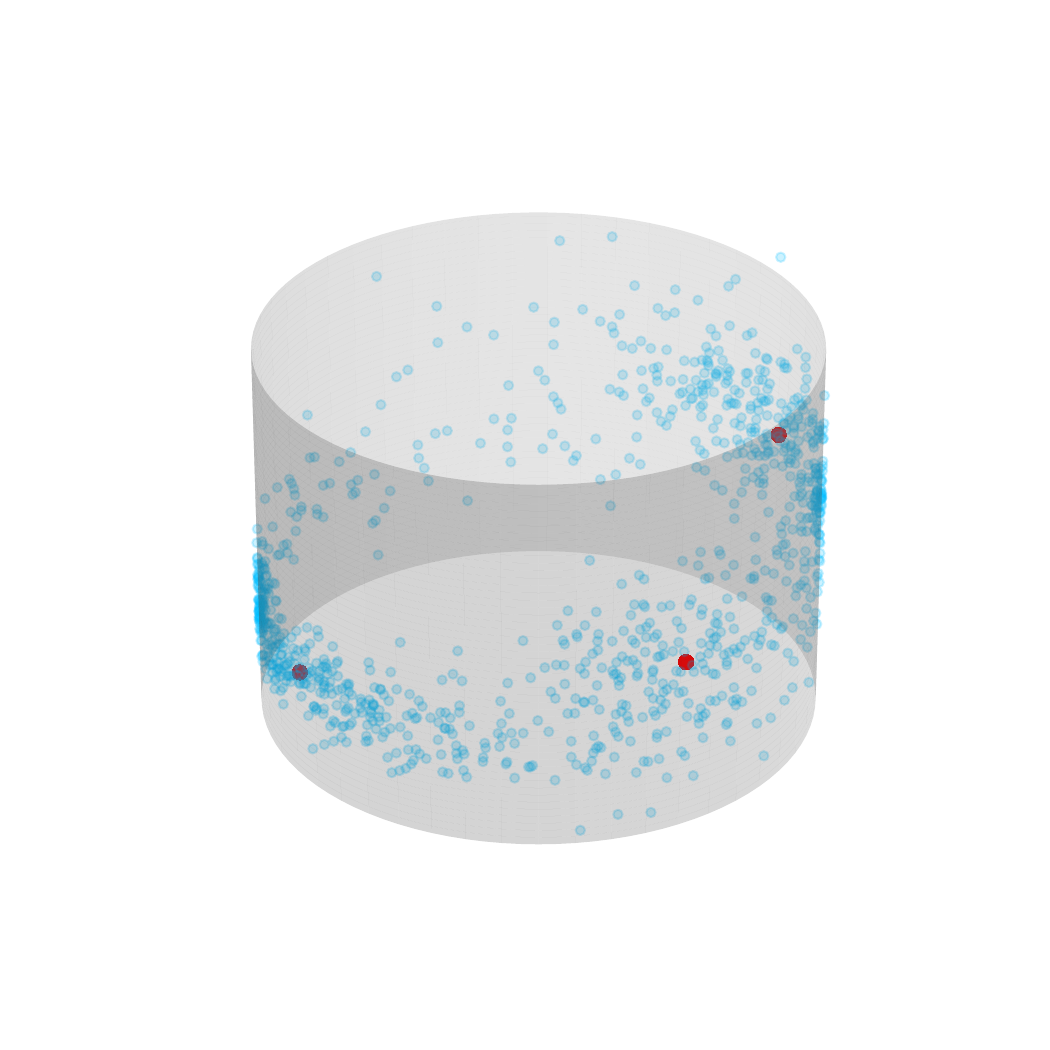}
		\caption{Simulation 1: DirLin data on a cylinder.}
	\end{subfigure}
	\hfil
	\begin{subfigure}[t]{.32\linewidth}
		\centering
		\includegraphics[width=1\linewidth]{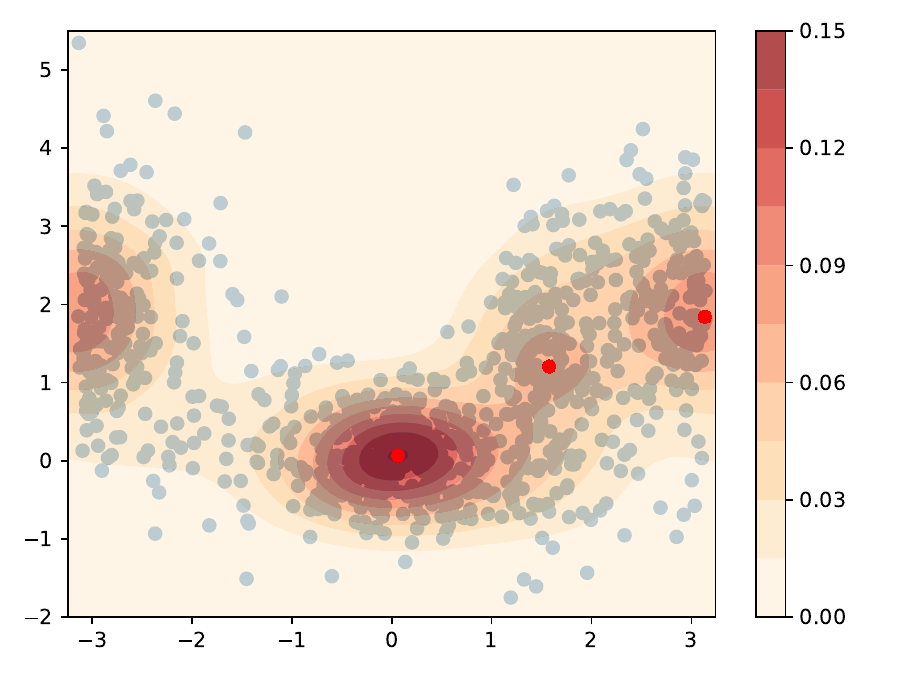}
		\caption{Simulation 1: DirLin data with the contour plot of KDE.}
	\end{subfigure}
	\hfil
	\begin{subfigure}[t]{.32\linewidth}
		\centering
		\includegraphics[width=1\linewidth]{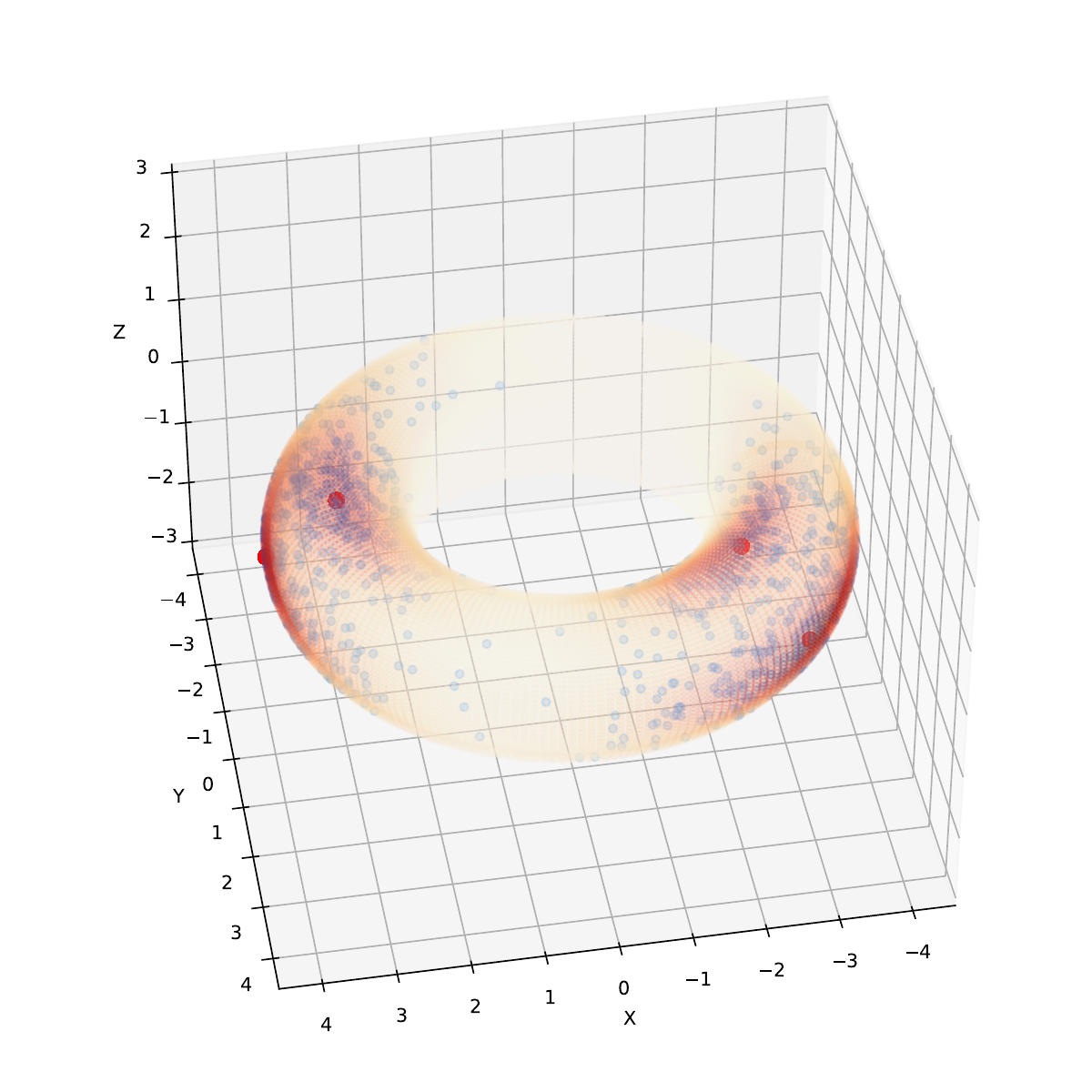}
		\caption{Simulation 2: DirDir data with the contour plot of KDE.}
	\end{subfigure}
	\caption{Local modes obtained by our mean shift algorithm on the simulated DirLin and DirDir data. In each panel, the red dots are estimated local modes while the blue dots are simulated points.}
	\label{fig:ModeSeek_sim}
\end{figure*}

We first simulate 1000 points from a vMF-Gaussian mixture model inspired by \cite{Dir_Linear2013}:
\begin{align*}
	&\frac{2}{5} \text{vMF}\left((1,0), 3\right) \mathcal{N}\left(0,\frac{1}{4} \right) + \frac{1}{5} \text{vMF}\left((0,1),10 \right) \mathcal{N}(1,1) + \frac{2}{5} \text{vMF}\left((-1,0),3 \right) \mathcal{N}\left(2, 1\right),
\end{align*}
where the original study compared directional-linear KDE's exact and asymptotic Mean Integrated Squared Error (MISE). Here, we extend their analysis to identify the local modes of the directional-linear KDE for this mixture data. 
The directional bandwidth $h_1\approx 0.454$ and linear bandwidth $h_2\approx 0.314$ are selected via the rules of thumb described in \autoref{Sec:Add_exp_bw}. Data points with density values below the 5\% quantiles of overall density estimates on the dataset are excluded to avoid spurious modes before applying the mean shift algorithm \eqref{MS_iter_sim} or \eqref{MS_iter_comp}. Panels (a-b) of \autoref{fig:ModeSeek_sim} present the estimated local modes 
$$(0.998, 0.064, 0.066), (0, 1, 1.20), (-1, 0, 1.84) \in \Omega_1 \times \mathbb{R}$$ 
obtained by our mean shift algorithm on this simulated DirLin data. To quantitatively compare our proposed mean shift algorithms with existing approaches, we estimate the set of local modes $\mathcal{M}$ using the following three methods:
\begin{enumerate}[label=(\roman*)]
	\item The standard Euclidean mean shift algorithm on $\mathbb{R}^3 \supset \Omega_1\times \mathbb{R}$.
	\item The directional mean shift algorithm on $\Omega_1$ and the Euclidean mean shift algorithm on $\mathbb{R}$ independently.
	\item Our proposed (DirLin) mean shift algorithm on $\Omega_1\times \mathbb{R}$.
\end{enumerate}
For Method (ii), since the local modes are estimated separately on $\Omega_1$ and $\mathbb{R}$, it may be difficult to combine the resulting local modes on $\Omega_1$ and $\mathbb{R}$ into a set of joint local modes on $\Omega_1\times \mathbb{R}$. Hence, we leverage the ground-truth $\mathcal{M}_1$ to perform an optimal matching. Then, we compare the average distances \eqref{avg_err_dist} from each estimated mode in $\hat{\mathcal{M}}_1$ yielded by the above three methods to the true mode $\mathcal{M}_1$ across different sample sizes over 1000 Monte Carlo simulations in \autoref{table:avg_dist_sim1}. The results demonstrate that our proposed mean shift algorithm outperforms the existing mean shift algorithm in estimating local modes from directional-linear data on $\Omega_1\times \mathbb{R}$. Notably, the undesired performance of Method (ii) reiterates the need to design a joint mode-seeking algorithm for a data-generating distribution supported on a product space. Additionally, Euclidean mean shift algorithm on $\mathbb{R}^3$ performs only slightly worse than our proposed method as the sample size increases. This is because $\Omega_1\times \mathbb{R}$ is locally Euclidean in $\mathbb{R}^3$ and simulated data become concentrated around the true local modes as the sample size increases. Finally, we observed in our experiments that the simultaneous \eqref{MS_iter_sim} and componentwise \eqref{MS_iter_comp} mean shift algorithms yield equivalent modes under von Mises and Gaussian kernels, while the simultaneous version excels in terms of time efficiency. 

\begin{table}[h]
	\centering
	\begin{tabular}{c c c c} 
		\hline
		 & {\bf Method (i)} & {\bf Method (ii)} & {\bf Method (iii) (Proposed)} \\
		\hline\hline
		$n=500$ & 0.252 ($3.641\times 10^{-3}$) & 0.882 ($1.698\times 10^{-3}$) & {\bf 0.180} ($2.578\times 10^{-3}$) \\ 
		$n=1000$ & 0.213 ($3.067\times 10^{-3}$) & 0.881 ($7.238\times 10^{-4}$) & {\bf 0.163} ($2.410\times 10^{-3}$) \\
		$n=2000$ & 0.181 ($2.517\times 10^{-3}$) & 0.880 ($5.467\times 10^{-4}$) & {\bf 0.148} ($2.217\times 10^{-3}$) \\
		\hline
	\end{tabular}
	\caption{The average distances $d_{\text{avg}}(\hat{\mathcal{M}}_1, \mathcal{M}_1) =\frac{1}{|\hat{\mathcal{M}}_1|}\sum_{\hat{\bm{m}}\in \hat{\mathcal{M}}_1} d(\hat{\bm{m}},\mathcal{M}_1)$ yielded by three different mean shift methods under different sample sizes in the context of Simulation 1. The standard errors are given in parenthesis.}
	\label{table:avg_dist_sim1}
\end{table}

%\begin{align*}
%	d_{\text{avg}}(\hat{\mathcal{M}}_1, \mathcal{M}_1) 
%	% &=\frac{1}{|\hat{\mathcal{M}}_1|}\sum_{\hat{\bm{m}}\in \hat{\mathcal{M}}_1} d(\hat{\bm{m}},\mathcal{M}_1) 
%	\approx\begin{cases}
%		0.2131 \;(\text{standard error: } 3.067\times 10^{-3})& \text{ Method (i)},\\
%		0.8805 \;(\text{standard error: } 7.238\times 10^{-4}) & \text{ Method (ii)},\\
%		\mathbf{0.1625} \;(\text{standard error: } 2.410 \times 10^{-3}) & \text{ Method (iii)}.
%	\end{cases}
%\end{align*}

\subsection{Simulation 2: Mode-Seeking on $\Omega_1\times \Omega_1$} 
\label{subsec:sim2}

For the mode seeking task on $\Omega_1\times \Omega_1$, we consider two different data-generating process.  

$\bullet$ {\bf Product of two vMF mixture densities:} We sample 1000 points from an independent product of two vMF mixture densities as:
\begin{align*}
	&\left[\frac{1}{2} \text{vMF}((1,0), 5) + \frac{1}{2} \text{vMF}((0,1), 5)\right] \times \left[\frac{1}{2} \text{vMF}\left((1,0), 7 \right) + \frac{1}{2} \text{vMF}\left((1/\sqrt{2}, -1/\sqrt{2}), 7 \right)\right],
\end{align*}
where the four true local modes are $(0,0)$, $(0,3\pi/4)$, $(\pi/2, 0)$, and $(\pi/2, 3\pi/4)$ under the angular coordinate system. We apply our mean shift algorithm on this simulated data with bandwidths $h_1\approx 0.218, h_2\approx 0.339$ selected via \eqref{bw_ROT_Dir} with $D_1=1$. Panel (c) of \autoref{fig:ModeSeek_sim} on a torus shows the estimated local modes $(-0.01, -0.05)$, $(0.01, 2.37)$, $(1.52, 0.04)$, and $(1.70, 2.38)$, which align with the KDE contours of \eqref{KDE_prod} and are close to the true modes. Furthermore, we compute the average distance $d_{\text{avg}}(\mathcal{M}_2,\hat{\mathcal{M}}_2)$ from $\mathcal{M}_2$ to its estimator $\hat{\mathcal{M}}_2$ yielded by the following two methods:
\begin{enumerate}[label=(\alph*)]
	\item The Euclidean mean shift algorithm on $\mathbb{R}^4 \supset \Omega_1\times \Omega_1$.
	\item Our proposed (DirDir) mean shift algorithm on $\Omega_1\times\Omega_1$.
\end{enumerate}
Both methods identify the correct number of local modes, making their comparison of $d_{\text{avg}}(\hat{\mathcal{M}}_2,\mathcal{M}_2)$ reasonable. We exclude the experiments that apply the directional mean shift algorithm independently to each coordinate of $\Omega_1\times \Omega_1$ due to its unidentifiability issue. The average distances from the estimated local modes yielded by these two methods to the true ones across different sample sizes over 1000 Monte Carlo simulations are shown in \autoref{table:avg_dist_sim2}, which again demonstrates the effectiveness of our proposed method on the product space $\Omega_1\times \Omega_1$.

\begin{table}[h]
	\centering
	\begin{tabular}{c c c} 
		\hline
		& {\bf Method (a)} & {\bf Method (b) (Proposed)} \\
		\hline\hline
		$n=500$ & 0.113 ($1.172\times 10^{-3}$) & {\bf 0.099} ($9.899\times 10^{-4}$) \\ 
		$n=1000$ & 0.085 ($8.118\times 10^{-4}$) & {\bf 0.078} ($7.812\times 10^{-4}$) \\
		$n=2000$ & 0.068 ($6.187\times 10^{-3}$) & {\bf 0.066} ($6.324\times 10^{-3}$) \\
		\hline
	\end{tabular}
	\caption{The average distances $d_{\text{avg}}(\hat{\mathcal{M}}_2,\mathcal{M}_2) =\frac{1}{|\mathcal{M}_2|}\sum_{\bm{m}\in \mathcal{M}_2} d(\bm{m},\hat{\mathcal{M}}_2)$ yielded by two different mean shift methods under different sample sizes when the underlying density is a product of two vMF densities on $\Omega_1\times \Omega_1$. The standard errors are given in parenthesis.}
	\label{table:avg_dist_sim2}
\end{table}

%\[
%d_{\text{avg}}(\hat{\mathcal{M}}_2,\mathcal{M}_2)
%% =\frac{1}{|\mathcal{M}_2|}\sum_{\bm{m}\in \mathcal{M}_2} d(\bm{m},\hat{\mathcal{M}}_2)
%\approx\begin{cases}
%	0.0855  \;(\text{standard error: } 8.118\times 10^{-4}) & \text{ Method (a)},\\
%	\mathbf{0.0784} \;(\text{standard error: } 7.812\times 10^{-4}) & \text{ Method (b)},
%\end{cases} 
%\]

$\bullet$ {\bf Mixture of bivariate von Mises densities:} The previous data-generating distribution was an independent product of two vMF mixture densities, which may inherently favor our proposed DirDir mean shift algorithm. To further illuminate the superiority of our proposed algorithm over the standard Euclidean mean shift method, we sample i.i.d. points $\left\{(\theta_i,\phi_i)\right\}_{i=1}^n$ from a mixture of bivariate von Mises densities on $\Omega_1\times \Omega_1$ as \citep{mardia1975statistics}:
\begin{align*}
&f(\theta,\phi) \\
&\propto \frac{1}{2}\exp\left[\kappa_1 \cos\left(\theta-\mu_1\right) + \kappa_2\cos(\phi - \mu_2) + \left(\cos(\theta-\mu_1), \sin(\theta-\mu_1)\right)\bm{A}_1 \begin{pmatrix}
	\cos(\phi-\mu_2)\\
	\sin(\phi - \mu_2)
\end{pmatrix} \right] \\
&\quad + \frac{1}{2}\exp\left[\kappa_3 \cos\left(\theta-\mu_3\right) + \kappa_4\cos(\phi - \mu_4) + \left(\cos(\theta-\mu_3), \sin(\theta-\mu_3)\right)\bm{A}_3 \begin{pmatrix}
	\cos(\phi-\mu_4)\\
	\sin(\phi - \mu_4)
\end{pmatrix} \right],
\end{align*}
where $\mu_1=\mu_2=0$, $\kappa_1=\kappa_2=10$, $\bm{A}_1=\begin{pmatrix}
	-1 & 0.1\\
	0.1 & 1
\end{pmatrix}$, $\mu_3=\frac{3\pi}{4}, \mu_4=\frac{\pi}{2}$, $\kappa_3=\kappa_4=5$, and $\bm{A}_3=\begin{pmatrix}
0 & 0\\
0 & 1
\end{pmatrix}$. Consequently, the two true local modes are $\left(0,0\right)$ and $\left(\frac{3\pi}{4}, \frac{\pi}{2}\right)$. As in the previous experiments, we compare the performance of our proposed DirDir mean shift algorithm with the Euclidean mean shift method across different sample sizes over 1000 Monte Carlo simulations. The results, shown in \autoref{table:avg_dist_sim2add}, further confirm the superior accuracy of our proposed algorithm when handling the data on product spaces.

\begin{table}[h]
	\centering
	\begin{tabular}{c c c} 
		\hline
		& {\bf Method (a)} & {\bf Method (b) (Proposed)} \\
		\hline\hline
		$n=500$ & 0.101 ($1.099\times 10^{-3}$) & $\mathbf{9.355\times 10^{-2}}$ ($9.771\times 10^{-4}$) \\ 
		$n=1000$ & $9.471\times 10^{-2}$ ($8.390\times 10^{-4}$) & $\mathbf{8.980\times 10^{-2}}$ ($7.934\times 10^{-4}$) \\
		$n=2000$ & $8.780\times 10^{-2}$ ($7.183\times 10^{-4}$) & $\mathbf{8.532\times 10^{-2}}$ ($6.324\times 10^{-3}$) \\
		\hline
	\end{tabular}
	\caption{The average distances $d_{\text{avg}}(\hat{\mathcal{M}}_2,\mathcal{M}_2) =\frac{1}{|\mathcal{M}_2|}\sum_{\bm{m}\in \mathcal{M}_2} d(\bm{m},\hat{\mathcal{M}}_2)$ yielded by two different mean shift methods under different sample sizes when the underlying density is a mixture of bivariate von Mises densities on $\Omega_1\times \Omega_1$. The standard errors are given in parenthesis.}
	\label{table:avg_dist_sim2add}
\end{table}

\subsection{Simulation 3: Ridge-Finding on $\Omega_2 \times \mathbb{R}$}
\label{subsec:sim3}

To compare our proposed SCMS method with the regular/Euclidean SCMS algorithm on $\Omega_2\times \mathbb{R}$, we study a spiral curve example defined as:
\begin{align*}
	\mathcal{C} &= \left\{\left(t\cos(\pi/6) \cos(5t), t\cos(\pi/6) \sin(5t), t\sin(\pi/6) \right): 0\leq t \leq 4 \right\}\\ 
	&\equiv \left\{(\xi, \phi, R) = \left((900t/\pi)^{\circ}, 30^{\circ}, t \right): 0\leq t\leq 4\right\},
\end{align*}
where we convert the first two radian coordinates into their degree measures in the second angular-linear representation of $\mathcal{C}$. We sample 1000 observations on the spiral curve with additive Gaussian noises $\mathcal{N}(0,0.2^2)$ to their angular-linear coordinates; see Panel (a) of \autoref{fig:spiral_cur} in the main paper. One may think of $(\xi,\phi,R)$ as (right ascension, declination, redshift) under the astronomical survey coordinate system; recall \autoref{subsec:cosmic_web} in the main paper. Hence, each simulated observation has different representations in three coordinate systems: 
\begin{enumerate}[label=(\roman*)]
	\item Cartesian in $\mathbb{R}^3$ as $(x_i,y_i,z_i)$;
	\item angular-linear on $\Omega_2\times \mathbb{R}$ as $(\xi_i,\phi_i, R_i)$;
	\item directional-linear on $\Omega_2\times \mathbb{R}$ as $(\bm{X}_i,R_i)$,
\end{enumerate}
where, for $i=1,...,1000$, 
\begin{align*}
(x_i,y_i,z_i) &= \left(R_i\cos\phi_i \cos\xi_i, R_i\cos\phi_i \sin\xi_i, R_i\sin\phi_i \right),\\
\bm{X}_i &= \left(\cos\phi_i \cos\xi_i, \cos\phi_i \sin\xi_i, \sin\phi_i \right).
\end{align*} 
We apply the regular SCMS algorithm to $\{(x_i,y_i,z_i)\}_{i=1}^{1000}$ and $\{(\xi_i,\phi_i, R_i)\}_{i=1}^{1000}$ as well as our proposed SCMS algorithm to $\{(\bm{X}_i,R_i)\}_{i=1}^{1000}$ based on the same simulated dataset. As shown in \autoref{fig:spiral_cur}, the regular SCMS algorithm fails to recover the underlying curve, while our proposed SCMS algorithm converges to estimated ridges close to the true structure. Quantitatively, our proposed SCMS algorithm also outperforms the regular SCMS algorithm in terms of the average manifold-recovering error measure in $\mathbb{R}^3$ defined in \autoref{Sec:Manifold_Rec_err} across different sample sizes over 1000 Monte Carlo simulations; see \autoref{table:avg_dist_sim3} below.

Other simulation studies on ridge estimation with our proposed SCMS algorithm \eqref{scaled_SCMS} on DirLin and DirDir data are delineated in \autoref{Sec:stepsize} when we investigate the effects of varying the step size $\eta$.

\begin{table}[h]
	\centering
	\begin{tabular}{c c c c} 
		\hline
		& {\bf Regular SCMS on $\mathbb{R}^3$} & {\bf Regular SCMS on $\Omega_2\times \mathbb{R}$} & {\bf Our proposed SCMS} \\
		\hline\hline
		$n=500$ & 0.258 ($1.7\times 10^{-5}$) & 0.537 ($2.1\times 10^{-5}$) & {\bf 0.0836} ($6.0\times 10^{-6}$) \\ 
		$n=1000$ & 0.195 ($1.2\times 10^{-5}$) & 0.532 ($1.6\times 10^{-5}$) & {\bf 0.0580} ($3.0\times 10^{-6}$) \\
		$n=2000$ & 0.138 ($6.0\times 10^{-6}$) & 0.528 ($1.1\times 10^{-5}$) & {\bf 0.0440} ($2.0\times 10^{-6}$) \\
		\hline
	\end{tabular}
	\caption{The average manifold-recovering error measure $d_H\left(\hat{\mathcal{R}}_1,\mathcal{C} \right)$ yielded by three different SCMS algorithms under different sample sizes in the context of Simulation 3. The standard errors are given in parenthesis.}
	\label{table:avg_dist_sim3}
\end{table}

%\[
%d_H\left(\hat{\mathcal{R}}_1,\mathcal{C} \right)\approx
%\begin{cases}
%	0.1950 \; (\text{standard error: } 1.2\times 10^{-5}) & \text{ for regular SCMS algorithm on $\mathbb{R}^3$},\\
%	0.5319  \; (\text{standard error: } 1.6\times 10^{-5}) & \text{ for regular SCMS algorithm on $\Omega_2\times \mathbb{R}$},\\
%	0.0580 \; (\text{standard error: } 3.0\times 10^{-6})  & \text{ for our proposed SCMS algorithm}.
%\end{cases}
%\]

\subsection{Simulation 4: Surface-Recovering on $\Omega_2\times \mathbb{R}$}

We have demonstrated the effectiveness of our proposed SCMS algorithm in recovering the underlying spiral curve from noisy observations in Simulation 3 of \autoref{subsec:sim3} above.
Additionally, our previous experiments have focused exclusively on identifying the local modes or density ridges with intrinsic (or manifold) dimension 1. To extend our analyses, we now present an example of recovering a hidden surface using our proposed SCMS algorithm and compare its performance with the regular SCMS algorithm. 

\begin{figure*}[t]
	\captionsetup[subfigure]{justification=centering}
	\centering
	\begin{subfigure}[t]{.24\linewidth}
		\centering
		\includegraphics[width=1\linewidth]{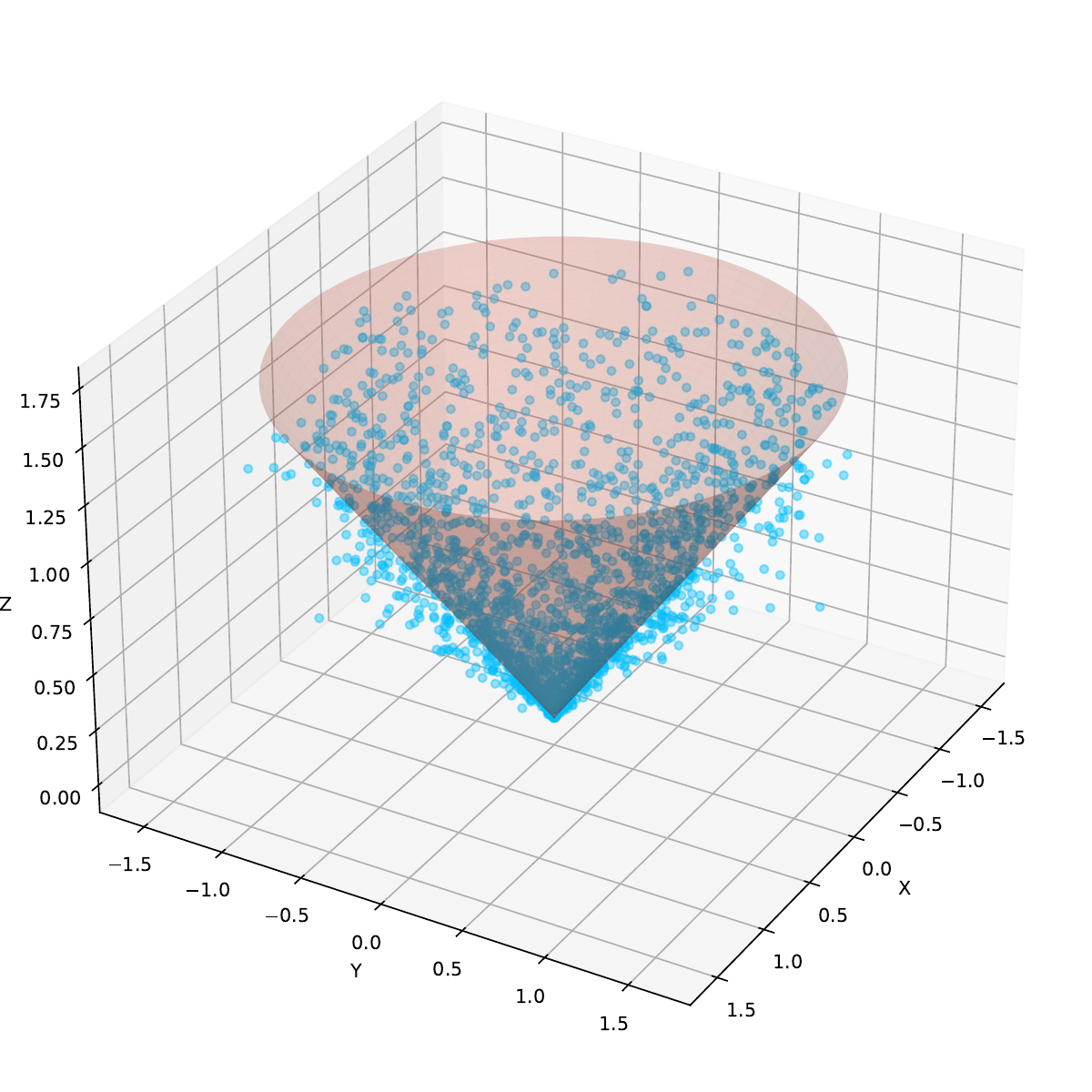}
		\caption{Simulated points around a spherical cone.}
	\end{subfigure}
	\hfil
	\begin{subfigure}[t]{.24\linewidth}
		\centering
		\includegraphics[width=1\linewidth]{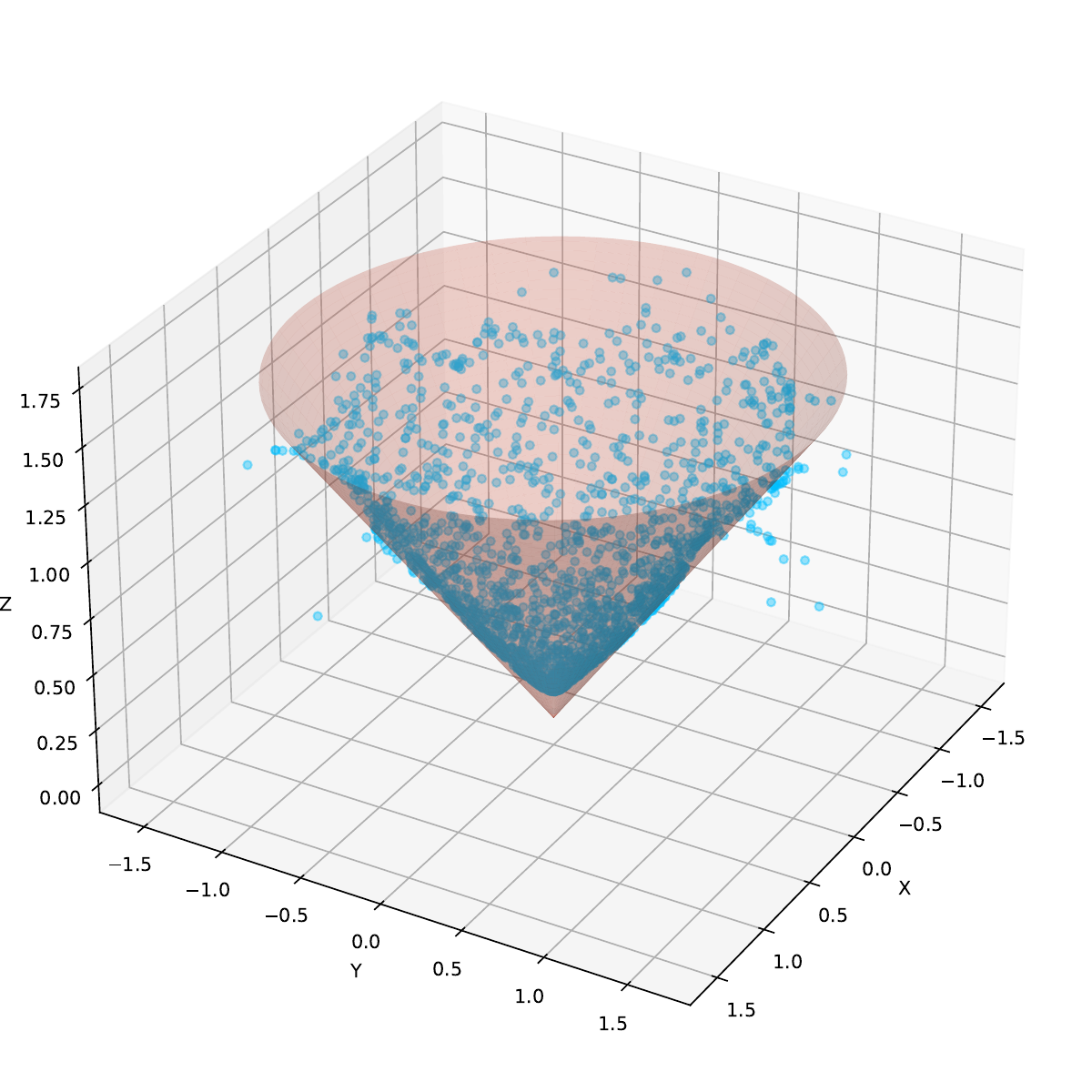}
		\caption{Regular SCMS on Cartesian coordinates in $\mathbb{R}^3$.}
	\end{subfigure}
	\hfil
	\begin{subfigure}[t]{.24\linewidth}
		\centering
		\includegraphics[width=1\linewidth]{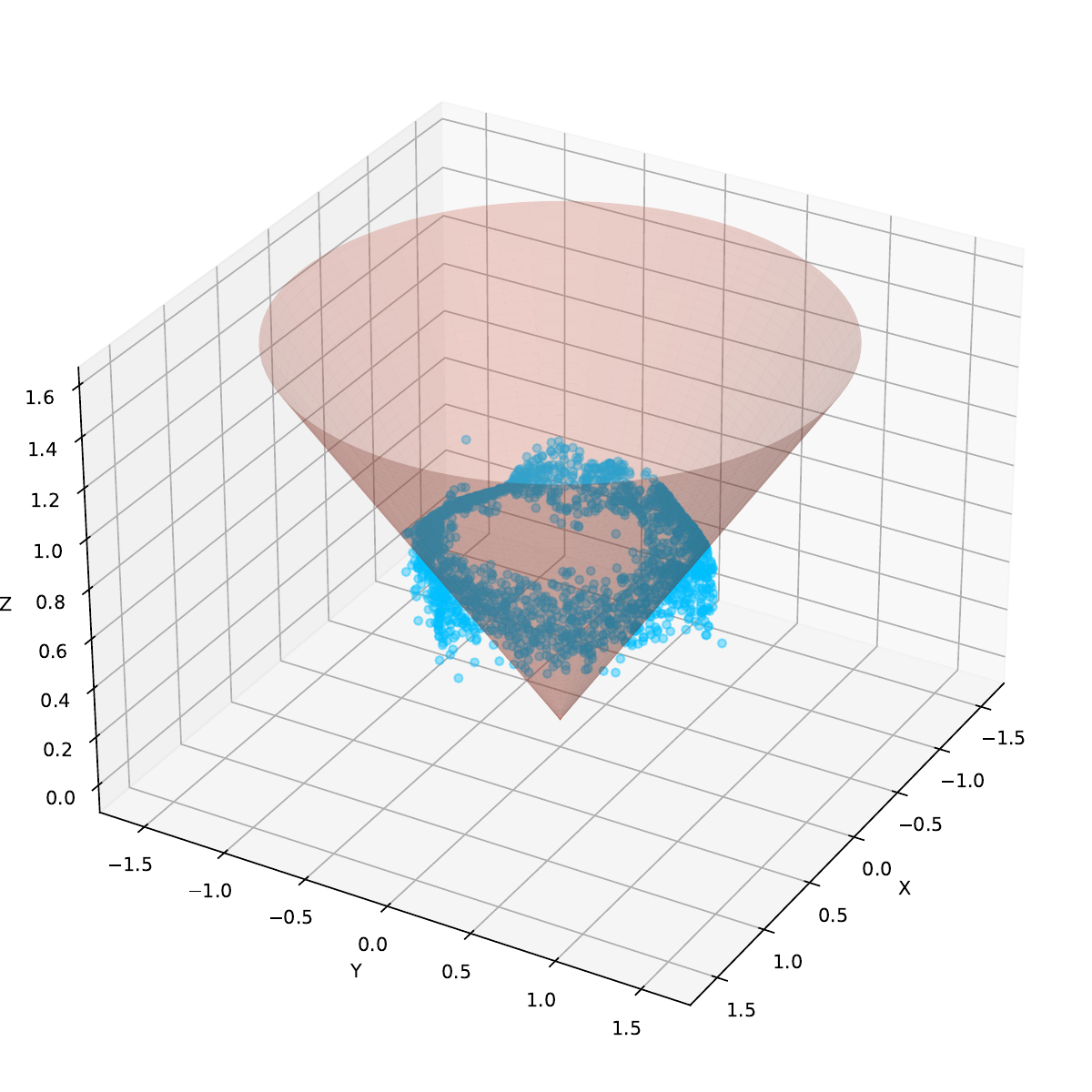}
		\caption{Regular SCMS on angular-linear coordinates $(\xi, \varphi,R)$ in $\Omega_2 \times \mathbb{R}$.}
	\end{subfigure}
	\hfil
	\begin{subfigure}[t]{.24\linewidth}
		\centering
		\includegraphics[width=1\linewidth]{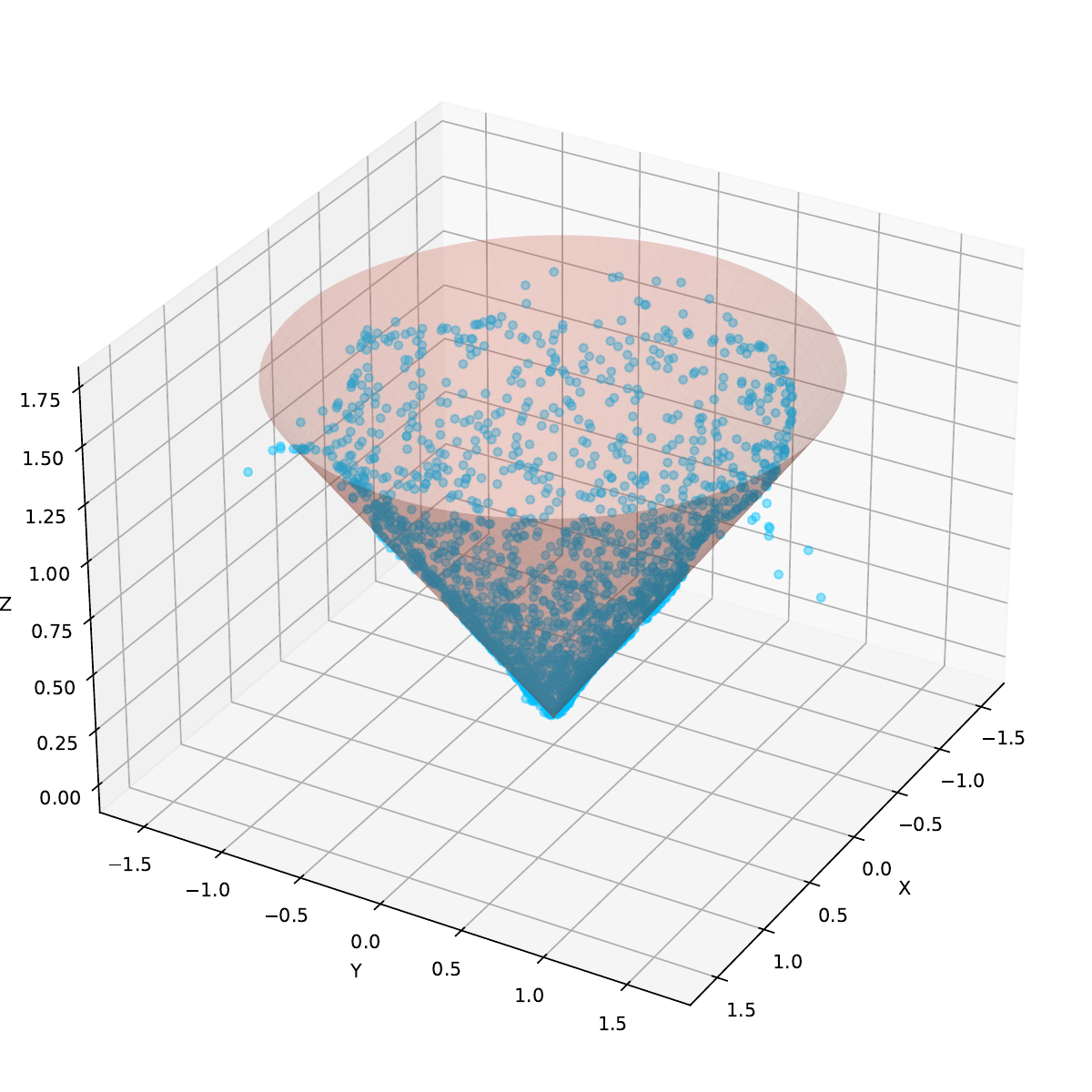}
		\caption{{\bf Our SCMS on directional-linear coordinates in $\Omega_2 \times \mathbb{R}$.}}
	\end{subfigure}
	\caption{Estimated ridges obtained by various SCMS algorithms on the spherical cone data. In each panel, the red surface is the hidden manifold structure while the blue dots are final convergent points of the corresponding SCMS algorithm.}
	\label{fig:sph_cone}
\end{figure*}

We simulate a dataset with 2000 points $\left\{\bm{X}_i \right\}_{i=1}^{2000}$ on the circle of latitude $45^{\circ}$ on the unit sphere $\Omega_2$ with additive Gaussian noises $\mathcal{N}(0,0.1^2)$ applied to their Cartesian coordinates in $\mathbb{R}^3$. To ensure that the noisy simulated data points $\left\{\bm{X}_i \right\}_{i=1}^{2000}$ remain on $\Omega_2$, we standardize them via $\bm{X}_i=\frac{\bm{X}_i}{\norm{\bm{X}_i}_2}$. Additionally, we sample 2000 observations $\left\{R_i\right\}_{i=1}^{2000}$ uniformly from the interval $[0,2]$, forming the directional-linear dataset $\left\{(\bm{X}_i,R_i) \right\}_{i=1}^{2000}$. For any data point $(\bm{X}_i,R_i)$, we can view $\bm{X}_i$ as its position on a sphere, while $R_i$ represents the sphere's radius for $i=1,...,2000$. Notably, the true manifold structure thus corresponds to a spherical cone $\mathcal{C}$ with its apex angle as $\frac{\pi}{2}$; see Panel (a) of \autoref{fig:sph_cone}. As in Simulation 3 above, the directional-linear dataset $\left\{(\bm{X}_i,R_i) \right\}_{i=1}^{2000}$ has alternative representations as the Cartesian coordinate in $\mathbb{R}^3: (x_i,y_i,z_i)$ and the angular-linear coordinate on $\Omega_2\times \mathbb{R}: (\xi_i,\phi_i, R_i),$
where $(\xi_i,\phi_i)$ is the angular representation of $\bm{X}_i$ in degree measure and 
$(x_i,y_i,z_i) = \left(R_i\cos\phi_i \cos\xi_i, R_i\cos\phi_i \sin\xi_i, R_i\sin\phi_i \right)$ 
for $i=1,...,2000$. We apply the regular SCMS algorithm to the data representations $\{(x_i,y_i,z_i)\}_{i=1}^{1000}$ and $\{(\xi_i,\phi_i, R_i)\}_{i=1}^{1000}$ as well as our proposed SCMS algorithm to $\{(\bm{X}_i,R_i)\}_{i=1}^{1000}$ based on the same simulated dataset in order to estimate density ridges (or principal surfaces) with dimension 2. The regular SCMS algorithm under the angular-linear coordinate system fails to approximate the true spherical cone with its yielded ridge; see Panel (c) of \autoref{fig:sph_cone}. While the regular SCMS algorithm in $\mathbb{R}^3$ produces an estimated ridge capturing most parts of the true spherical cone, it struggles to recover the surface near the apex; see Panel (b) of \autoref{fig:sph_cone}. In contrast, our proposed SCMS algorithm converges to an estimated ridge close to the true structure. Quantitatively, our proposed SCMS algorithm also outperforms the regular one in terms of the average manifold-recovering error measure \eqref{manifold_rec_err} in $\mathbb{R}^3$ across different sample sizes over 1000 Monte Carlo simulations; see \autoref{table:avg_dist_sim4} below.

\begin{table}[h]
	\centering
	\begin{tabular}{c c c c} 
		\hline
		& {\bf Regular SCMS on $\mathbb{R}^3$} & {\bf Regular SCMS on $\Omega_2\times \mathbb{R}$} & {\bf Our proposed SCMS} \\
		\hline\hline
		$n=500$ &  0.0724 ($3.0\times 10^{-6}$) &  0.231 ($1.3\times 10^{-5}$) & {\bf 0.0551} ($3.0\times 10^{-6}$) \\ 
		$n=1000$ & 0.0588 ($2.0\times 10^{-6}$) & 0.242 ($1.1\times 10^{-6}$) & {\bf 0.0424} ($2.0\times 10^{-6}$) \\
		$n=2000$ & 0.0488 ($2.0\times 10^{-6}$) & 0.252 ($9.0\times 10^{-6}$) & {\bf 0.0334} ($2.0\times 10^{-6}$) \\
		\hline
	\end{tabular}
	\caption{The average manifold-recovering error measure $d_H\left(\hat{\mathcal{R}}_2,\mathcal{C} \right)$ yielded by three different SCMS algorithms under different sample sizes in the context of Simulation 4. The standard errors are given in parenthesis.}
	\label{table:avg_dist_sim4}
\end{table}

%\[
%d_H\left(\hat{\mathcal{R}}_2,\mathcal{C} \right)\approx
%\begin{cases}
%	0.0488 \; (\text{standard error: } 2.0\times 10^{-6}) & \text{ for regular SCMS algorithm on $\mathbb{R}^3$},\\
%	0.2526 \; (\text{standard error: } 9.0\times 10^{-6}) & \text{ for regular SCMS algorithm on $\Omega_2\times \mathbb{R}$},\\
%	0.0334 \; (\text{standard error: } 2.0\times 10^{-6}) & \text{ for our proposed SCMS algorithm}.
%\end{cases}
%\]

\section{Statistical Consistency}
\label{Sec:Cons_Theory}

This section is devoted to the discussion of consistency results for 
$\hat{f}_{\bm{h}}(\bm{z})$
defined in \eqref{KDE_prod} of the main paper (and its derivatives), its induced mode estimator $\hat{\mathcal{M}}$ in \eqref{ProdDen_Mode_est}, and ridge estimator $\hat{\mathcal{R}}_d$ in \eqref{ProdDen_Ridge_est} on the product space $\mathcal{S}_1\times \mathcal{S}_2$. 

We reserve the notation $D^{[\tau]}f = f^{(\tau)}$ for the $|[\tau]|$-th order partial derivative operator with the coordinate chart of $\mathcal{S}_1\times \mathcal{S}_2$, where $[\tau]=(\tau_1,...,\tau_{D_T})$ is a multi-index and $D_T=D_1+D_2$ is the sum of intrinsic dimensions of $\mathcal{S}_1$ and $\mathcal{S}_2$. For $\ell=0,1,...$, we define the functional norms 
$\norm{f}_{\infty}^{(\ell)}= \max_{\tau:|[\tau]|=\ell} \sup_{\bm{z}\in \mathcal{S}_1\times \mathcal{S}_2} |f^{(\tau)}(\bm{z})|$.
These functional norms are defined via the derivatives of $f$ within the tangent space $T_{\bm{z}}\equiv T_{\bm{x}}(\mathcal{S}_1)\times T_{\bm{y}}(\mathcal{S}_2)$. We also define $\norm{f}_{\infty,\ell}^* = \max_{i=0,...,\ell}\norm{f}_{\infty}^{(i)}$.
For any matrix $A\in \mathbb{R}^{m\times n}$, its maximum norm is defined as $\norm{A}_{\max}=\max_{i,j} A_{ij}$ with $A_{ij}$ being the $(i,j)$ entry of $A$. 

\subsection{Assumptions and Consistency of KDE}
\label{Sec:Assum_Cons}

Besides assuming that the projection of $f$ on $\mathcal{S}_j$ has a compact support when $\mathcal{S}_j=\mathbb{R}^{D_j}$ is Euclidean for $j=1,2$, we impose the following conditions on $f$ as well as the linear and directional kernel profiles $k$ and $L$:
\begin{itemize}
	\item {\bf (A1)} Under the above extension of the (density) function $f$ on $\mathcal{S}_1\times \mathcal{S}_2$, we assume that the total gradient $\nabla f(\bm{z})$, total Hessian $\nabla^2 f(\bm{z})$, and third-order derivative tensor $\nabla^3 f(\bm{z})$ exists and are continuous in $\mathcal{S}\setminus N_A$. Moreover, they are assumed to be square integrable on $\mathcal{S}_1 \times \mathcal{S}_2$. Finally, $f$ is assumed to have bounded fourth-order derivatives on $\mathcal{S}_1 \times \mathcal{S}_2$.
	
	\item {\bf (A2)} Given the KDE $\hat{f}_{\bm{h}}$ defined in \eqref{KDE_prod} and \eqref{kernel_fun} in the main paper, we assume that its kernel profiles $k_1,k_2:[0,\infty) \to [0,\infty)$ are non-increasing and at least three times continuously differentiable with bounded fourth-order partial derivatives. We also assume that the resulting kernel functions satisfy
	$\int_{\mathbb{R}^{D_j}} \norm{\bm{x}}_2^2 K_j^{(\tau)}(\bm{x}) \,d\bm{x} < \infty$ and $\int_{\mathbb{R}^{D_j}} \left[K_j^{(\tau)}(\bm{x}) \right]^2 d\bm{x} < \infty$
	for any multi-index with $|[\tau]|\leq 3$ if $K_j$ has a linear kernel profile for $j=1,2$, and $0 < \int_0^{\infty} |L^{(\ell)}(r)|^p r^{\frac{D_j}{2}-1} dr < \infty$
	for all $D_j \geq 1$, $p=1,2$, and $\ell =0,1,2,3$ when $K_j$ has a directional kernel profile $L$ for $j=1,2$.
	
	\item {\bf (A3)} Let 
	\begin{align*}
		\mathcal{K} &= \Bigg\{(\bm{u},\bm{v}) \mapsto K_1^{(\tau_1)}\left(\frac{\bm{x}-\bm{u}}{h_1} \right) K_2^{(\tau_2)}\left(\frac{\bm{y}-\bm{v}}{h_2} \right):\\
		& \bm{x},\bm{u}\in \mathcal{S}_1, \bm{y},\bm{v}\in\mathcal{S}_2, |[\tau_1]|,|[\tau_2]|=0,1,2,3, h_1,h_2>0\Bigg\}.
	\end{align*}
	We assume that $\mathcal{K}$ is a bounded VC (subgraph) class of measurable functions on $\mathcal{S}_1\times \mathcal{S}_2$; that is, there exists constants $A,\upsilon > 0$ such that
	$\sup_Q N\left(\mathcal{K}, \mathcal{L}_2(Q), \epsilon ||F||_{\mathcal{L}_2(Q)} \right) \leq \left(\frac{A}{\epsilon} \right)^{\upsilon}$ for any $0<\epsilon <1$,
	where $N(T,d_T, \epsilon)$ is the $\epsilon$-covering number of the pseudometric space $(T, d_T)$, $Q$ is any probability measure with the same support as $P$ on $\mathcal{S}_1\times \mathcal{S}_2$, and $F$ is an envelope function of $\mathcal{K}$. The constants $A$ and $\upsilon$ are usually called the VC characteristics of $\mathcal{K}$, and the norm $\norm{F}_{L_2(Q)}$ is defined as $\left[\int_{\mathcal{S}_1\times \mathcal{S}_2} |F(\bm{x})|^2 dQ(\bm{x}) \right]^{\frac{1}{2}}$.
\end{itemize}

Conditions (A1) and (A2) are standard assumptions for establishing the (pointwise) consistency results for KDE and its derivatives in directional and/or linear product spaces; see \cite{wand1994kernel,All_nonpara2006} for linear/Euclidean KDE and \cite{KDE_Sphe1987,KLEMELA2000,Zhao2001} for directional KDE. The VC-typed assumption (A3) is a well-known regularity condition developed by \cite{Gine2002,Einmahl2005} for the uniform consistency of KDE and its derivatives. The techniques in \cite{KDE_Sphe1987,Asymp_deri_KDE2011,Dir_Linear2013,garcia2015central,DirMS2020} can be adopted to establish the following pointwise rates of convergence for KDE $\hat{f}_{\bm{h}}$ and its derivatives.

\begin{lemma}
	\label{pw_rate_KDE}
	Assume conditions (A1-2). Then, 
	\begin{equation*}
		\bm{\bar{\nabla}}_{\bm{v}}^{\ell} \hat{f}_{\bm{h}}(\bm{z}) - \bm{\bar{\nabla}}_{\bm{v}}^{\ell} f(\bm{z}) = O\left(h^2 \right) + O_P\left(\sqrt{\frac{1}{nh^{D_1+D_2+2\ell}}} \right)
	\end{equation*}
	as $\max(\bm{h}) = \max\left\{h_1,h_2\right\} \lesssim h \to 0$ and $nh^{D_1+D_2+2\ell} \to \infty$ for any integer $\ell$, where $\bm{z}=(\bm{x},\bm{y})\in \mathcal{S}_1\times \mathcal{S}_2$ and $\bm{\bar{\nabla}}_{\bm{v}}$ is the Riemannian connection for any unit $\bm{v}\in T_{\bm{z}} = T_{\bm{x}}(\mathcal{S}_1) \times T_{\bm{y}}(\mathcal{S}_2)$ so that $\bm{\bar{\nabla}}_{\bm{v}} f(\bm{z}) = \langle \grad f(\bm{z}), \bm{v} \rangle$, $\bm{\bar{\nabla}}_{\bm{v}}^2 f(\bm{z}) = \langle \bm{\bar{\nabla}}_{\bm{v}}\grad f(\bm{z}), \bm{v} \rangle= \langle \mathcal{H} f(\bm{z})[\bm{v}], \bm{v} \rangle$, and $\bm{\bar{\nabla}}_{\bm{v}}^3 f(\bm{z}) = \langle \bm{\bar{\nabla}}_{\bm{v}}\mathcal{H}f(\bm{z})[\bm{v}], \bm{v} \rangle$; recall \eqref{Riem_Hess_def} and see also Section 5.3 in \cite{Op_algo_mat_manifolds2009}.
\end{lemma}

The pointwise consistency results in Lemma~\ref{pw_rate_KDE} can be improved as the following uniform rates of convergence using the arguments in \cite{Gine2002,Einmahl2005,Asymp_deri_KDE2011,KDE_direct1988,DirMS2020}.
\begin{lemma}
	\label{unif_rate_KDE}
	Assume conditions (A1-3). Then, 
	$$\norm{\hat{f}_{\bm{h}} - f}_{\infty}^{(\ell)} = \sup_{\bm{z}\in \mathcal{S}_1\times \mathcal{S}_2} \norm{\nabla_{T_{\bm{z}}}^{\ell} \hat{f}_{\bm{h}}(\bm{z}) - \nabla_{T_{\bm{z}}}^{\ell} f(\bm{z})}_{\max} = O(h^2) + O_P\left(\sqrt{\frac{|\log h|}{nh^{D_1+D_2+2\ell}}} \right)$$
	as $\max(\bm{h}) = \max\left\{h_1,h_2\right\} \lesssim h \to 0$ and $\frac{nh^{D_1+D_2+2\ell}}{|\log h|} \to \infty$ for any integer $\ell$.
\end{lemma}

The pointwise and uniform consistency results for $\hat{f}_{\bm{h}}$ only hold within the tangent space/bundle. The (uniform) consistency results are particularly useful for establishing the mode and ridge consistency theory below. More importantly, they serve as a key building block for establishing the (linear) convergence of our proposed SCMS algorithm; see \autoref{Sec:stepsize} for details.

\subsection{Mode Consistency}
\label{Sec:Mode_cons}

Recall that $\mathcal{M}=\left\{\bm{z} \in \mathcal{S}_1\times \mathcal{S}_2: \grad f(\bm{z})=\bm{0}, \lambda_1(\bm{z})<0 \right\} \equiv \left\{\bm{m}_1,...,\bm{m}_N \right\}$ is the set of local modes of the true (density) function $f$ on $\mathcal{S}_1\times \mathcal{S}_2$, while $$\hat{\mathcal{M}} = \left\{\bm{z} \in \mathcal{S}_1\times \mathcal{S}_2: \grad \hat{f}_{\bm{h}}(\bm{z})=\bm{0}, \hat{\lambda}_1(\bm{z})<0 \right\} \equiv \left\{\hat{\bm{m}}_1,...,\hat{\bm{m}}_{\hat{N}} \right\}$$ 
denotes the collection of estimated local modes obtained by KDE $\hat{f}_{\bm{h}}$. Here, $N$ refers to the number of true local modes, and $\hat{N}$ is number of estimated local modes. To quantify the difference between two sets of points $\mathcal{M}$ and $\hat{\mathcal{M}}$, we leverage the Hausdorff distance as:
\begin{align}
	\label{Haus_dist}
	\begin{split}
		\Haus(\mathcal{M}, \hat{\mathcal{M}}) &= \max\left\{\sup_{\bm{m}\in \mathcal{M}} d(\bm{m},\hat{\mathcal{M}}), \sup_{\hat{\bm{m}}\in \hat{\mathcal{M}}} d(\hat{\bm{m}},\mathcal{M}) \right\} \\
		&= \inf\left\{\epsilon>0: \mathcal{M} \subset \hat{\mathcal{M}} \oplus \epsilon, \hat{\mathcal{M}} \subset \mathcal{M} \oplus \epsilon \right\}.
	\end{split}
\end{align}
Here, $d(\hat{\bm{m}}, \mathcal{M}) = \inf\left\{\norm{\bm{m}-\hat{\bm{m}}}_2: \bm{m} \in \mathcal{M} \right\}$, $\mathcal{M} \oplus \epsilon = \left\{\bm{z}\in \mathcal{S}_1\times \mathcal{S}_2: d(\bm{z},\mathcal{M}) \leq \epsilon \right\}$, and other quantities follow similarly.

\begin{theorem}
	\label{Thm:Mode_cons}
	Assumes conditions (A1-3) and there exist constants $\lambda^*, \Theta^*, C^* >0$ such that
	$$0 < \lambda^* \leq |\lambda_1(\bm{m})| \quad \text{ for any } \bm{m} \in \mathcal{M} \quad \text{(strict definiteness condition)}$$
	as well as
	$\left\{\bm{z}\in \mathcal{S}_1\times \mathcal{S}_2: \norm{\grad f(\bm{z})}_2 \leq \Theta^*, \lambda_1(\bm{z}) \leq -\frac{\lambda^*}{2} < 0 \right\} \subset \mathcal{M} \oplus \frac{C^*\lambda^*}{\norm{f}_{\infty}^{(3)}},$
	where we recall that $\lambda_1(\bm{z}) \geq \cdots \geq \lambda_{D_1+D_2}(\bm{z})$ are the eigenvalues of the Riemannian Hessian $\mathcal{H}f(\bm{z})$ associated with the eigenvectors within the tangent space $T_{\bm{z}} = T_{\bm{x}}(\mathcal{S}_1) \times T_{\bm{y}}(\mathcal{S}_2)$. Then, when $\norm{\hat{f}_{\bm{h}} - f}_{\infty,2}^*$ is sufficiently small, there exist constant $A^*, B^* >0$ such that
	$$\mathbb{P}\left(\hat{N} \neq N \right) \leq B^* e^{-A^* nh^{D_1+D_2+4}} \quad \text{ and } \quad \Haus(\mathcal{M}, \hat{\mathcal{M}}) = O\left(h^2 \right) + O_P\left(\sqrt{\frac{1}{nh^{D_1+D_2+2}}} \right)$$
	with $\max(\bm{h}) = \max\left\{h_1,h_2 \right\} \lesssim h$.
\end{theorem}

The assumptions imposed in Theorem~\ref{Thm:Mode_cons} ensure that $f$ is strongly concave around its local modes so that the local modes are well-separated. The proof is a direct application of the arguments in Theorem 1 of \cite{Mode_clu2016} and Theorem 6 of \cite{DirMS2020}, and is therefore omitted here. The key argument of the proof relies on the Taylor's expansion of $f$ on $\mathcal{S}_1\times \mathcal{S}_2$ as \citep{pennec2006intrinsic}:
\begin{equation}
	\label{taylor}
	f(\Exp_{\bm{z}}(\bm{v})) = f(\bm{z}) + \langle \grad f(\bm{z}), \bm{v} \rangle + \frac{1}{2} \bm{v}^{\top} \mathcal{H} f(\bm{z}) \bm{v} + O(\norm{\bm{v}}_2^3)
\end{equation}
for any $\bm{v} \in T_{\bm{z}}$, where $\Exp_{\bm{z}}: T_{\bm{z}} \to \mathcal{S}_1\times \mathcal{S}_2$ is the \emph{exponential map} at $\bm{z} \in \mathcal{S}_1\times \mathcal{S}_2$ such that the tangent vector $\bm{v} \in T_{\bm{z}}$ is mapped to the point $\bm{u}:= \Exp_{\bm{z}}(\bm{v}) \in \mathcal{S}_1\times \mathcal{S}_2$ along a curve $\alpha$ on $\mathcal{S}_1\times \mathcal{S}_2$ with $\alpha(0)=\bm{z}, \alpha(1)=\bm{u}$ and $\alpha'(0)=\bm{v}$. Here, $\alpha:[0,1] \to \mathcal{S}_1\times \mathcal{S}_2$ is the curve of minimum length between $\bm{z}$ and $\bm{u}$ (i.e., the so-called geodesic on $\mathcal{S}_1\times \mathcal{S}_2$).

From \eqref{unif_rate_KDE}, it is not difficult to see that the required condition that $\norm{\hat{f}_{\bm{h}} - f}_{\infty,2}^*$ is sufficiently small is satisfied as $\max(\bm{h}) = \max\left\{h_1,h_2 \right\} \lesssim h \to 0$ and $\frac{nh^{D_1+D_2+4}}{|\log h|} \to \infty$. Notably, the rate of convergence of $\Haus(\mathcal{M},\hat{\mathcal{M}})$ is identical to the pointwise asymptotic rate of the (Riemannian) gradient estimator. Nevertheless, in order for the rate of convergence of $\Haus(\mathcal{M},\hat{\mathcal{M}})$ to be valid, the true (density) function $f$ and its estimator $\hat{f}_{\bm{h}}$ must be uniformly close, including their (Riemannian) Hessians.

Finally, our mode consistency results are closely related to those in the context of modal regression, which focuses on identifying the local modes of the conditional density of an outcome variable given the covariate \citep{einbeck2006modelling,chen2016nonparametric,chen2018modal,alonso2020introduction}. In particular, \cite{alonso2022new,alonso2023analyzing} utilized the mean shift algorithm and studied the asymptotic properties of multi-modal regression for circular-linear and the circular-circular data.

\subsection{Ridge Consistency}
\label{Sec:Ridge_cons}

Recall that our definition of the order-$d$ (density) ridge 
$$\mathcal{R}_d=\left\{\bm{z} \in \mathcal{S}_1\times \mathcal{S}_2: V_d(\bm{z}) V_d(\bm{z})^{\top} \grad f(\bm{z}) =\bm{0}, \lambda_{d+1}(\bm{z}) < 0 \right\},$$
which can be regarded as a generalized version of the set of local modes within the eigenspace spanned by the last $(D_1+D_2-d)$ eigenvector of the Riemannian Hessian $\mathcal{H} f(\bm{z})$ (i.e., the column of $V_d(\bm{z})$). Given a natural estimator $\hat{\mathcal{R}}_d$ in \eqref{ProdDen_Ridge_est} of the ridge provided by the KDE $\hat{f}_{\bm{h}}$, we modify the assumptions in Theorem 4 in \cite{Non_ridge_est2014} and Theorem 4.1 in \cite{DirSCMS2021} to obtain its consistency under the Hausdorff distance.

\begin{theorem}
	\label{Thm:Ridge_cons}
	Assume conditions (A1-3) and there exist constants $r_*, \beta_0 >0$ such that
	$$\lambda_{d+1}(\bm{z}) \leq -\beta_0, \quad \lambda_d(\bm{z}) -\lambda_{d+1}(\bm{z}) \geq \beta_0 \quad \text{ (eigengap condition)}$$
	as well as
	\begin{align*}
		&(D_1+D_2)^{\frac{3}{2}}\norm{\left[\bm{I}_{D_1+\mathbbm{1}_{\{\mathcal{S}_1=\Omega_{D_1}\}}+D_2+\mathbbm{1}_{\{\mathcal{S}_2=\Omega_{D_2}\}}} - V_d(\bm{z})V_d(\bm{z})^{\top} \right] \grad f(\bm{z})}_2 \norm{\nabla^3 f(\bm{z})}_{\max} \leq \frac{\beta_0^2}{4}
	\end{align*}
	for any $\bm{z}\in \mathcal{R}_d \oplus r_*$. Then, when $\norm{\hat{f}_{\bm{h}} - f}_{\infty,3}^*$ is sufficiently small,
	$\Haus(\hat{\mathcal{R}}_d, \mathcal{R}_d) = O\left(h^2 \right) + O_P\left(\sqrt{\frac{|\log h|}{nh^{D_1+D_2+4}}} \right)$
	with $\max(\bm{h}) = \max\left\{h_1,h_2 \right\} \lesssim h$.
\end{theorem}

The eigengap condition in Theorem~\ref{Thm:Ridge_cons} is a generalization of the strict definiteness condition in Theorem~\ref{Thm:Mode_cons}. It requires $f$ to be (locally) concave within the eigenspace spanned by the last $(D_1+D_2-d)$ eigenvector of the Riemannian Hessian $\mathcal{H}f(\bm{z})$ so that every order-$d$ ridge $\mathcal{R}_d$ contains the lower order ridges but is isolated from the ridges with the same order $d$. As our conditions in Theorem~\ref{Thm:Ridge_cons} imply the conditions in \cite{Non_ridge_est2014,DirSCMS2021}, the proof follows from their arguments.

% \section{Choice of Step Sizes and Linear Convergence of the Proposed SCMS Algorithm on $\mathcal{S}_1\times \mathcal{S}_2$}
\section{Linear Convergence of the Mean Shift and SCMS Algorithms on $\mathcal{S}_1\times \mathcal{S}_2$}
\label{Sec:Lin_conv}

This section is dedicated to formulating the general (subspace constrained) gradient ascent framework on the product space/manifold $\mathcal{S}_1\times \mathcal{S}_2$ and establishing the linear convergence of our proposed mean shift and SCMS algorithms (Theorems~\ref{Thm:MS_lin_conv} and \ref{Thm:SCMS_lin_conv} in the main paper) under this framework. In particular, we argue that our rule of thumb for the step size $\eta$ in the proposed SCMS algorithm \eqref{scaled_SCMS} is well-suited for the linear convergence of the algorithm when the bandwidth parameters $h_1,h_2\lesssim h$ are chosen to be small. In contrast, we demonstrate via the subspace constrained gradient ascent framework and empirical examples that it is inappropriate to set a constant step size, says $\eta=1$, with respect to the bandwidth parameters $h_1,h_2$ in our proposed SCMS algorithm \eqref{scaled_SCMS}.

\subsection{(Subspace Constrained) Gradient Ascent Framework on $\mathcal{S}_1\times \mathcal{S}_2$}

Whenever $\mathcal{S}_j=\Omega_{D_j}$ is a nonlinear manifold for $j=1$ or $2$, the gradient ascent method with an objective function $f$ on $\mathcal{S}_1\times \mathcal{S}_2$ is defined through the exponential map $\Exp_{\bm{z}}: T_{\bm{z}}\equiv T_{\bm{x}}(\mathcal{S}_1) \times T_{\bm{y}}(\mathcal{S}_2) \to \mathcal{S}_1\times \mathcal{S}_2$ as \citep{Op_algo_mat_manifolds2009,Geo_Convex_Op2016}:
\begin{equation}
	\label{GA_manifold}
	\bm{z}^{(t+1)} \gets \Exp_{\bm{z}^{(t)}}\left(\tilde{\eta}_t \cdot \grad f(\bm{z}^{(t)}) \right)
\end{equation}
for $t=0,1,...$. Here, we call $\tilde{\eta}_t$ the \emph{intrinsic step size} of the gradient ascent method, whose subscript $t$ indicates its potential dependence with respect to the iteration number $t$. While the exponential map on the entire product space $\mathcal{S}_1\times \mathcal{S}_2$ may be difficult to work with, one may resort to the following lemma and express it in terms of the exponential maps on $\mathcal{S}_1$ and $\mathcal{S}_2$.

\begin{lemma}
	\label{exp_map_decomp}
	Given the exponential maps $\Exp_{\bm{x}}:T_{\bm{x}}(\mathcal{S}_1) \to \mathcal{S}_1$, $\Exp_{\bm{y}}:T_{\bm{y}}(\mathcal{S}_2) \to \mathcal{S}_2$, and $\Exp_{\bm{z}}:T_{\bm{x}}(\mathcal{S}_1) \times T_{\bm{y}}(\mathcal{S}_2) \to \mathcal{S}_1\times \mathcal{S}_2$ with $\bm{z}=(\bm{x},\bm{y}) \in \mathcal{S}_1\times \mathcal{S}_2$, we have the following property:
	$$\Exp_{\bm{z}}(\bm{v}) = \left(\Exp_{\bm{x}}(\bm{v}_{\bm{x}}),\, \Exp_{\bm{y}}(\bm{v}_{\bm{y}}) \right),$$
	where $\bm{v}=(\bm{v}_{\bm{x}}, \bm{v}_{\bm{y}}) \in T_{\bm{z}} \equiv T_{\bm{x}}(\mathcal{S}_1) \times T_{\bm{y}}(\mathcal{S}_2)$ and $\bm{v}_{\bm{x}} \in T_{\bm{x}}(\mathcal{S}_1), \bm{v}_{\bm{y}} \in T_{\bm{y}}(\mathcal{S}_2)$.
\end{lemma}

Under the product metric, Lemma~\ref{exp_map_decomp} follows easily from the definition of exponential maps. One can start by verifying that the Riemannian connection $\bm{\bar{\nabla}}$ of $\mathcal{S}_1\times \mathcal{S}_2$ is given by $\bm{\bar{\nabla}}_{Y_1+Y_2}(X_1+X_2) = \bm{\bar{\nabla}}_{Y_1}^{(1)}(X_1) + \bm{\bar{\nabla}}_{Y_2}^{(2)}(X_2)$ with $\bm{\bar{\nabla}}^{(j)}$ being the Riemannian connection of $\mathcal{S}_j$ respectively and $X_j,Y_j$ as differentiable vector fields in $\mathcal{S}_j$ for $j=1,2$; see Exercise 1 of Chapter 6 in \cite{do2013riemannian}. Then, it is justifiable that $\gamma(t)=(\alpha_1(t),\alpha_2(t))$ is the unique geodesic on $\mathcal{S}_1\times \mathcal{S}_2$ when $\alpha_j:[0,1] \to \mathcal{S}_j, j=1,2$ are geodesics on $\mathcal{S}_j$; see Chapter 3 of \cite{doCarmo2016} and Chapter 5 of \cite{lee2006riemannian}. We omit the detailed proof here.

Let $\Pi_j:\mathcal{S}_1\times \mathcal{S}_2 \to \mathcal{S}_j$ for $j=1,2$ be the projection maps from $\mathcal{S}_1\times \mathcal{S}_2$ to $\mathcal{S}_1$ and $\mathcal{S}_2$, respectively. By Lemma~\ref{exp_map_decomp}, the above iterative equation \eqref{GA_manifold} can be written in terms of the gradient ascent methods within $\mathcal{S}_1$ and $\mathcal{S}_2$ as:
\begin{equation}
	\label{GA_manifold_comp}
	\bm{z}^{(t+1)} \gets \left(\Exp_{\bm{x}^{(t)}}\left[\tilde{\eta}_t^{(1)} \cdot \Pi_1\left(\grad f(\bm{z}^{(t)}) \right)\right],\, \Exp_{\bm{y}^{(t)}}\left[\tilde{\eta}_t^{(2)} \cdot \Pi_2\left(\grad f(\bm{z}^{(t)}) \right)\right] \right),
\end{equation}
where $\tilde{\eta}_t^{(j)}, j=1,2$ are the corresponding (componentwise) intrinsic step sizes.

By replacing the Riemannian gradient $\grad f(\bm{z})$ in \eqref{GA_manifold} or \eqref{GA_manifold_comp} with the subspace constrained (Riemannian) gradient $V_d(\bm{z}) V_d(\bm{z})^{\top} \grad f(\bm{z}) = V_d(\bm{z}) V_d(\bm{z})^{\top} \nabla f(\bm{z}),$ 
we obtain the following iterative formula for the subspace constrained gradient ascent method on $\mathcal{S}_1\times\mathcal{S}_2$ as:
\begin{align}
	\label{SCGA}
	\begin{split}
		&\bm{z}^{(t+1)} \\
		&\gets \Exp_{\bm{z}^{(t)}}\left(\tilde{\eta}_t \cdot V_d(\bm{z}) V_d(\bm{z})^{\top} \nabla f(\bm{z}^{(t)}) \right)\\
		&= \left(\Exp_{\bm{x}^{(t)}}\left[\tilde{\eta}_t^{(1)} \cdot V_d^{(1)}(\bm{z}^{(t)}) V_d(\bm{z}^{(t)})^{\top} \nabla f(\bm{z}^{(t)}) \right],\, \Exp_{\bm{y}^{(t)}}\left[\tilde{\eta}_t^{(2)} \cdot V_d^{(2)}(\bm{z}^{(t)}) V_d(\bm{z}^{(t)})^{\top} \nabla f(\bm{z}^{(t)})\right] \right),
	\end{split}
\end{align}
where we write the projection matrix
\begin{equation}
	\label{V_d_exp}
	V_d(\bm{z}) = \left(\bm{v}_{d+1}(\bm{z}),...,\bm{v}_{D_T}(\bm{z}) \right) = \begin{pmatrix}
		V_d^{(1)}(\bm{z})\\
		V_d^{(2)}(\bm{z})
	\end{pmatrix} \in \mathbb{R}^{D_F \times (D_F-d)}
\end{equation}
with $D_F=D_1+\mathbbm{1}_{\{\mathcal{S}_1=\Omega_{D_1}\}} + D_2+ \mathbbm{1}_{\{\mathcal{S}_2=\Omega_{D_2}\}}$, $\bm{v}_{d+1}(\bm{z}),...,\bm{v}_{D_T}(\bm{z})$ are the last $(D_1+D_2-d)$ eigenvectors of $\mathcal{H}f(\bm{z})$ within the tangent space $T_{\bm{z}}$, and $V_d^{(j)}(\bm{z}) \in \mathbb{R}^{D_{F_j}\times (D_{F_j} -d)}$ with $D_{F_j} = D_j+\mathbbm{1}_{\{\mathcal{S}_j=\Omega_{D_j}\}}$ for $j=1,2$.

\subsection{Intrinsic Step Size of the Proposed Mean Shift Algorithms Under the Gradient Ascent Framework on $\mathcal{S}_1\times \mathcal{S}_2$}
\label{Sec:MS_stepsize}

We first derive the intrinsic step size parameters $\tilde{\eta}_t^{(1)}, \tilde{\eta}_t^{(2)}$ of the simultaneous mean shift algorithm on $\mathcal{S}_1\times \mathcal{S}_2$ (Version A) under the above gradient ascent framework. Recall from \autoref{Sec:MS_prod} in the main paper that the simultaneous mean shift algorithm on $\mathcal{S}_1\times \mathcal{S}_2$ iterates the following formula until convergence:
\begin{align}
	\label{MS_iter}
	\begin{split}
		\left(\bm{z}^{(t+1)} \right)^{\top} &=\left(\bm{x}^{(t+1)}, \bm{y}^{(t+1)} \right)^{\top} \\
		&\gets \left(\bm{x}^{(t)}, \bm{y}^{(t)} \right)^{\top} + \Xi(\bm{x}^{(t)}, \bm{y}^{(t)}) = \begin{pmatrix}
			\frac{\sum\limits_{i=1}^n \bm{X}_i k_1'\left(\norm{\frac{\bm{x}^{(t)} -\bm{X}_i}{h_1}}_2^2 \right)  K_2\left(\frac{\bm{y}^{(t)}-\bm{Y}_i}{h_2} \right) }{\sum\limits_{i=1}^n k_1'\left(\norm{\frac{\bm{x}^{(t)} -\bm{X}_i}{h_1}}_2^2 \right) K_2\left(\frac{\bm{y}^{(t)}-\bm{Y}_i}{h_2} \right)}\\ 
			\frac{\sum\limits_{i=1}^n \bm{Y}_i K_1\left(\frac{\bm{x}^{(t)} -\bm{X}_i}{h_1} \right)  k_2'\left(\norm{\frac{\bm{y}^{(t)}-\bm{Y}_i}{h_2}}_2^2 \right) }{\sum\limits_{i=1}^n K_1\left(\frac{\bm{x}^{(t)} -\bm{X}_i}{h_1} \right)  k_2'\left(\norm{\frac{\bm{y}^{(t)}-\bm{Y}_i}{h_2}}_2^2 \right)}
		\end{pmatrix}
	\end{split}
\end{align}
with some additional standardization whenever $\mathcal{S}_1$ and/or $\mathcal{S}_2$ are directional for $t=0,1,...$. Additionally, we remind the reader of the definition of KDE on $\mathcal{S}_1\times \mathcal{S}_2$ as:
\begin{align}
	\label{KDE_prod1}
	\begin{split}
		\hat{f}_{\bm{h}}(\bm{z}) = \hat{f}_{\bm{h}}(\bm{x},\bm{y}) &= \frac{1}{n} \sum_{i=1}^n K_1\left(\frac{\bm{x}-\bm{X}_i}{h_1} \right) K_2\left(\frac{\bm{y}-\bm{Y}_i}{h_2} \right) \\
		&= \frac{\prod_{j=1}^2 C_{k_j,D_j}(h_j)}{n} \sum_{i=1}^n k_1\left(\norm{\frac{\bm{x}-\bm{X}_i}{h_1}}_2^2 \right) k_2\left(\norm{\frac{\bm{y}-\bm{Y}_i}{h_2}}_2^2 \right),
	\end{split}
\end{align}
whose (total) gradient is (c.f. Equation~\eqref{KDE_tot_grad} in the main paper)
\begin{align}
	\label{KDE_tot_grad2}
	\begin{split}
		\nabla \hat{f}_{\bm{h}}(\bm{z}) &= \begin{pmatrix}
			\nabla_{\bm{x}} \hat{f}_{\bm{h}}(\bm{z})\\
			\nabla_{\bm{y}} \hat{f}_{\bm{h}}(\bm{z})
		\end{pmatrix} \\
	&= \frac{2\prod_{j=1}^2 C_{k_j,D_j}(h_j)}{n}\begin{pmatrix}
			\frac{1}{h_1^2} \sum\limits_{i=1}^n (\bm{x}-\bm{X}_i) \cdot k_1'\left(\norm{\frac{\bm{x} -\bm{X}_i}{h_1}}_2^2 \right) k_2\left(\norm{\frac{\bm{y}-\bm{Y}_i}{h_2}}_2^2 \right)\\
			\frac{1}{h_2^2} \sum\limits_{i=1}^n (\bm{y}-\bm{Y}_i) \cdot k_1\left(\norm{\frac{\bm{x} -\bm{X}_i}{h_1}}_2^2 \right) k_2'\left(\norm{\frac{\bm{y}-\bm{Y}_i}{h_2}}_2^2 \right)
		\end{pmatrix} \\
	&= \begin{pmatrix}
			G_{\bm{x}} \left[\frac{\sum\limits_{i=1}^n \bm{X}_i \cdot  k_1'\left(\norm{\frac{\bm{x} -\bm{X}_i}{h_1}}_2^2 \right)  k_2\left(\norm{\frac{\bm{y}-\bm{Y}_i}{h_2}}_2^2 \right) }{\sum\limits_{i=1}^n k_1'\left(\norm{\frac{\bm{x} -\bm{X}_i}{h_1}}_2^2 \right) k_2\left(\norm{\frac{\bm{y}-\bm{Y}_i}{h_2}}_2^2 \right)} -\bm{x} \right]\\
			G_{\bm{y}} \left[\frac{\sum\limits_{i=1}^n \bm{Y}_i \cdot k_1\left(\norm{\frac{\bm{x}-\bm{X}_i}{h_1}}_2^2\right)  k_2'\left(\norm{\frac{\bm{y}-\bm{Y}_i}{h_2}}_2^2 \right) }{\sum\limits_{i=1}^n k_1\left(\norm{\frac{\bm{x}-\bm{X}_i}{h_1}}_2^2\right)  k_2'\left(\norm{\frac{\bm{y}-\bm{Y}_i}{h_2}}_2^2 \right)} -\bm{y} \right]
		\end{pmatrix}
	\end{split}
\end{align}
with 
\begin{align*}
G_{\bm{x}} &= -\frac{2\prod_{j=1}^2 C_{k_j,D_j}(h_j)}{nh_1^2} \sum_{i=1}^n k_1'\left(\norm{\frac{\bm{x}-\bm{X}_i}{h_1}}_2^2\right) k_2\left(\norm{\frac{\bm{y}-\bm{Y}_i}{h_2}}_2^2 \right),\\
G_{\bm{y}} &= -\frac{2\prod_{j=1}^2 C_{k_j,D_j}(h_j)}{nh_2^2} \sum_{i=1}^n k_1\left(\norm{\frac{\bm{x}-\bm{X}_i}{h_1}}_2^2\right) k_2'\left(\norm{\frac{\bm{y}-\bm{Y}_i}{h_2}}_2^2\right).
\end{align*}
Thus, the mean shift vector $\Xi(\bm{z})$ can be written in terms of the (total) gradient estimator $\nabla \hat{f}_{\bm{h}}(\bm{z})$ as:
\begin{equation}
	\label{MS_tot_grad}
	\Xi(\bm{z}) = \begin{pmatrix}
		\nabla_{\bm{x}} \hat{f}_{\bm{h}}(\bm{z}) / G_{\bm{x}}\\
		\nabla_{\bm{y}} \hat{f}_{\bm{h}}(\bm{z}) / G_{\bm{y}}
	\end{pmatrix}.
\end{equation}
We consider the intrinsic step size under three different combinations of the product space.\\

$\bullet$ {\bf Case 1: $\mathcal{S}_1\times \mathcal{S}_2=\mathbb{R}^{D_1}\times \mathbb{R}^{D_2}$}. Given \eqref{MS_tot_grad}, the simultaneous mean shift iterative formula can be written as 
$\bm{z}^{(t+1)} \gets \bm{z}^{(t)} +  \begin{pmatrix}
	\nabla_{\bm{x}} \hat{f}_{\bm{h}}(\bm{z}^{(t)}) / G_{\bm{x}^{(t)}}\\
	\nabla_{\bm{y}} \hat{f}_{\bm{h}}(\bm{z}^{(t)}) / G_{\bm{y}^{(t)}}
\end{pmatrix}.$
Note that under linear kernel profiles, the normalizing constants $C_{k_1,D_1}, C_{k_2,D_2}$ are independent of the bandwidth parameters $h_1,h_2$. Hence, it is natural to see that the intrinsic step size parameters $\tilde{\eta}_t^{(1)}, \tilde{\eta}_t^{(2)}$ are
\begin{align*}
	\tilde{\eta}_t^{(1)} &= \frac{1}{G_{\bm{x}^{(t)}}} = -\frac{h_1^2}{\frac{2\prod_{j=1}^2 C_{k_j,D_j}}{n} \sum_{i=1}^n k_1'\left(\norm{\frac{\bm{x}^{(t)}-\bm{X}_i}{h_1}}_2^2\right) k_2\left(\norm{\frac{\bm{y}^{(t)}-\bm{Y}_i}{h_2}}_2^2 \right)},\\
	\tilde{\eta}_t^{(2)} &= \frac{1}{G_{\bm{y}^{(t)}}} = -\frac{h_2^2}{\frac{2\prod_{j=1}^2 C_{k_j,D_j}}{n} \sum_{i=1}^n k_1\left(\norm{\frac{\bm{x}^{(t)}-\bm{X}_i}{h_1}}_2^2\right) k_2'\left(\norm{\frac{\bm{y}^{(t)}-\bm{Y}_i}{h_2}}_2^2 \right)}.
\end{align*}
The following Lemma~\ref{Thm:MS_Eu_rate} elucidates the asymptotic rates of $\tilde{\eta}_t^{(1)}, \tilde{\eta}_t^{(2)}$ in terms of the bandwidth parameters.

\begin{lemma}
	\label{Thm:MS_Eu_rate}
	Assume conditions (A1-2) and $\mathcal{S}_1\times \mathcal{S}_2=\mathbb{R}^{D_1}\times \mathbb{R}^{D_2}$. Then, when $h_1,h_2 \lesssim h \to 0$ and $nh^{D_1+D_2} \to \infty$, we have that
	\begin{align*}
		h_1^2\cdot G_{\bm{x}} &=-\frac{2\prod_{j=1}^2 C_{k_j,D_j}}{n} \sum_{i=1}^n k_1'\left(\norm{\frac{\bm{x}-\bm{X}_i}{h_1}}_2^2\right) k_2\left(\norm{\frac{\bm{y}-\bm{Y}_i}{h_2}}_2^2 \right) \\
		&= -2\prod_{j=1}^2 C_{k_j,D_j} \cdot f(\bm{z}) \int_{\mathcal{S}_1\times \mathcal{S}_2} k_1'\left(\norm{\bm{x}}_2^2 \right) k_2\left(\norm{\bm{y}}_2^2 \right)\, d\bm{x} d\bm{y} + O(h^2) + O_P\left(\sqrt{\frac{1}{nh^{D_1+D_2}}} \right)\\
		&= O(1) + O(h^2) + O_P\left(\sqrt{\frac{1}{nh^{D_1+D_2}}} \right),
	\end{align*}
	and similarly, 
	\begin{align*}
		h_2^2\cdot G_{\bm{y}} &=-\frac{2\prod_{j=1}^2 C_{k_j,D_j}}{n} \sum_{i=1}^n k_1\left(\norm{\frac{\bm{x}-\bm{X}_i}{h_1}}_2^2\right) k_2'\left(\norm{\frac{\bm{y}-\bm{Y}_i}{h_2}}_2^2 \right) \\
		&= -2\prod_{j=1}^2 C_{k_j,D_j} \cdot f(\bm{z}) \int_{\mathcal{S}_1\times \mathcal{S}_2} k_1\left(\norm{\bm{x}}_2^2 \right) k_2'\left(\norm{\bm{y}}_2^2 \right)\, d\bm{x} d\bm{y} + O(h^2) + O_P\left(\sqrt{\frac{1}{nh^{D_1+D_2}}} \right)\\
		&= O(1)+ O(h^2) + O_P\left(\sqrt{\frac{1}{nh^{D_1+D_2}}} \right).
	\end{align*}
\end{lemma}

The proof of Lemma~\ref{Thm:MS_Eu_rate} is similar to Theorem 1 in \cite{MS1995} and Lemma 3.2 in \cite{DirSCMS2021}, which makes use of the Taylor's theorem; see the proof of Proposition~\ref{Thm:trans_grad_conv} in \autoref{Sec:Conv_pf} for a sketch. It implies that $\tilde{\eta}_t^{(1)}, \tilde{\eta}_t^{(2)} =O(h^2)$ as the bandwidths $h_1,h_2 \lesssim h \to 0$ and $nh^{D_1+D_2} \to \infty$.\\

\begin{figure}
	\centering
	\includegraphics[width=0.45\linewidth]{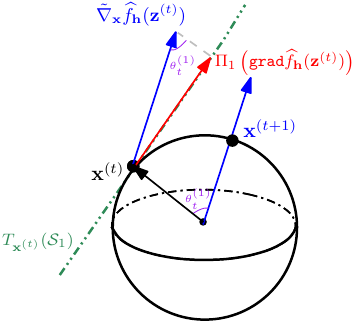}
	\caption{One-step iteration of the (simultaneous) mean shift algorithm on $\Omega_{D_1}\times \mathbb{R}^{D_2}$ projected onto $\Omega_{D_1}$.}
	\label{fig:MS_one_step}
\end{figure}

$\bullet$ {\bf Case 2: $\mathcal{S}_1\times \mathcal{S}_2=\Omega_{D_1}\times \mathbb{R}^{D_2}$}. Apart from updating the sequence $\left\{\bm{z}^{(t)} \right\}_{t=0}^{\infty} = \left\{(\bm{x}^{(t)}, \bm{y}^{(t)}) \right\}_{t=0}^{\infty}$ with \eqref{MS_iter}, we need to standardize $\bm{x}^{(t+1)}$ as $\frac{\bm{x}^{(t+1)}}{\norm{\bm{x}^{(t+1)}}_2} \in \Omega_{D_1}$. Hence, a complete mean shift iterative step reads
\begin{align*}
	\bm{x}^{(t+1)} &\gets -\frac{\sum\limits_{i=1}^n \bm{X}_i \cdot  k_1'\left(\norm{\frac{\bm{x}^{(t)} -\bm{X}_i}{h_1}}_2^2 \right)  k_2\left(\norm{\frac{\bm{y}^{(t)}-\bm{Y}_i}{h_2}}_2^2 \right) }{\norm{\sum\limits_{i=1}^n \bm{X}_i \cdot  k_1'\left(\norm{\frac{\bm{x}^{(t)} -\bm{X}_i}{h_1}}_2^2 \right)  k_2\left(\norm{\frac{\bm{y}^{(t)}-\bm{Y}_i}{h_2}}_2^2 \right)}_2} \\
	&= -\frac{\sum\limits_{i=1}^n \bm{X}_i \cdot  L'\left(\frac{1-\bm{X}_i^{\top}\bm{x}^{(t)}}{h_1^2} \right)  k_2\left(\norm{\frac{\bm{y}^{(t)}-\bm{Y}_i}{h_2}}_2^2 \right) }{\norm{\sum\limits_{i=1}^n \bm{X}_i \cdot  L'\left(\frac{1-\bm{X}_i^{\top}\bm{x}^{(t)}}{h_1^2} \right)  k_2\left(\norm{\frac{\bm{y}^{(t)}-\bm{Y}_i}{h_2}}_2^2 \right)}_2},\\
	\bm{y}^{(t+1)} &\gets \frac{\sum\limits_{i=1}^n \bm{Y}_i \cdot k_1\left(\norm{\frac{\bm{x}^{(t)}-\bm{X}_i}{h_1}}_2^2\right)  k_2'\left(\norm{\frac{\bm{y}^{(t)}-\bm{Y}_i}{h_2}}_2^2 \right) }{\sum\limits_{i=1}^n k_1\left(\norm{\frac{\bm{x}^{(t)}-\bm{X}_i}{h_1}}_2^2\right)  k_2'\left(\norm{\frac{\bm{y}^{(t)}-\bm{Y}_i}{h_2}}_2^2 \right)} \\
	& = \bm{y}^{(t)} + \frac{1}{G_{\bm{y}^{(t)}}} \cdot \nabla_{\bm{y}} \hat{f}_{\bm{h}}(\bm{z}^{(t)})
\end{align*}
with the directional kernel profile $k_1\left(\norm{\bm{u}}_2^2 \right) = L\left(\frac{1}{2}\norm{\bm{u}}_2^2 \right)$.
Then, one can observe that when $h_1,h_2 \lesssim h$ are small,
\begin{align*}
	\tilde{\eta}_t^{(2)} &= \frac{1}{G_{\bm{y}^{(t)}}} = - \frac{h_2^2}{\frac{2C_{k_1,D_1}(h_1) \cdot C_{k_2,D_2}}{n} \sum\limits_{i=1}^n k_1\left(\norm{\frac{\bm{x}^{(t)}-\bm{X}_i}{h_1}}_2^2\right) k_2'\left(\norm{\frac{\bm{y}^{(t)}-\bm{Y}_i}{h_2}}_2^2 \right)} = O(h^2).
\end{align*}
To derive $\tilde{\eta}_t^{(1)}$ and its asymptotic rate, we denote the angle between $\bm{x}^{(t+1)}$ and $\bm{x}^{(t)}$ by $\theta_t^{(1)}$. Under the gradient ascent framework \eqref{GA_manifold_comp} (projected to $\mathcal{S}_1$), we have the following equality for the geodesic distance $d_g(\bm{x}^{(t+1)},\bm{x}^{(t)})$ on $\Omega_{D_1}$ as:
\begin{align*}
	\arccos\left(\left(\bm{x}^{(t+1)}\right)^{\top} \bm{x}^{(t)} \right) = \theta_t^{(1)} &= \tilde{\eta}_t^{(1)} \norm{\Pi_1\left(\grad \hat{f}_{\bm{h}}(\bm{z}^{(t)})\right)}_2 = \tilde{\eta}_t^{(1)} \cdot \tilde{\nabla}_{\bm{x}} \hat{f}_{\bm{h}}(\bm{z}^{(t)}) \cdot \sin \theta_t^{(1)},
\end{align*}
where 
$$\tilde{\nabla}_{\bm{x}} \hat{f}_{\bm{h}}(\bm{z}) = -\frac{C_{k_1,D_1}(h_1) \cdot C_{k_2,D_2}}{nh_1^2 h_2^{D_2}} \sum\limits_{i=1}^n \bm{X}_i \cdot  L'\left(\frac{1-\bm{X}_i^{\top}\bm{x}}{h_1^2} \right)  k_2\left(\norm{\frac{\bm{y}-\bm{Y}_i}{h_2}}_2^2 \right).$$ 
See \autoref{fig:MS_one_step} for a graphical illustration. Indeed, $\tilde{\nabla}_{\bm{x}} \hat{f}_{\bm{h}}(\bm{z})$ is also a total gradient of $\hat{f}_{\bm{h}}(\bm{z})$ within $\mathcal{S}_1$ when $\mathcal{S}_1\times \mathcal{S}_2=\Omega_{D_1}\times \mathbb{R}^{D_2}$. Thus, the intrinsic step size on $\Omega_{D_1}$ is
$$\tilde{\eta}_t^{(1)} = \frac{\theta_t^{(1)}}{\tilde{\nabla}_{\bm{x}} \hat{f}_{\bm{h}}(\bm{z}^{(t)}) \cdot \sin \theta_t^{(1)}}.$$
As $t\to \infty$ and the initial point $\bm{z}^{(0)}$ is near an estimated local mode $\hat{\bm{m}} \in \hat{\mathcal{M}}$, we know that $\theta_t^{(1)} \to 0$ and $\frac{\theta_t^{(1)}}{\sin \theta_t^{(1)}} \to 1$ will be upper bounded. Therefore, the intrinsic step size $\tilde{\eta}_t^{(1)}$ is essentially determined by $\tilde{\nabla}_{\bm{x}} \hat{f}_{\bm{h}}(\bm{z}^{(t)})$. The following Lemma~\ref{Thm:MS_Dir_rate} illuminates the asymptotic rate of the intrinsic step size $\tilde{\eta}_t^{(1)}$.

\begin{lemma}
	\label{Thm:MS_Dir_rate}
	Assume conditions (A1-2). Then, for any fixed $\bm{z}=(\bm{x},\bm{y}) \in \mathcal{S}_1\times \mathcal{S}_2=\Omega_{D_1} \times \mathbb{R}^{D_2}$, we have that
	$$h_1^2 \cdot \tilde{\nabla}_{\bm{x}} \hat{f}_{\bm{h}}(\bm{z}) = \bm{x} f(\bm{z}) C_{L,D_1} + o(1) + O_P\left(\sqrt{\frac{1}{nh^{D_1+D_2}}} \right)$$
	as $h_1,h_2 \lesssim h \to 0$ and $nh^{D_1+D_2} \to \infty$, where $C_{L,D_1} = -\frac{\int_0^{\infty} L'(r) r^{\frac{D_1}{2}-1} dr}{\int_0^{\infty} L(r) r^{\frac{D_1}{2}-1} dr} >0$ is a constant depending exclusively on the directional kernel profile $L$ and dimension $D_1$ of $\Omega_{D_1}$.
\end{lemma}

The proof of Lemma~\ref{Thm:MS_Dir_rate} is almost identical to Lemma 10 in \cite{DirMS2020} and is thus omitted. It indicates that the limiting behavior of $\tilde{\nabla}_{\bm{x}} \hat{f}_{\bm{h}}(\bm{z})$ will approach the radial direction $\bm{x}$ on $\mathcal{S}_1=\Omega_{D_1}$. More importantly, the result teaches us that $\tilde{\eta}_t^{(1)} = O(h^2)$ as the bandwidths $h_1,h_2 \lesssim h$ are small and the sample size $n$ is large.\\

$\bullet$ {\bf Case 3: $\mathcal{S}_1\times \mathcal{S}_2=\Omega_{D_1}\times \Omega_{D_2}$}. Given the mean shift iteration \eqref{MS_iter} and extra standardizations $\frac{\bm{x}^{(t+1)}}{\norm{\bm{x}^{(t+1)}}_2}$ and $\frac{\bm{y}^{(t+1)}}{\norm{\bm{y}^{(t+1)}}_2}$, an entire mean shift iterative step on $\Omega_{D_1}\times \Omega_{D_2}$ updates the sequence $\left\{\bm{z}^{(t)} \right\}_{t=0}^{\infty} = \left\{(\bm{x}^{(t)}, \bm{y}^{(t)}) \right\}_{t=0}^{\infty}$ as:
\begin{align*}
	\bm{x}^{(t+1)} &\gets -\frac{\sum\limits_{i=1}^n \bm{X}_i \cdot  k_1'\left(\norm{\frac{\bm{x}^{(t)} -\bm{X}_i}{h_1}}_2^2 \right)  k_2\left(\norm{\frac{\bm{y}^{(t)}-\bm{Y}_i}{h_2}}_2^2 \right) }{\norm{\sum\limits_{i=1}^n \bm{X}_i \cdot  k_1'\left(\norm{\frac{\bm{x}^{(t)} -\bm{X}_i}{h_1}}_2^2 \right)  k_2\left(\norm{\frac{\bm{y}^{(t)}-\bm{Y}_i}{h_2}}_2^2 \right)}_2} \\
	&= -\frac{\sum\limits_{i=1}^n \bm{X}_i \cdot  L_1'\left(\frac{1-\bm{X}_i^{\top}\bm{x}^{(t)}}{h_1^2} \right)  L_2\left(\frac{1-\bm{Y}_i^{\top}\bm{y}^{(t)}}{h_2^2} \right) }{\norm{\sum\limits_{i=1}^n \bm{X}_i \cdot  L_1'\left(\frac{1-\bm{X}_i^{\top}\bm{x}^{(t)}}{h_1^2} \right)  L_2\left(\frac{1-\bm{Y}_i^{\top}\bm{y}^{(t)}}{h_2^2} \right)}_2},\\
	\bm{y}^{(t+1)} &\gets -\frac{\sum\limits_{i=1}^n \bm{Y}_i \cdot  k_1\left(\norm{\frac{\bm{x}^{(t)} -\bm{X}_i}{h_1}}_2^2 \right)  k_2'\left(\norm{\frac{\bm{y}^{(t)}-\bm{Y}_i}{h_2}}_2^2 \right) }{\norm{\sum\limits_{i=1}^n \bm{Y}_i \cdot  k_1\left(\norm{\frac{\bm{x}^{(t)} -\bm{X}_i}{h_1}}_2^2 \right)  k_2'\left(\norm{\frac{\bm{y}^{(t)}-\bm{Y}_i}{h_2}}_2^2 \right)}_2} \\
	&= -\frac{\sum\limits_{i=1}^n \bm{Y}_i \cdot  L_1\left(\frac{1-\bm{X}_i^{\top}\bm{x}^{(t)}}{h_1^2} \right)  L_2'\left(\frac{1-\bm{Y}_i^{\top}\bm{y}^{(t)}}{h_2^2} \right) }{\norm{\sum\limits_{i=1}^n \bm{Y}_i \cdot  L_1\left(\frac{1-\bm{X}_i^{\top}\bm{x}^{(t)}}{h_1^2} \right)  L_2'\left(\frac{1-\bm{Y}_i^{\top}\bm{y}^{(t)}}{h_2^2} \right)}_2}
\end{align*}
with the directional kernel profile $k_j\left(\norm{\bm{u}}_2^2 \right) = L_j\left(\frac{1}{2}\norm{\bm{u}}_2^2 \right)$ for $j=1,2$. Let $\theta_t^{(1)}$ be the angle between $\bm{x}^{(t+1)}$ and $\bm{x}^{(t)}$ as well as $\theta_t^{(2)}$ be the angle between $\bm{y}^{(t+1)}$ and $\bm{y}^{(t)}$. If we denote 
\begin{align*}
	\tilde{\nabla}_{\bm{x}}\hat{f}_{\bm{h}}(\bm{z}) &= -\frac{\prod_{j=1}^2 C_{k_j,D_j}(h_j)}{n} \sum\limits_{i=1}^n \bm{X}_i \cdot  L_1'\left(\frac{1-\bm{X}_i^{\top}\bm{x}^{(t)}}{h_1^2} \right)  L_2\left(\frac{1-\bm{Y}_i^{\top}\bm{y}^{(t)}}{h_2^2} \right),\\
	\tilde{\nabla}_{\bm{y}} \hat{f}_{\bm{h}}(\bm{z}) &= -\frac{\prod_{j=1}^2 C_{k_j,D_j}(h_j)}{n} \sum\limits_{i=1}^n \bm{Y}_i \cdot  L_1\left(\frac{1-\bm{X}_i^{\top}\bm{x}^{(t)}}{h_1^2} \right)  L_2'\left(\frac{1-\bm{Y}_i^{\top}\bm{y}^{(t)}}{h_2^2} \right),
\end{align*}
then the arguments in {\bf Case 2} can be adopted to show that the intrinsic step sizes are
$$\tilde{\eta}_t^{(1)} = \frac{\theta_t^{(1)}}{\tilde{\nabla}_{\bm{x}}\hat{f}_{\bm{h}}(\bm{z}^{(t)}) \cdot \sin \theta_t^{(1)}}\quad \text{ and } \quad \tilde{\eta}_t^{(2)} = \frac{\theta_t^{(2)}}{\tilde{\nabla}_{\bm{y}}\hat{f}_{\bm{h}}(\bm{z}^{(t)}) \cdot \sin \theta_t^{(2)}}.$$%
Therefore, by Lemma~\ref{Thm:MS_Dir_rate}, $\tilde{\eta}_t^{(j)}=O(h^2)$ for $j=1,2$ when bandwidths $h_1,h_2\lesssim h$ are small and the sample size $n$ is large.\\

We have discussed the intrinsic step sizes of the simultaneous mean shift algorithm (Version A) under the gradient ascent framework \eqref{GA_manifold_comp}. One can follow our preceding arguments to derive the intrinsic step sizes of the componentwise mean shift algorithm (Version B) as:
\begin{itemize}
	\item {\bf Case 1: } When $\mathcal{S}_1\times \mathcal{S}_2 = \mathbb{R}^{D_1} \times \mathbb{R}^{D_2}$,
	\begin{align*}
		\tilde{\eta}_t^{(1)} &= -\frac{nh_1^2}{2\prod\limits_{j=1}^2 C_{k_j,D_j} \sum\limits_{i=1}^n k_1'\left(\norm{\frac{\bm{x}^{(t)}-\bm{X}_i}{h_1}}_2^2\right) k_2\left(\norm{\frac{\bm{y}^{(t)}-\bm{Y}_i}{h_2}}_2^2 \right)} \\
		\text{ and } & \quad 
		\tilde{\eta}_t^{(2)} = -\frac{nh_2^2}{2\prod\limits_{j=1}^2 C_{k_j,D_j} \sum\limits_{i=1}^n k_1\left(\norm{\frac{\bm{x}^{(t+1)}-\bm{X}_i}{h_1}}_2^2\right) k_2'\left(\norm{\frac{\bm{y}^{(t)}-\bm{Y}_i}{h_2}}_2^2 \right)}.
	\end{align*}
	
	\item {\bf Case 2: } When $\mathcal{S}_1\times \mathcal{S}_2 = \Omega_{D_1} \times \mathbb{R}^{D_2}$, 
	\begin{align*}
		\tilde{\eta}_t^{(1)} &= \frac{\theta_t^{(1)}}{\tilde{\nabla}_{\bm{x}}\hat{f}_{\bm{h}}(\bm{x}^{(t)}, \bm{y}^{(t)}) \cdot \sin \theta_t^{(1)}} \\
		\text{ and } & \quad
		\tilde{\eta}_t^{(2)} = -\frac{nh_2^2}{2\prod\limits_{j=1}^2 C_{k_j,D_j} \sum\limits_{i=1}^n k_1\left(\norm{\frac{\bm{x}^{(t+1)}-\bm{X}_i}{h_1}}_2^2\right) k_2'\left(\norm{\frac{\bm{y}^{(t)}-\bm{Y}_i}{h_2}}_2^2 \right)}.
	\end{align*}
	
	\item {\bf Case 3: } When $\mathcal{S}_1\times \mathcal{S}_2 = \Omega_{D_1}\times \Omega_{D_2}$,
	$$\tilde{\eta}_t^{(1)} = \frac{\theta_t^{(1)}}{\tilde{\nabla}_{\bm{x}}\hat{f}_{\bm{h}}(\bm{x}^{(t)}, \bm{y}^{(t)}) \cdot \sin \theta_t^{(1)}}\quad \text{ and } \quad \tilde{\eta}_t^{(2)} = \frac{\theta_t^{(2)}}{\tilde{\nabla}_{\bm{y}}\hat{f}_{\bm{h}}(\bm{x}^{(t+1)},\bm{y}^{(t)}) \cdot \sin \theta_t^{(2)}}.$$
\end{itemize}
Notice that each full iteration of the componentwise mean shift algorithm (Version B) iterates the gradient ascent method \eqref{GA_manifold_comp} twice. Finally, analogous to the simultaneous mean shift algorithm, the intrinsic step sizes of the componentwise mean shift algorithm have their asymptotic rates $\tilde{\eta}_t^{(j)} = O(h^2)$ for $j=1,2$ when the bandwidths $h_1,h_2\lesssim h$ are small and the sample size $n$ is large.

\subsection{Intrinsic Step Size of the Proposed SCMS Algorithm Under the Subspace Constrained Gradient Ascent Framework on $\mathcal{S}_1\times \mathcal{S}_2$}
\label{Sec:stepsize}

Recall from \autoref{Sec:SCMS_prod} in the main paper that our proposed SCMS algorithm \eqref{scaled_SCMS} on $\mathcal{S}_1\times \mathcal{S}_2$ with Gaussian and/or von Mises kernels iterates the following equation:
\begin{equation*}
	\bm{z}^{(t+1)} \gets \bm{z}^{(t)} + \eta\cdot \hat{V}_d(\bm{z}^{(t)}) \hat{V}_d(\bm{z}^{(t)})^{\top} \bm{H}^{-1} \Xi(\bm{z}^{(t)})
\end{equation*}
with an additional standardization whenever $\mathcal{S}_1$ and/or $\mathcal{S}_2$ are directional for $t=0,1,...$, where $\eta >0$ is the step size (i.e., a tuning parameter). We now derive the intrinsic step size parameter $\tilde{\eta}_t$ (or $\tilde{\eta}_t^{(1)}, \tilde{\eta}_t^{(2)}$) of the SCMS algorithm \eqref{scaled_SCMS} under the subspace constrained gradient ascent framework \eqref{SCGA} on $\mathcal{S}_1\times \mathcal{S}_2$. There are three different cases based on the configurations of $\mathcal{S}_1\times\mathcal{S}_2$.\\

$\bullet$ {\bf Case 1: $\mathcal{S}_1\times \mathcal{S}_2=\mathbb{R}^{D_1}\times \mathbb{R}^{D_2}$}. Some simple algebra show that $\bm{H}^{-1} \Xi(\bm{z}) = \frac{\nabla \hat{f}_{\bm{h}}(\bm{z})}{\hat{f}_{\bm{h}}(\bm{z})}$
under Gaussian and/or von Mises kernels; see also \cite{chacon2013data,chacon2013comparison}. Thus, the iterative equation \eqref{scaled_SCMS} becomes
\begin{equation}
	\label{scale_SCMS_grad}
	\bm{z}^{(t+1)} \gets \bm{z}^{(t)} + \eta \hat{V}_d(\bm{z}^{(t)}) \hat{V}_d(\bm{z}^{(t)})^{\top} \cdot \frac{\nabla\hat{f}_{\bm{h}}(\bm{z}^{(t)})}{\hat{f}_{\bm{h}}(\bm{z}^{(t)})},
\end{equation}
so it is intuitive to see that the intrinsic step size is $\tilde{\eta}_t = \frac{\eta}{\hat{f}_{\bm{h}}(\bm{z}^{(t)})}$.\\

$\bullet$ {\bf Case 2: $\mathcal{S}_1\times \mathcal{S}_2=\Omega_{D_1}\times \mathbb{R}^{D_2}$}. Besides updating the sequence $\left\{\bm{z}^{(t)} \right\}_{t=0}^{\infty} = \left\{(\bm{x}^{(t)}, \bm{y}^{(t)}) \right\}_{t=0}^{\infty}$ with \eqref{scale_SCMS_grad}, we need to standardize $\bm{x}^{(t+1)}$ as $\frac{\bm{x}^{(t+1)}}{\norm{\bm{x}^{(t+1)}}_2}$. Hence, a complete SCMS iterative step reads
\begin{align*}
	\bm{x}^{(t+1)} &\gets \frac{\hat{f}_{\bm{h}}(\bm{z}^{(t)}) \bm{x}^{(t)} + \eta \hat{V}_d^{(1)}(\bm{z}^{(t)}) \hat{V}_d(\bm{z}^{(t)})^{\top} \nabla \hat{f}_{\bm{h}}(\bm{z}^{(t)})}{\norm{\hat{f}_{\bm{h}}(\bm{z}^{(t)}) \bm{x}^{(t)} + \eta \hat{V}_d^{(1)}(\bm{z}^{(t)}) \hat{V}_d(\bm{z}^{(t)})^{\top} \nabla \hat{f}_{\bm{h}}(\bm{z}^{(t)})}_2}, \\ 
	\bm{y}^{(t+1)} &\gets \bm{y}^{(t)} + \frac{\eta}{\hat{f}_{\bm{h}}(\bm{z}^{(t)})} \cdot \hat{V}_d^{(2)}(\bm{z}^{(t)}) \hat{V}_d(\bm{z}^{(t)})^{\top} \nabla \hat{f}_{\bm{h}}(\bm{z}^{(t)}),
\end{align*}
where we decompose $\hat{V}_d(\bm{z})$ into two parts $\hat{V}_d^{(j)}(\bm{z}), j=1,2$ within $\mathcal{S}_1$ and $\mathcal{S}_2$ respectively as in \eqref{V_d_exp}. Compared with \eqref{SCGA}, it is natural to obtain that $\tilde{\eta}_t^{(2)} = \frac{\eta}{\hat{f}_{\bm{h}}(\bm{z}^{(t)})}$. 

\begin{figure}
	\centering
	\includegraphics[width=0.65\linewidth]{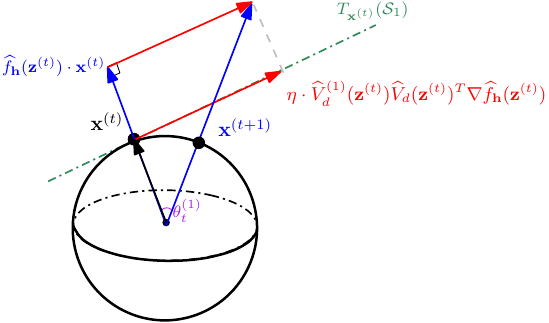}
	\caption{One-step iteration of the proposed SCMS algorithm on $\Omega_{D_1}\times \mathbb{R}^{D_2}$ projected onto $\Omega_{D_1}$.}
	\label{fig:SCMS_one_step}
\end{figure}

As for $\tilde{\eta}_t^{(1)}$ within $\mathcal{S}_1=\Omega_{D_1}$, we first notice that $\bm{x}$ is orthogonal to the columns of $\hat{V}_d^{(1)}(\bm{z})$ for any $\bm{z}=(\bm{x},\bm{y})\in \mathcal{S}_1\times \mathcal{S}_2$. This is because $(\bm{x},\bm{0}) \in \mathcal{S}_1\times \mathcal{S}_2$ is a unit eigenvector of the Riemannian Hessian $\mathcal{H}\hat{f}_{\bm{h}}(\bm{z})$ that is normal to the tangent space $T_{\bm{z}} = T_{\bm{x}}(\mathcal{S}_1) \times T_{\bm{y}}(\mathcal{S}_2)$. Given that the columns of $\hat{V}_d(\bm{z})$ consist of the last $(D_1+D_2-d)$ eigenvectors of $\mathcal{H}\hat{f}_{\bm{h}}(\bm{z})$ within the tangent space $T_{\bm{z}}$, it follows that $\bm{x}^{\top} \hat{V}_d^{(1)}(\bm{z}) = \bm{0} \in \mathbb{R}^{1\times (D_1+D_2-d)}$; see \autoref{fig:SCMS_one_step} for a graphical illustration.
If we denote the angle between $\bm{x}^{(t+1)}$ and $\bm{x}^{(t)}$ by $\theta_t^{(1)}$, then we have the following equality for the geodesic distance $d_g(\bm{x}^{(t)},\bm{x}^{(t+1)})$ on $\Omega_{D_1}$ as:
\begin{align*}
	\theta_t^{(1)} &= \tilde{\eta}_t^{(1)} \norm{\hat{V}_d^{(1)}(\bm{z}^{(t)}) \hat{V}_d(\bm{z}^{(t)})^{\top} \nabla \hat{f}_{\bm{h}}(\bm{z}^{(t)})}_2 = \frac{\tilde{\eta}_t^{(1)}}{\eta} \cdot\hat{f}_{\bm{h}}(\bm{z}^{(t)}) \tan\theta_t^{(1)},
\end{align*}
where we leverage the orthogonality between $\bm{x}^{(t)}$ and the columns of $\hat{V}_d^{(1)}(\bm{z}^{(t)})$ to obtain the second equality; see \autoref{fig:SCMS_one_step}. Thus, the intrinsic step on $\Omega_{D_1}$ is
$\tilde{\eta}_t^{(1)} = \frac{\eta}{\hat{f}_{\bm{h}}(\bm{z}^{(t)})} \cdot \frac{\theta_t^{(1)}}{\tan\theta_t^{(1)}}.$

$\bullet$ {\bf Case 3: $\mathcal{S}_1\times \mathcal{S}_2=\Omega_{D_1}\times \Omega_{D_2}$}. Under the iteration \eqref{scale_SCMS_grad} and extra standardizations $\frac{\bm{x}^{(t+1)}}{\norm{\bm{x}^{(t+1)}}_2}$ and $\frac{\bm{y}^{(t+1)}}{\norm{\bm{y}^{(t+1)}}_2}$, a complete SCMS iteration on $\Omega_{D_1}\times \Omega_{D_2}$ updates the sequence $\left\{\bm{z}^{(t)} \right\}_{t=0}^{\infty} = \left\{(\bm{x}^{(t)}, \bm{y}^{(t)}) \right\}_{t=0}^{\infty}$ as:
\begin{align*}
	\bm{x}^{(t+1)} &\gets \frac{\hat{f}_{\bm{h}}(\bm{z}^{(t)}) \bm{x}^{(t)} + \eta \hat{V}_d^{(1)}(\bm{z}^{(t)}) \hat{V}_d(\bm{z}^{(t)})^{\top} \nabla \hat{f}_{\bm{h}}(\bm{z}^{(t)})}{\norm{\hat{f}_{\bm{h}}(\bm{z}^{(t)}) \bm{x}^{(t)} + \eta \hat{V}_d^{(1)}(\bm{z}^{(t)}) \hat{V}_d(\bm{z}^{(t)})^{\top} \nabla \hat{f}_{\bm{h}}(\bm{z}^{(t)})}_2} \\ 
	\text{ and } &\quad \bm{y}^{(t+1)} \gets \frac{\hat{f}_{\bm{h}}(\bm{z}^{(t)}) \bm{y}^{(t)} + \eta \hat{V}_d^{(2)}(\bm{z}^{(t)}) \hat{V}_d(\bm{z}^{(t)})^{\top} \nabla \hat{f}_{\bm{h}}(\bm{z}^{(t)})}{\norm{\hat{f}_{\bm{h}}(\bm{z}^{(t)}) \bm{y}^{(t)} + \eta \hat{V}_d^{(2)}(\bm{z}^{(t)}) \hat{V}_d(\bm{z}^{(t)})^{\top} \nabla \hat{f}_{\bm{h}}(\bm{z}^{(t)})}_2},
\end{align*}
where we again leverage the decomposition of $\hat{V}_d(\bm{z})$ in \eqref{V_d_exp}. Moreover, since $(\bm{x},\bm{0}), (\bm{0},\bm{y}) \in \mathcal{S}_1\times \mathcal{S}_2$ are unit eigenvectors of the Riemannian Hessian $\mathcal{H} \hat{f}_{\bm{h}}(\bm{x})$ that are orthogonal to the tangent space $T_{\bm{x}}(\mathcal{S}_1)\times T_{\bm{y}}(\mathcal{S}_2)$, the same argument in {\bf Case 2} implies that $\bm{x}^{\top}\hat{V}_d^{(1)}(\bm{z}) = \bm{y}^{\top} \hat{V}_d(\bm{z}) = \bm{0}$. Let $\theta_t^{(1)}$ be the angle between $\bm{x}^{(t)}$ and $\bm{x}^{(t+1)}$, and denote the angle between $\bm{y}^{(t)}$ and $\bm{y}^{(t+1)}$ by $\theta_t^{(2)}$. Thus, based on the arguments in {\bf Case 2}, we know that the intrinsic steps on $\Omega_{D_1}$ and $\Omega_{D_2}$ are
$\tilde{\eta}_t^{(1)} = \frac{\eta}{\hat{f}_{\bm{h}}(\bm{z}^{(t)})} \cdot \frac{\theta_t^{(1)}}{\tan\theta_t^{(1)}} \text{ and } \tilde{\eta}_t^{(2)} = \frac{\eta}{\hat{f}_{\bm{h}}(\bm{z}^{(t)})} \cdot \frac{\theta_t^{(2)}}{\tan\theta_t^{(2)}}.$

Our above derivations indicate that the intrinsic step sizes $\tilde{\eta}_t$ or $\tilde{\eta}_t^{(j)},j=1,2$ of our proposed SCMS algorithm \eqref{scaled_SCMS} on $\mathcal{S}_1\times \mathcal{S}_2$ under the subspace constrained gradient ascent framework \eqref{SCGA} embrace the following bounds near $\hat{\mathcal{R}}_d$:
\begin{equation}
	\label{intrinsic_step_size_bound}
	\frac{\tilde{A} \eta}{\hat{f}_{\bm{h}}(\bm{z}^{(t)})} \leq \tilde{\eta}_t \text{ or } \tilde{\eta}_t^{(1)}, \tilde{\eta}_t^{(2)} \leq \frac{\eta}{\hat{f}_{\bm{h}}(\bm{z}^{(t)})},
\end{equation}
because $\frac{\theta_t^{(j)}}{\tan \theta_t^{(j)}} \leq 1$ but $\frac{\theta_t^{(j)}}{\tan \theta_t^{(j)}} \to 1$ as $\theta_t^{(j)} \to 0$ for $j=1,2$, where $\tilde{A}$ is a constant that is independent of $t$ and bandwidth $\bm{h}=(h_1,h_2)$. Therefore, if we set the tuning parameter $\eta=1$ in our proposed SCMS algorithm as in the standard/naive SCMS iteration, the intrinsic step sizes have a lower bound around $\hat{\mathcal{R}}_d$:
$$\tilde{\eta}_t \text{ or } \tilde{\eta}_t^{(1)}, \tilde{\eta}_t^{(2)} \geq \frac{\tilde{A}}{\hat{f}_{\bm{h}}(\bm{z}^{(t)})} \geq \frac{\tilde{A}}{\hat{f}_{\bm{h}}(\hat{\bm{m}})}$$
for some $\hat{\bm{m}} \in \hat{\mathcal{M}}$. Asymptotically, when the bandwidths $h_1,h_2 \lesssim h$ are small and the sample size $n$ is large, we obtain by Lemma~\ref{pw_rate_KDE} of $\hat{f}_{\bm{h}}$ that
\begin{align*}
	\mathcal{P}_{\bm{z}}\bm{H}^{-1} \Xi(\bm{z}) &= \frac{\grad \hat{f}_{\bm{h}}(\bm{z})}{\hat{f}_{\bm{h}}(\bm{z})} = \frac{\grad f(\bm{z}) + O(h^2) + O_P\left(\sqrt{\frac{1}{nh^{D_1+D_2+2}}} \right)}{f(\bm{z}) + O(h^2) + O_P\left(\sqrt{\frac{1}{nh^{D_1+D_2}}} \right)} \asymp \frac{\grad f(\bm{z})}{f(\bm{z})}
\end{align*}
under Gaussian and/or von Mises kernels, where $\mathcal{P}_{\bm{z}}$ is the projection matrix onto the tangent space $T_{\bm{z}} = T_{\bm{x}}(\mathcal{S}_1)\times T_{\bm{y}}(\mathcal{S}_2)$ defined in \eqref{Riem_grad} of the main paper and ``$\asymp$'' stands for the asymptotic equivalence. Therefore, the intrinsic adaptive step sizes under $\eta=1$ in our proposed SCMS algorithm \eqref{scaled_SCMS} have an asymptotic lower bound around $\mathcal{R}_d$:
$$\tilde{\eta}_t \text{ or } \tilde{\eta}_t^{(1)}, \tilde{\eta}_t^{(2)} \geq \frac{1}{f(\bm{z})} \geq \frac{1}{f(\bm{m})}$$
when $h$ is sufficiently small and $nh^{D_1+D_2+2}$ is sufficiently large.

Ideally, the intrinsic step sizes of the proposed SCMS algorithm should decrease to zero as the sample size increases to ensure (linear) convergence. To empirically validate the \emph{undesirable} behavior of the SCMS algorithm when a constant step size of $\eta=1$ is used, we simulate two datasets with true manifold structures under the directional-linear and directional-directional data settings.

\begin{figure}[!t]
	\captionsetup[subfigure]{justification=centering}
	\centering
	\begin{subfigure}[t]{.32\textwidth}
		\centering
		\includegraphics[width=1\linewidth]{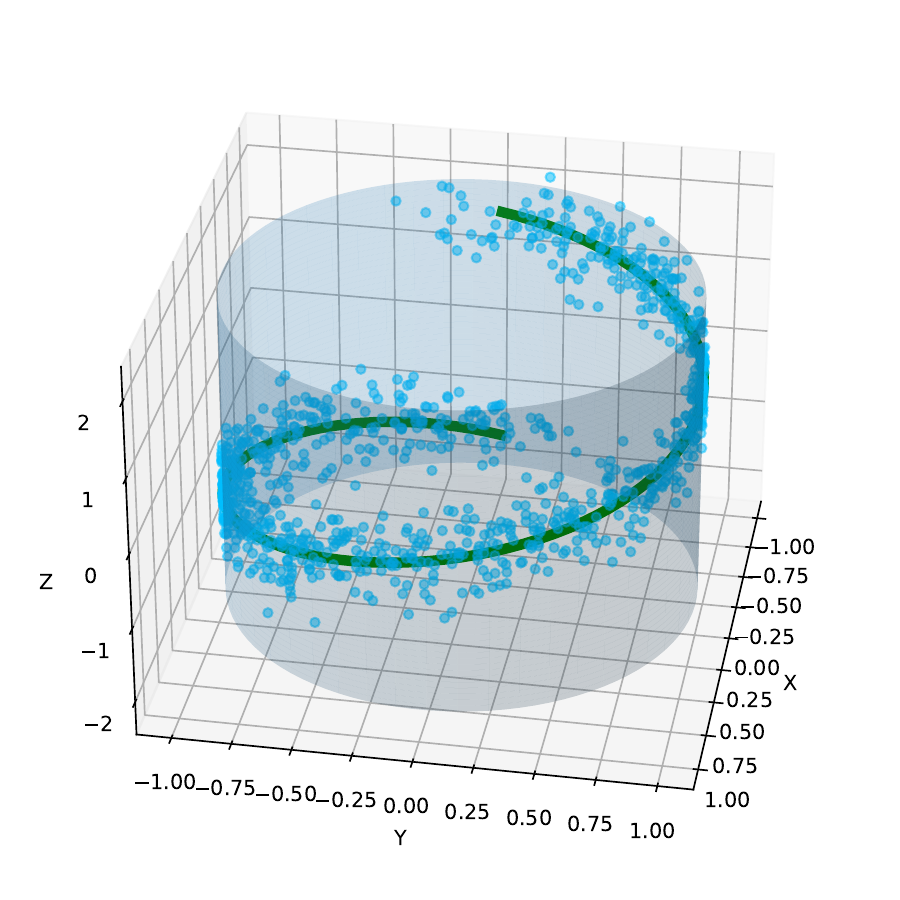}
		\caption{Simulated points around the underlying curve structure.}
	\end{subfigure}
	\hfil
	\begin{subfigure}[t]{.32\textwidth}
		\centering
		\includegraphics[width=1\linewidth]{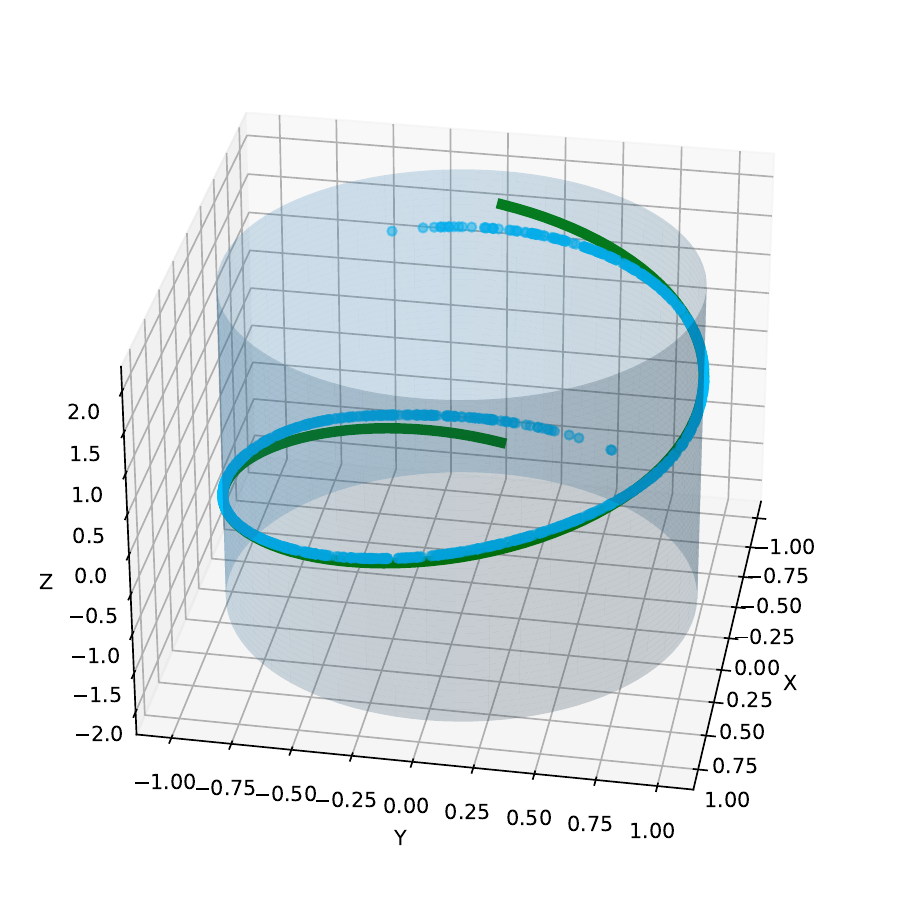}
		\caption{SCMS converged at step 270 with $\eta=0.1\cdot h_1 h_2=0.022$.}
	\end{subfigure}
	\hfil
	\begin{subfigure}[t]{.32\textwidth}
		\centering
		\includegraphics[width=1\linewidth]{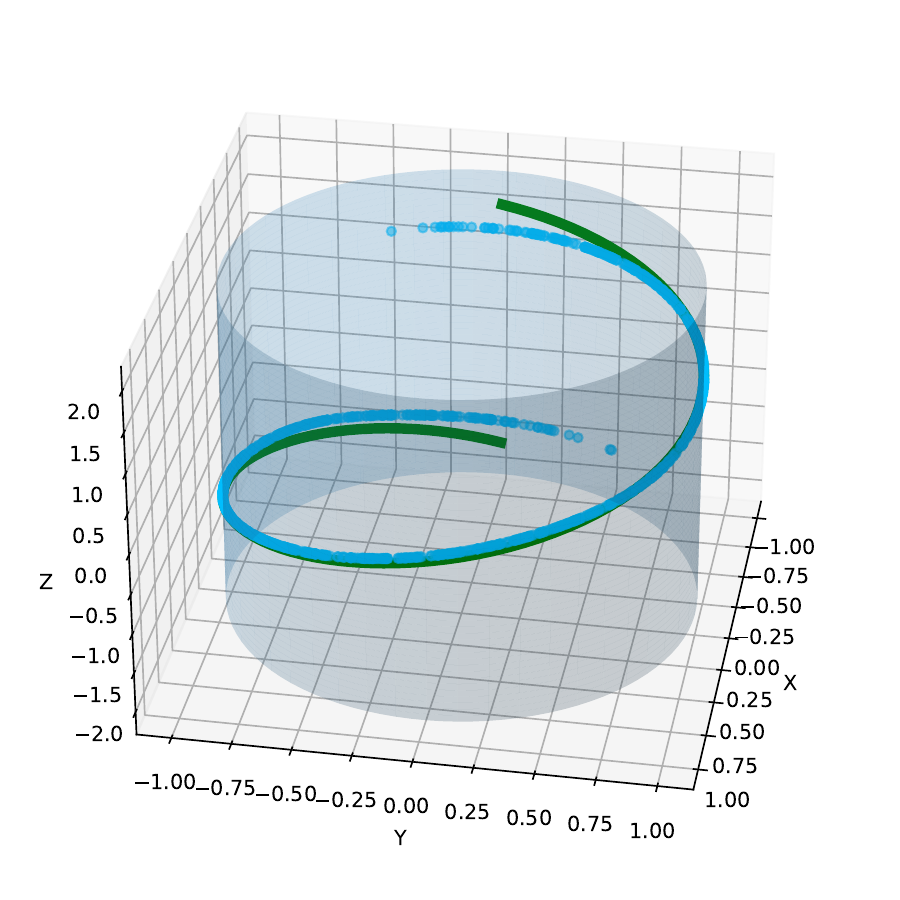}
		\caption{SCMS converged at step 48 with $\eta=0.5\cdot h_1 h_2=0.110$.}
	\end{subfigure}
	\begin{subfigure}[t]{.32\textwidth}
		\centering
		\includegraphics[width=1\linewidth]{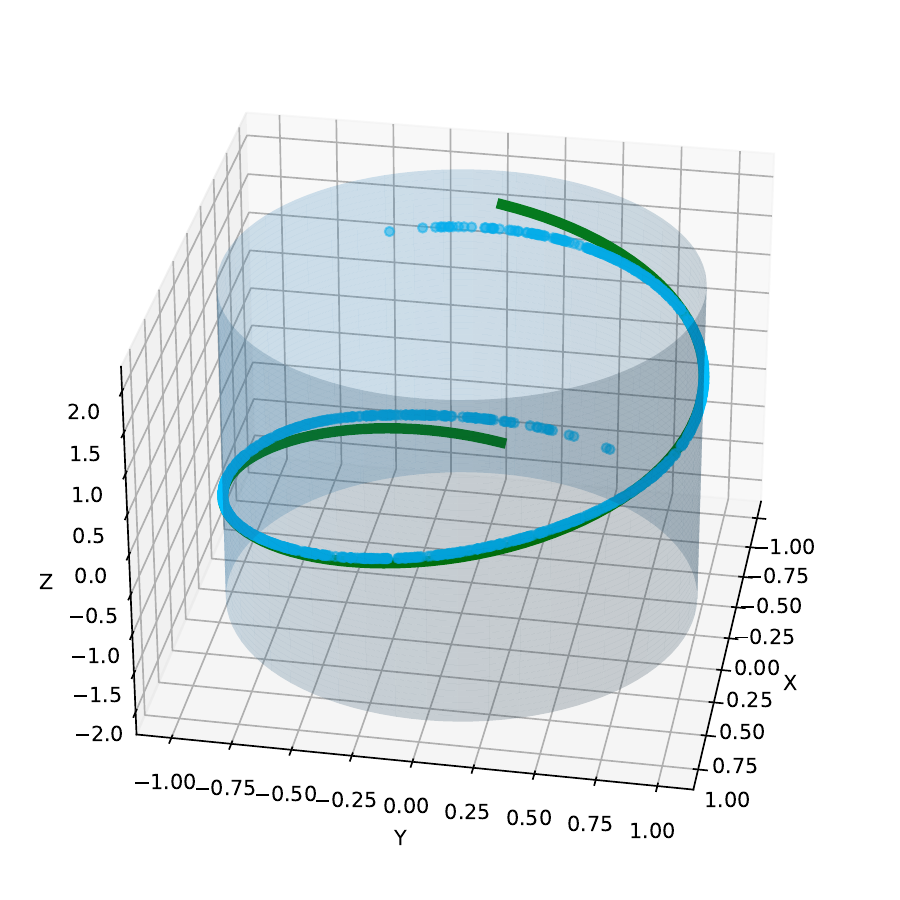}
		\caption{{\bf SCMS converged at step 20 with $\eta=h_1 h_2=0.221$ (proposed)}.}
	\end{subfigure}
	\hfil
	\begin{subfigure}[t]{.32\textwidth}
		\centering
		\includegraphics[width=1\linewidth]{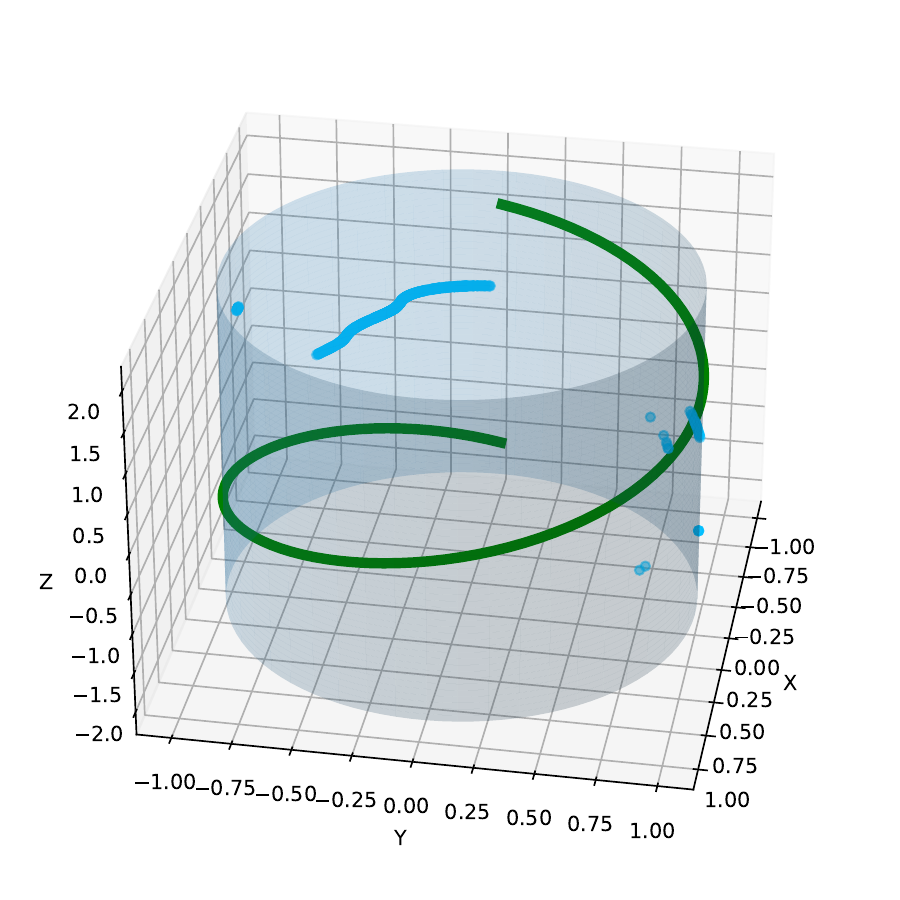}
		\caption{SCMS converged at step 771 with $\eta=1$.}
	\end{subfigure}
	\hfil
	\begin{subfigure}[t]{.32\textwidth}
		\centering
		\includegraphics[width=1\linewidth]{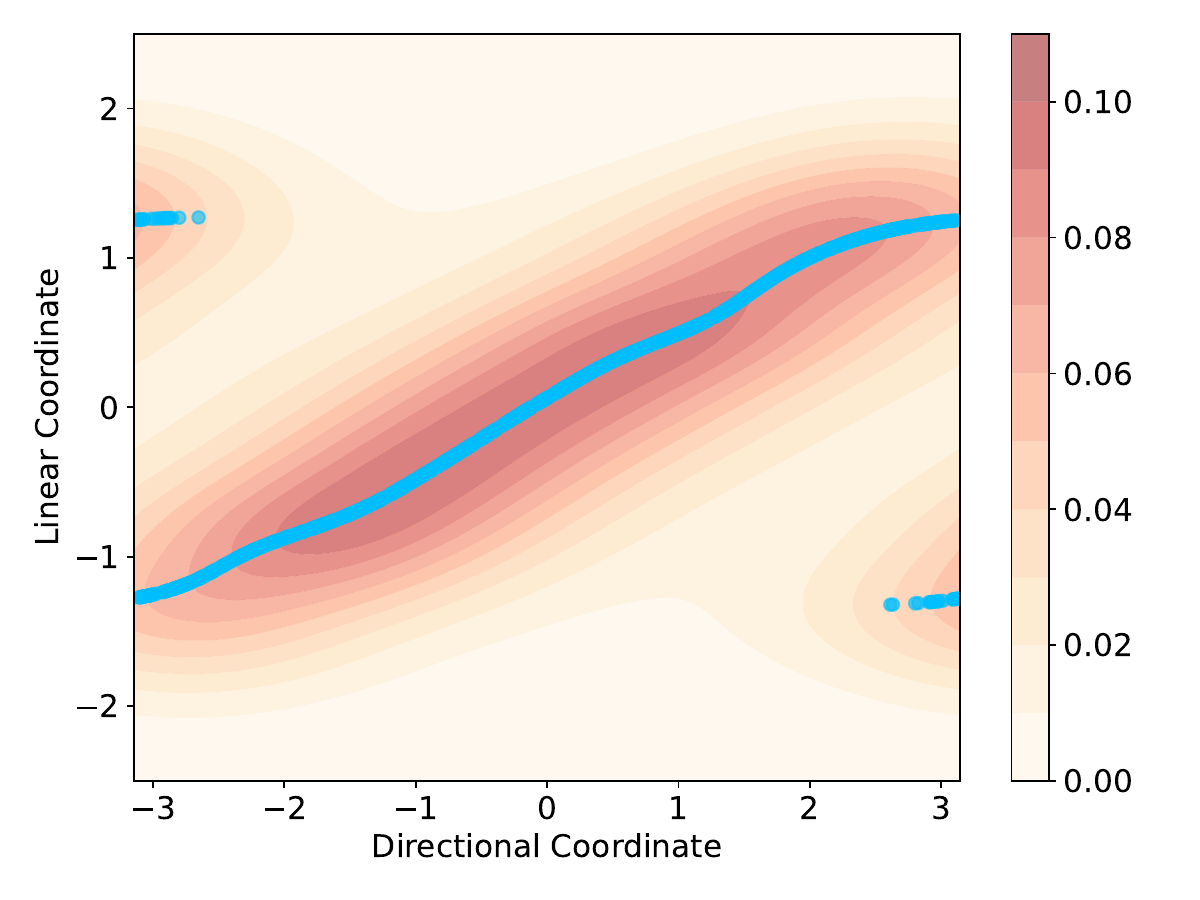}
		\caption{Estimated ridge by the SCMS algorithm with our proposed step size on the contour of $\hat{f}_{\bm{h}}$.}
	\end{subfigure}
	\caption{Proposed SCMS algorithm with different choices of the step size $\eta$ on the simulated directional-linear dataset. In Panels (b-e), the blue points are the final convergent points of the SCMS algorithm under a certain step size $\eta$, while the green curve indicates the underlying curve structure.}
	\label{fig:stepsize_DirLin}
\end{figure}

\begin{example}[Directional-Linear Data]
	We sample 1000 points $t_i, i=1,...,1000$ uniformly from $[-\pi,\pi)$ to obtain the dataset
	$$\left\{\left(\cos(t_i+\epsilon_{i,1}),\sin(t_i + \epsilon_{i,1}), t_i/2 + \epsilon_{i,2}\right) \right\}_{i=1}^{1000} \subset \Omega_1 \times \mathbb{R}$$ 
	lying around an underlying directional-linear curve $(\cos t, \sin t, t/2)$ on the cylinder with radius 1, where $t\in [-\pi,\pi)$ and $\epsilon_{i,1},\epsilon_{i,2} \sim \mathcal{N}(0,\sigma^2)$ are independent Gaussian noises with $\sigma=0.3$; see Panel (a) of \autoref{fig:stepsize_DirLin}. The first two coordinates of each simulated point come from the circular space $\Omega_1$, while the last coordinate lies on the Euclidean/linear space $\mathbb{R}$. We apply the SCMS algorithm \eqref{scaled_SCMS} in the main paper to this simulated dataset with different values of the step size $\eta$; see \autoref{fig:stepsize_DirLin} for details. The bandwidth parameter $h_1$ for the directional component is selected via the rule of thumb \ref{bw_ROT_Dir} in the main paper. In this case, the directional dimension is $D_1=1$ and $h_1\approx0.886$. Furthermore, the bandwidth parameter for the linear part is chosen through the normal reference rule \eqref{NR_rule} with $h_2\approx 0.249$ under the linear dimension as $D_2=1$. The initial set of mesh points for the SCMS algorithm is chosen as the simulated dataset, and the tolerance level for terminating the algorithm is always set to $\epsilon=10^{-7}$.
	
	As shown in \autoref{fig:stepsize_DirLin}, our proposed rule of thumb \eqref{ROT_stepsize} for the step size $\eta$ in the main paper strikes an effective balance between convergence speed and the precision of locating the (estimated) ridge line. A smaller step size than the proposed rule results in slower convergence of the SCMS algorithm, while an excessively large step size relative to the bandwidth parameters causes overshooting, leading to convergence to incorrect structures. The bias observed near the endpoints of the underlying curve is attributed to the boundary bias inherent in density estimation \citep{marron1994transformations,ruppert1994bias}. As we observe from Panel (f) of \autoref{fig:stepsize_DirLin}, the ridge line identified by our proposed SCMS algorithm is consistent with the contour plot of the estimated directional-linear density $\hat{f}_{\bm{h}}$, in which we unfold the cylinder along its linear dimension for a better visualization.
\end{example}

\begin{figure}[!t]
	\captionsetup[subfigure]{justification=centering}
	\centering
	\begin{subfigure}[t]{.32\textwidth}
		\centering
		\includegraphics[width=1\linewidth]{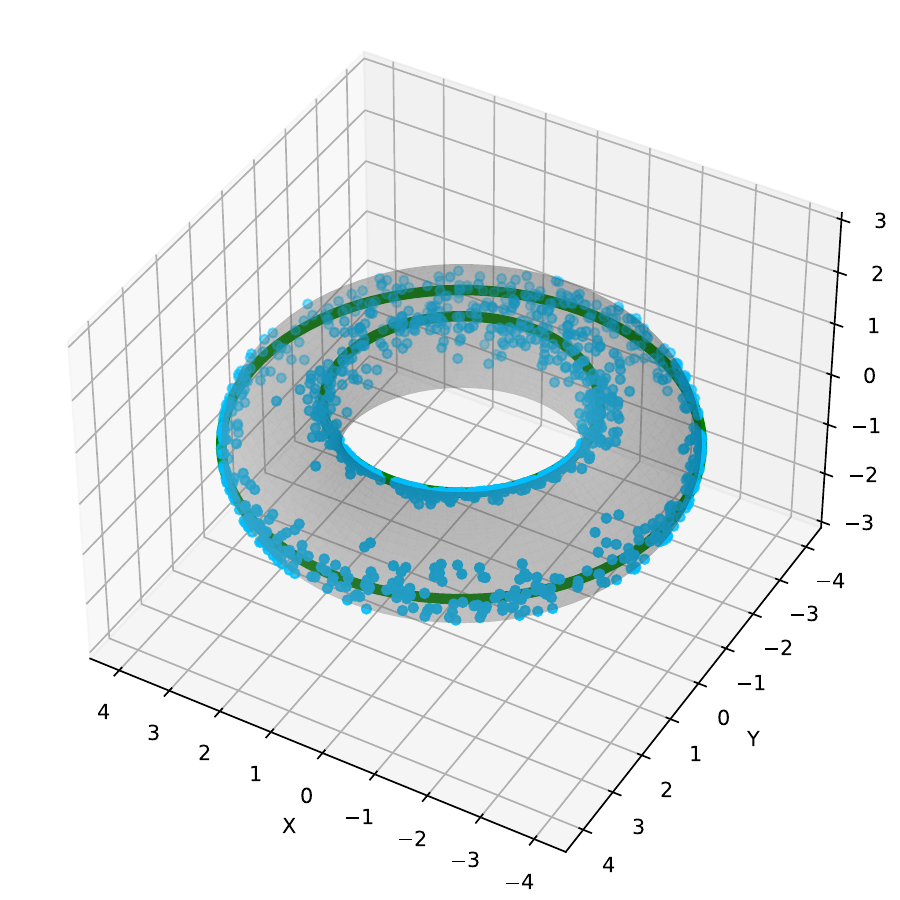}
		\caption{Simulated points around two underlying curve structures.}
	\end{subfigure}
	\hfil
	\begin{subfigure}[t]{.32\textwidth}
		\centering
		\includegraphics[width=1\linewidth]{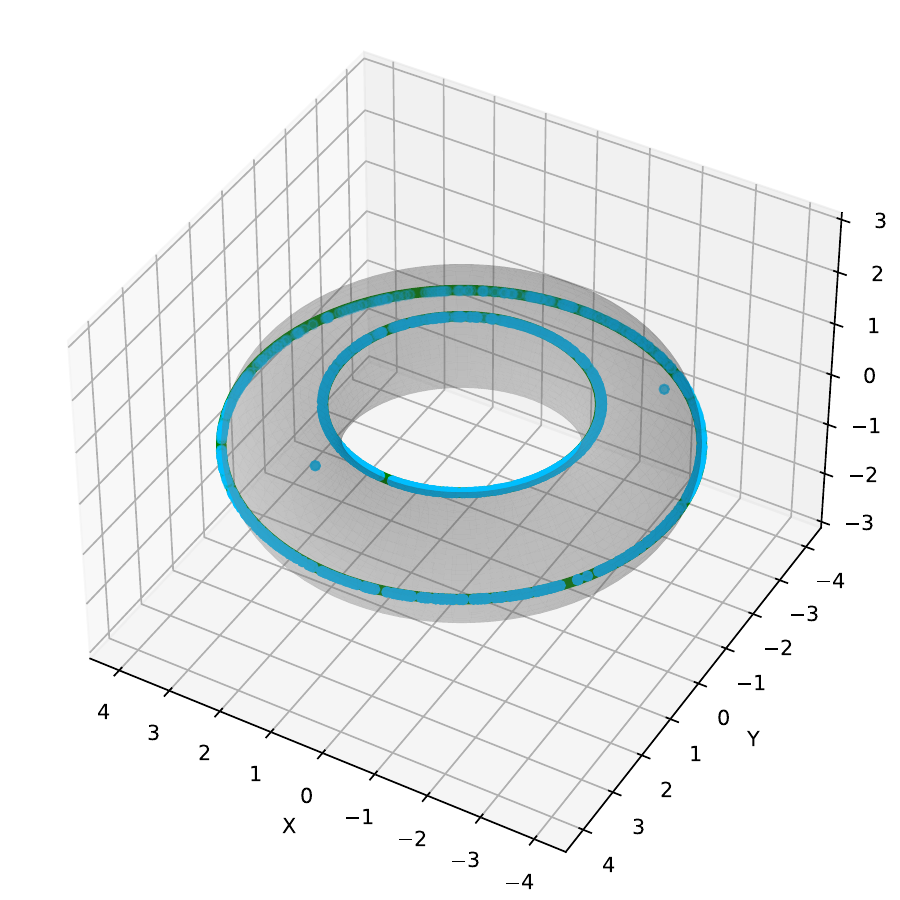}
		\caption{SCMS converged at step 1465 with $\eta=0.1\cdot h_1 h_2=0.024$.}
	\end{subfigure}
	\hfil
	\begin{subfigure}[t]{.32\textwidth}
		\centering
		\includegraphics[width=1\linewidth]{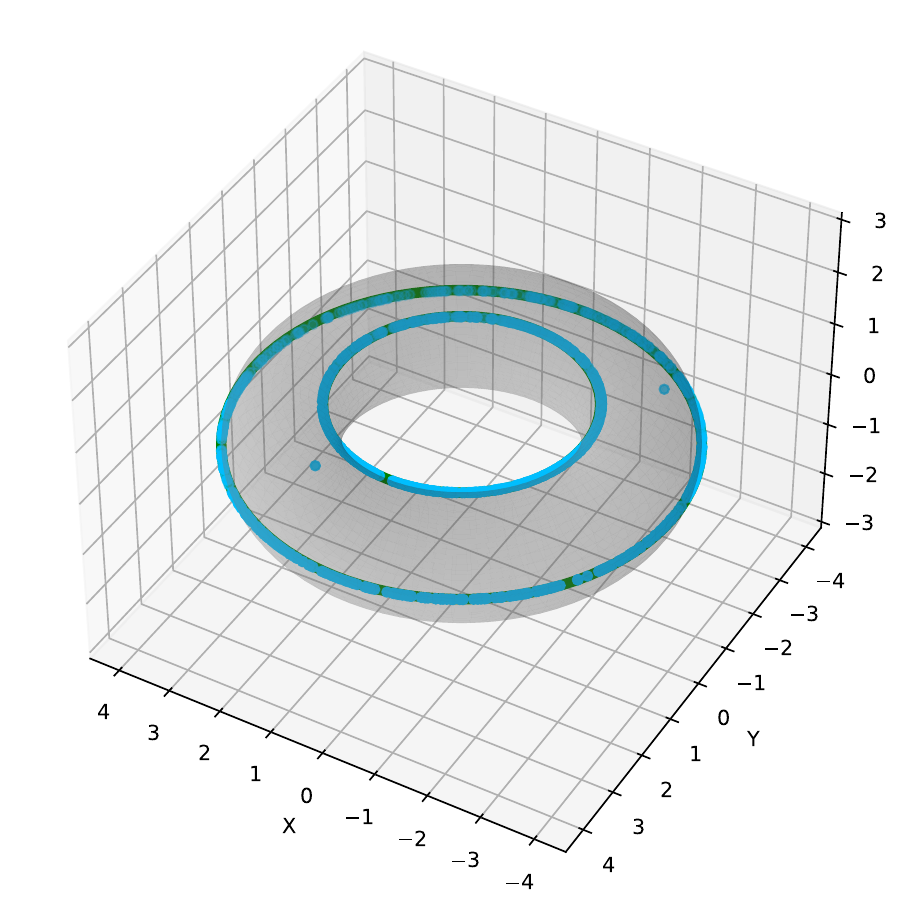}
		\caption{SCMS converged at step 289 with $\eta=0.5\cdot h_1 h_2=0.120$.}
	\end{subfigure}
	\begin{subfigure}[t]{.32\textwidth}
		\centering
		\includegraphics[width=1\linewidth]{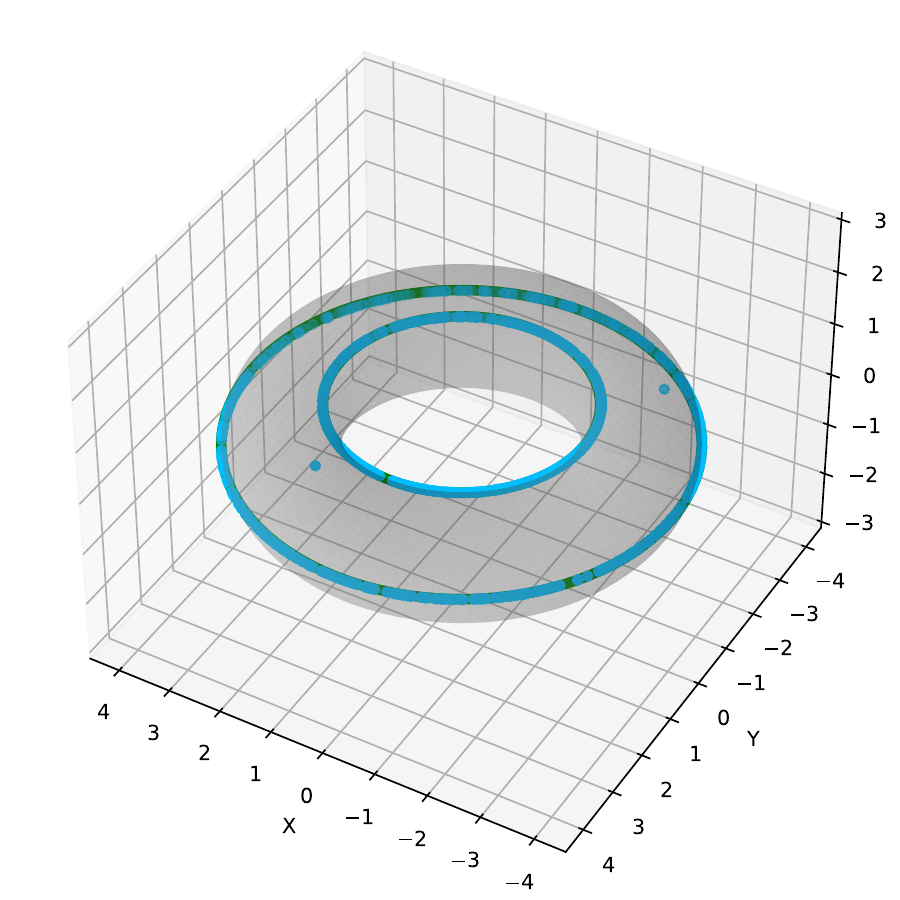}
		\caption{{\bf SCMS converged at step 142 with $\eta=h_1 h_2=0.241$ (proposed)}.}
	\end{subfigure}
	\hfil
	\begin{subfigure}[t]{.32\textwidth}
		\centering
		\includegraphics[width=1\linewidth]{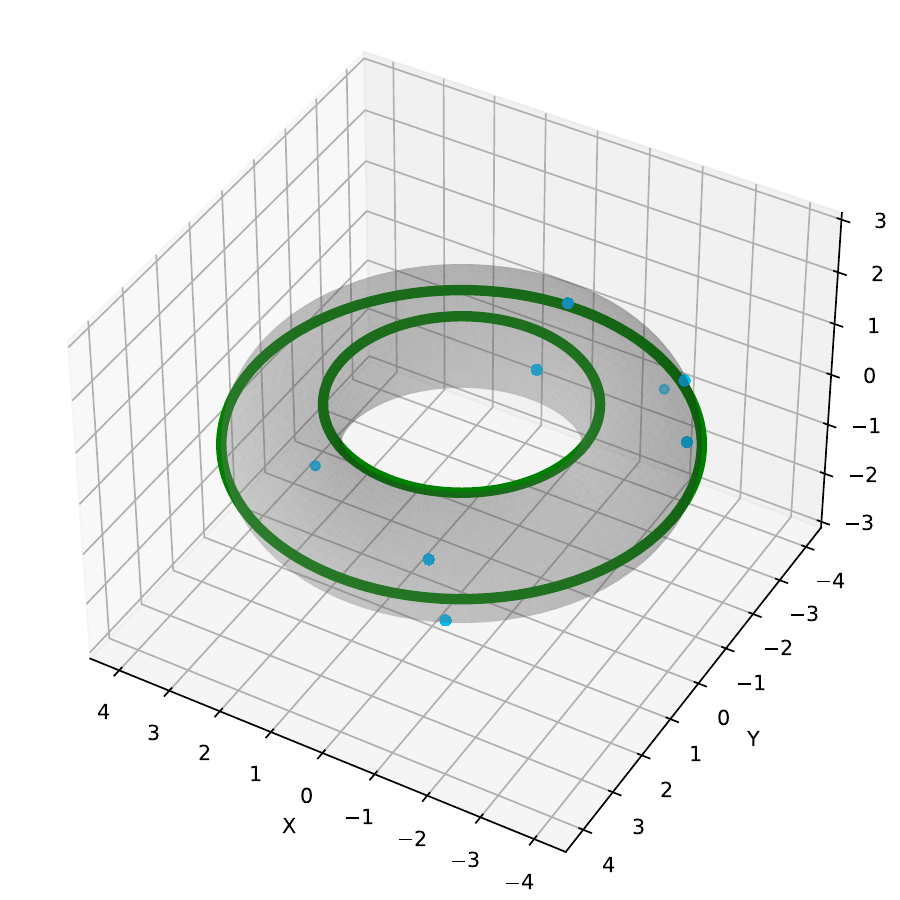}
		\caption{SCMS does not converge after 5000 iterations with $\eta=1$.}
	\end{subfigure}
	\hfil
	\begin{subfigure}[t]{.32\textwidth}
		\centering
		\includegraphics[width=1\linewidth]{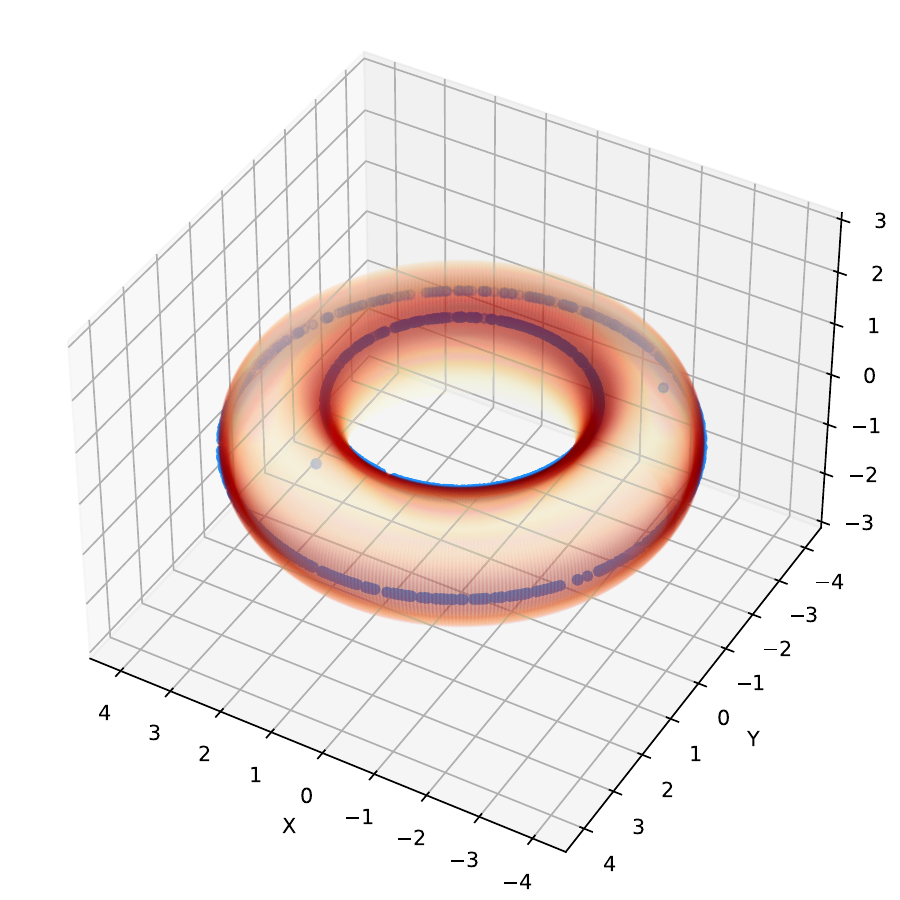}
		\caption{Estimated ridge by the SCMS algorithm with proposed step size on the contour of $\hat{f}_{\bm{h}}$.}
	\end{subfigure}
	\caption{Proposed SCMS algorithm with different choices of the step size $\eta$ on the simulated directional-directional dataset. In Panels (b-e), the blue points are the final convergent points of the SCMS algorithm under a certain step size $\eta$, while the green curve indicates the hidden manifold structure on a torus.}
	\label{fig:stepsize_DirDir}
\end{figure}

\begin{example}[Directional-Directional Data]
	We randomly sample 1000 points $\left\{(\theta_i,\phi_i)\right\}_{i=1}^{1000}$, in which $\theta_i = \theta + \epsilon_{i,1}$ with
	\[
	\theta = \begin{cases}
		0 & \text{ with probability } \frac{1}{2},\\
		\frac{3\pi}{4} & \text{ with probability } \frac{1}{2},
	\end{cases}
	\quad \text{ and } \quad \phi_i=\phi + \epsilon_{i,2} \text{ with } \phi \sim \text{Unif}[-\pi,\pi).
	\]
	Here, the two coordinates of each simulated point are independent and $\epsilon_{i,1},\epsilon_{i,2} \sim \mathcal{N}(0,\sigma^2)$ are independent Gaussian noises with $\sigma=0.3$; see Panel (a) of \autoref{fig:stepsize_DirDir}. The hidden manifold structures can be regarded as two curves in the toroidal direction on a torus. In \autoref{fig:stepsize_DirDir}, we set the distance from the center of the tube to the center of the torus as $c=3$ and the radius of the tube as $a=1$ so that in the 3D Cartesian coordinate system, each directional-directional point $(\theta,\phi)$ has its coordinate as $(X,Y,Z) = \left((c+a\cos\theta) \cos\phi, (c + a\cos\theta) \sin\phi, a\sin\theta\right)$.
	The simulated angular-angular dataset has an alternative directional-directional representation as:
	$\left\{(\cos\theta_i,\sin \theta_i, \cos\phi_i, \sin\phi_i) \right\}_{i=1}^{1000} \subset \Omega_1 \times \Omega_1,$
	to which we apply the SCMS algorithm \eqref{scaled_SCMS} with different values of the step size $\eta$. The bandwidth parameters $h_1,h_2$ are both selected using \eqref{bw_ROT_Dir} in the main paper so that $h_1=0.327$ and $h_2=0.736$. The initial set of mesh points is again chosen as the simulated dataset, and the tolerance level for terminating the algorithm is $\epsilon=10^{-7}$.
	
	On the one hand, consistent with the results for the directional-linear data, \autoref{fig:stepsize_DirDir} suggests that our SCMS algorithm with the proposed step size \eqref{ROT_stepsize} effectively identifies the estimated density ridge and approximates the hidden manifold structure. n the other hand, this example highlights that selecting a step size of $\eta=1$ is suboptimal and leads to undesirable outcomes.
\end{example}

\subsection{Linear Convergence of the Proposed Mean Shift and SCMS Algorithm on $\mathcal{S}_1\times \mathcal{S}_2$}
\label{Sec:Lin_conv_SCMS}

Our derivations in \autoref{Sec:MS_stepsize} demonstrate that the intrinsic step sizes $\tilde{\eta}_t^{(1)}, \tilde{\eta}_t^{(2)}$ of our proposed mean shift algorithm on $\mathcal{S}_1\times \mathcal{S}_2$ under the gradient ascent framework \eqref{GA_manifold_comp} have an asymptotic order $O(h^2)$ as the bandwidths $h_1,h_2\lesssim h$ are small and the sample size $n$ is large. Therefore, one can choose small bandwidths $h_1,h_2$ in order for the mean shift algorithm (either Version A or B) to converge linearly to an (estimated) local mode when the algorithm is initialized around the local mode, leading to the proof of Theorem~\ref{Thm:MS_lin_conv} in the main paper as follows.

\begin{customthm}{2}
	Assume that the conditions of Theorem~\ref{Thm:Mode_cons} hold. Given the sequence $\left\{\bm{z}^{(t)}\right\}_{t=0}^{\infty}$ defined by our mean shift algorithm \eqref{MS_iter_sim} or \eqref{MS_iter_comp}, there exist constants $\tilde{r}_1 >0, \Upsilon_1 \in (0,1)$ such that
	$$d_g(\bm{z}^{(t)}, \hat{\bm{m}}) \leq \Upsilon_1^t \cdot d_g(\bm{z}^{(0)}, \hat{\bm{m}}) \quad \text{ and } \quad d_g(\bm{z}^{(t)}, \bm{m}) \leq \Upsilon_1^t \cdot d_g(\bm{z}^{(0)}, \bm{m}) + O(h^2) + O_P\left(\sqrt{\frac{|\log h|}{nh^{D_1+D_2}}} \right)$$
	when $\bm{z}^{(0)} \in \left\{\bm{z}\in \mathcal{S}_1\times\mathcal{S}_2: d_g(\bm{z},\bm{m}) \leq \tilde{r}_1 \right\}$ with $\bm{m}\in \mathcal{M}$, $\max(\bm{h}) =\max\left\{h_1,h_2\right\} \lesssim h$ is sufficiently small, and the sample size $n$ is sufficiently large. Here, $D_T=D_1+D_2$ is the intrinsic dimension and $d_g(\cdot,\cdot)$ is the geodesic distance on $\mathcal{S}_1\times \mathcal{S}_2$.
\end{customthm}

\begin{proof}[Proof of Theorem~\ref{Thm:MS_lin_conv}]
	First, by Lemma~\ref{unif_rate_KDE}, we know that when the bandwidths $h_1,h_2 \lesssim h$ are sufficiently small and $\frac{nh^{D_1+D_2+4}}{|\log h|}$ is sufficiently large, the KDE $\hat{f}_{\bm{h}}$ also satisfies the condition in Theorem~\ref{Thm:Mode_cons}.
	
	Second, we argue that given the chosen bandwidths $h_1,h_2$, the intrinsic step sizes $\tilde{\eta}_t^{(1)}, \tilde{\eta}_t^{(2)}$ of our proposed mean shift algorithm on $\mathcal{S}_1\times \mathcal{S}_2$ under the gradient ascent framework \eqref{GA_manifold_comp} are lower bounded away from zero. Recall from \autoref{Sec:MS_stepsize} that the intrinsic step sizes $\tilde{\eta}_t^{(1)}, \tilde{\eta}_t^{(2)}$ of the simultaneous mean shift algorithm (Version A in \autoref{Sec:MS_prod}) on $\mathcal{S}_1\times \mathcal{S}_2$ are given by
	\begin{itemize}
		\item {\bf Case 1: } When $\mathcal{S}_1\times \mathcal{S}_2 = \mathbb{R}^{D_1} \times \mathbb{R}^{D_2}$,
		\begin{align*}
			\tilde{\eta}_t^{(1)} &= -\frac{h_1^2}{\frac{2\prod_{j=1}^2 C_{k_j,D_j}}{n} \sum\limits_{i=1}^n k_1'\left(\norm{\frac{\bm{x}^{(t)}-\bm{X}_i}{h_1}}_2^2\right) k_2\left(\norm{\frac{\bm{y}^{(t)}-\bm{Y}_i}{h_2}}_2^2 \right)} \\ 
			\text{ and } & \quad \tilde{\eta}_t^{(2)} = -\frac{h_2^2}{\frac{2\prod_{j=1}^2 C_{k_j,D_j}}{n} \sum\limits_{i=1}^n k_1\left(\norm{\frac{\bm{x}^{(t)}-\bm{X}_i}{h_1}}_2^2\right) k_2'\left(\norm{\frac{\bm{y}^{(t)}-\bm{Y}_i}{h_2}}_2^2 \right)}
		\end{align*}
		with $\bm{z}^{(t)}=\left(\bm{x}^{(t)},\bm{y}^{(t)}\right)$.
		
		\item {\bf Case 2: } When $\mathcal{S}_1\times \mathcal{S}_2 = \Omega_{D_1} \times \mathbb{R}^{D_2}$, 
		\begin{align*}
			\tilde{\eta}_t^{(1)} &= \frac{\theta_t^{(1)}}{\tilde{\nabla}_{\bm{x}}\hat{f}_{\bm{h}}(\bm{z}^{(t)}) \cdot \sin \theta_t^{(1)}} \\ 
			\text{ and } &\quad
			\tilde{\eta}_t^{(2)} = -\frac{h_2^2}{\frac{2\prod_{j=1}^2 C_{k_j,D_j}}{n} \sum\limits_{i=1}^n k_1\left(\norm{\frac{\bm{x}^{(t)}-\bm{X}_i}{h_1}}_2^2\right) k_2'\left(\norm{\frac{\bm{y}^{(t)}-\bm{Y}_i}{h_2}}_2^2 \right)},
		\end{align*}
		where $\tilde{\nabla}_{\bm{x}} \hat{f}_{\bm{h}}(\bm{z}) = -\frac{C_{k_1,D_1}(h_1) \cdot C_{k_2,D_2}}{nh_1^2 h_2^{D_2}} \sum\limits_{i=1}^n \bm{X}_i \cdot  L'\left(\frac{1-\bm{X}_i^{\top}\bm{x}}{h_1^2} \right)  k_2\left(\norm{\frac{\bm{y}-\bm{Y}_i}{h_2}}_2^2 \right)$.
		
		\item {\bf Case 3: } When $\mathcal{S}_1\times \mathcal{S}_2 = \Omega_{D_1}\times \Omega_{D_2}$,
		$$\tilde{\eta}_t^{(1)} = \frac{\theta_t^{(1)}}{\tilde{\nabla}_{\bm{x}}\hat{f}_{\bm{h}}(\bm{z}^{(t)}) \cdot \sin \theta_t^{(1)}}\quad \text{ and } \quad \tilde{\eta}_t^{(2)} = \frac{\theta_t^{(2)}}{\tilde{\nabla}_{\bm{y}}\hat{f}_{\bm{h}}(\bm{z}^{(t)}) \cdot \sin \theta_t^{(2)}}.$$
	\end{itemize}
	Under the differentiability condition (A2) on $k_1,k_2$ and the fact that $\frac{\theta_t^{(j)}}{\sin\theta_t^{(j)}} \to 1$ as $\theta_t^{(j)}\to 0$ for $j=1,2$, we know that the denominators of $\tilde{\eta}_t^{(j)}, j=1,2$ are upper bounded by some universal constant. Therefore, when $h_1,h_2$ are chosen, the intrinsic step sizes $\tilde{\eta}_t^{(j)}, j=1,2$ are lower bounded away from zero. 
	% (Notice that the denominators of $\tilde{\eta}_t^{(j)}, j=1,2$ are nonzero, i.e., the kernel profiles and its first-order derivatives are nonzero almost everywhere under the data configuration $\left\{\bm{Z}_1,...,\bm{Z}_n \right\}$ and the mean shift sequence $\left\{\bm{z}^{(t)}\right\}_{t=0}^{\infty}$.)
	
	Let $\Upsilon_1 = \sqrt{1-\frac{\lambda^* \tilde{\eta}^*}{2}}$, where $\lambda^*$ is the lower bound of absolute eigenvalues of the Riemannian Hessian $\mathcal{H} f(\bm{z})$ of the true (density) function $f$ defined in Theorem \eqref{Thm:Mode_cons} and $\tilde{\eta}^* = \min\left\{\inf_t \tilde{\eta}_t^{(1)}, \inf_t \tilde{\eta}_t^{(2)} \right\} =O(h^2) >0$ when the bandwidths satisfy $h_1,h_2\lesssim h$. As $h_1,h_2\lesssim h$ are chosen to be small enough and the sample size $n$ is sufficiently large, $\tilde{\eta}^*$ can fall below the step size threshold of the linear convergence; see Theorem 2.1.15 in \cite{nesterov2003introductory} and Theorem 12 in \cite{DirMS2020}. Since $\tilde{\eta}^*$ is a fixed constant under a given but large sample size $n$ and the chosen but small bandwidths $h_1,h_2$, the results follow from their arguments for the standard gradient ascent methods in the Euclidean or directional spaces under the (geodesically) strong concavity around the (estimated) local modes $\bm{m}$ or $\hat{\bm{m}}$. 
\end{proof}

Now, we consider the (linear) convergence of our proposed SCMS algorithm on $\mathcal{S}_1\times \mathcal{S}_2$. Given the bounds \eqref{intrinsic_step_size_bound} and Lemma~\ref{pw_rate_KDE}, the intrinsic step size of the algorithm with our proposed rule of thumb for the step size 
$\eta=\min\left\{\max(\bm{h})\cdot \min(\bm{h}),1 \right\}=\min\left\{h_1h_2,1\right\}$
under the subspace constrained gradient ascent framework \eqref{SCGA} has an asymptotic rate $O(h^2)$ when the bandwidths $h_1,h_2 \lesssim h$ are small and the sample size $n$ is large. The shrinking property of the intrinsic step size with respect to the bandwidths $h_1,h_2 \lesssim h$ guarantees that our SCMS algorithm on $\mathcal{S}_1\times \mathcal{S}_2$ converges (linearly) to the estimated ridge $\hat{\mathcal{R}}_d$ (and true ridge $\mathcal{R}_d$); see the full version of Theorem~\ref{Thm:SCMS_lin_conv} as follows.

\begin{theorem}
	\label{Thm:SCMS_lin_conv_full}
	Assume that all the assumptions in Theorem~\ref{Thm:Ridge_cons} hold. Under the constant $\beta_0 >0$ defined in Theorem~\ref{Thm:Ridge_cons}, we also assume that 
	\begin{itemize}
		\item {\bf (A4)} there exists a constant $\beta_1 \in (0, \beta_0)$ such that 
		$$d \cdot (D_1+D_2)^{\frac{3}{2}} \norm{\nabla f(\bm{z})}_2 \norm{\nabla^3 f(\bm{z})}_{\max} \leq \beta_0 (\beta_0 -\beta_1)$$ 
		for all $\bm{z}\in \mathcal{R}_d$.
	\end{itemize}
	Given the sequence $\left\{\bm{z}^{(t)} \right\}_{t=0}^{\infty}$ defined by our SCMS algorithm \eqref{scaled_SCMS} with step size $\eta$ dominated by $\max(\bm{h}) \lesssim h$, there exist a constant $\tilde{r}_2 >0$ such that with probability at least $1-\delta$ for any $\delta \in (0,1)$,
	\[
	\lim_{t\to\infty} d_g\left(\bm{z}^{(t)}, \hat{\mathcal{R}}_d\right) = 0 \;\; \text{ and } \;\; \lim_{t\to\infty} d_g\left(\bm{z}^{(t)}, \mathcal{R}_d\right) = O(h^2) + O_P\left(\sqrt{\frac{|\log h|}{nh^{D_1+D_2+4}}} \right)
	\]
	when $\bm{z}^{(0)} \in \hat{\mathcal{R}}_d \oplus \tilde{r}_2=\left\{\bm{z}\in \mathcal{S}_1\times \mathcal{S}_2: d_g(\bm{z},\hat{\mathcal{R}}_d) \leq \tilde{r}_2 \right\}$, $\max(\bm{h})\lesssim h$ is sufficiently small, and the sample size $n$ is sufficiently large. Here, $d_g(\bm{z},\mathcal{R}_d) = \inf\left\{d_g(\bm{z},\bm{x}): \bm{x}\in \mathcal{R}_d \right\}$ with $d(\cdot,\cdot)$ being the geodesic distance on $\mathcal{S}_1\times \mathcal{S}_2$. Denote the limiting point of the above SCMS sequence $\left\{\bm{z}^{(t)} \right\}_{t=0}^{\infty}$ by $\bm{z}^*$. Further, we assume that 
	\begin{itemize}
		\item {\bf (A5)} the sequence $\left\{\bm{z}^{(t)} \right\}_{t=0}^{\infty}$ satisfies the subspace constrained version of the Polyak-{\L}ojasiewicz inequality \citep{karimi2016linear} as:
		$$\frac{1}{2} \norm{V_d(\bm{z}^{(t)})^{\top} \nabla f(\bm{z}^{(t)})}_2^2 \geq \beta_2\left[ f(\bm{z}^*) -f(\bm{z}^{(t)})\right]$$
		for all $t\geq 0$ and some constant $\beta_2>0$.
	\end{itemize} 
	Then, there exist a constant $\tilde{r}_3>0$ and $\Upsilon_2\in (0,1)$ such that
	\begin{align*}
		&d_g\left(\bm{z}^{(t)},\hat{\mathcal{R}}_d \right) \leq \Upsilon_2^t \cdot d_g\left(\bm{z}^{(0)}, \hat{\mathcal{R}}_d \right) \\ 
		&\text{ and } \quad d_g\left(\bm{z}^{(t)},\mathcal{R}_d \right) \leq \Upsilon_2^t \cdot d_g\left(\bm{z}^{(0)},\mathcal{R}_d \right) + O(h^2) + O_P\left(\sqrt{\frac{|\log h|}{nh^{D_1+D_2+4}}} \right)
	\end{align*}
	when $\bm{z}^{(0)} \in \mathcal{R}_d \oplus \tilde{r}_3=\left\{\bm{z}\in \mathcal{S}_1\times \mathcal{S}_2: d_g(\bm{z},\mathcal{R}_d) \leq \tilde{r}_3 \right\}$, $\max(\bm{h})\lesssim h$ is sufficiently small, and the sample size $n$ is sufficiently large.
\end{theorem}

Condition (A4) is the regularity assumption on the third-order derivative proposed by \cite{Asymp_ridge2015} in order for a well-defined local normal coordinate along the ridge $\mathcal{R}_d$. Under this condition and assumptions in Theorem~\ref{Thm:Ridge_cons}, our proposed SCMS algorithm on $\mathcal{S}_1\times \mathcal{S}_2$ is guaranteed to converge to the estimated density ridge $\hat{\mathcal{R}}_d$ around its small neighborhood when the bandwidths $h_1,h_2$ are small and the sample size $n$ is large. In order for the SCMS sequence to converge linearly, we require the sequence to satisfy the subspace constrained Polyak-{\L}ojasiewicz inequality (A5). In other words, the objective (density) function $f$ should be sharply decayed off the ridge $\mathcal{R}_d$ in the direction of the projected (Riemannian) gradient $V_d(\bm{z}) V_d(\bm{z})^{\top} \grad f(\bm{z}) = V_d(\bm{z}) V_d(\bm{z})^{\top} \nabla f(\bm{z})$.

\begin{proof}[Proof of Theorem~\ref{Thm:SCMS_lin_conv_full}]
	On the one hand, given the bounds \eqref{intrinsic_step_size_bound}, the intrinsic step sizes $\tilde{\eta}_t$ or $\tilde{\eta}_t^{(j)}, j=1,2$ of our proposed SCMS algorithm under the subspace constrained gradient ascent framework on $\mathcal{S}_1\times \mathcal{S}_2$ are of the order $O(h^2)$ and can be small enough when the bandwidths $h_1,h_2\lesssim h$ are chosen to be sufficiently small. Then, by Corollaries 3.4 and 4.5 in \cite{DirSCMS2021}, we know that $\lim_{t\to\infty} d_g\left(\bm{z}^{(t)}, \hat{\mathcal{R}}_d\right) = 0$ with high probability when the SCMS sequence is initialized within a small neighborhood of $\hat{\mathcal{R}}_d$. In addition, the result $\lim_{t\to\infty} d_g\left(\bm{z}^{(t)}, \mathcal{R}_d\right) = O(h^2) + O_P\left(\sqrt{\frac{|\log h|}{nh^{D_1+D_2+4}}} \right)$ follows from the stability of density ridges on $\mathcal{S}_1\times\mathcal{S}_2$ (Lemma~\ref{Thm:Ridge_cons} here).
	
	On the other hand, given the bounds \eqref{intrinsic_step_size_bound} and condition (A2), we know that, when the bandwidths $h_1,h_2\lesssim h$ are chosen to be sufficiently small and $\frac{nh^{D_1+D_2+6}}{|\log h|}$ is large but fixed, the intrinsic step sizes $\tilde{\eta}_t$ or $\tilde{\eta}_t^{(j)}, j=1,2$ of our proposed SCMS algorithm under the subspace constrained gradient ascent framework on $\mathcal{S}_1\times \mathcal{S}_2$ are uniformly lower bounded away from zero with respect to the iteration number $t$, assuming that $\hat{f}_{\bm{h}}(\bm{z}^{(t)}), t=0,1,...$ or their population quantities $f(\bm{z}^{(t)}), t=0,1,...$ are nonzero. 
	
	Let $\Upsilon_2= \sqrt{1-\frac{\beta_0\tilde{\eta}^{**}}{4}}$, where $\beta_0>0$ is the eigengap between the $d$-th and $(d+1)$-th eigenvalues of the Riemannian Hessian $\mathcal{H}f(\bm{z})$ within the tangent space $T_{\bm{z}}$ defined in Theorem~\ref{Thm:Ridge_cons} and $\tilde{\eta}^{**}=\inf_t \tilde{\eta}_t \text{ or } \min\left\{\inf_t \tilde{\eta}_t^{(1)}, \inf_t \tilde{\eta}_t^{(2)} \right\}=O(h^2) >0$ when the bandwidths $h_1,h_2$ are dominated by $h$. As $h_1,h_2\lesssim h$ are chosen to be small and the sample size $n$ is large, $\tilde{\eta}^{**}$ can fall below the step size threshold of the linear convergence; see Theorems 3.6 and 4.6 in \cite{DirSCMS2021}. Notice that $\tilde{\eta}^{**}$ is a fixed constant once the large sample size $n$ is given and the bandwidths $h_1,h_2$ are chosen to be small. Therefore, our results follow from the arguments in Theorems 3.6 and 4.6 as well as Corollaries 3.7 and 4.7 in \cite{DirSCMS2021}.
\end{proof}

\section{Proofs of Theorem~\ref{Thm:MS_conv} and Proposition~\ref{Thm:trans_grad_conv}}
\label{Sec:Conv_pf}

\begin{customthm}{1}
	Denote the sequence defined by the mean shift algorithm by $\left\{\bm{z}^{(t)}\right\}_{t=0}^{\infty}= \left\{(\bm{x}^{(t)},\bm{y}^{(t)}) \right\}\subset \mathcal{S}_1\times\mathcal{S}_2$. Assume that 
	
	$\bullet$ {\bf (C1)} The kernel profiles $k_1,k_2$ (either linear $k$ or directional $L$) are strictly decreasing and differentiable on $[0,\infty)$ with $k_1(0), k_2(0) < \infty$.
	
	$\bullet$ {\bf (Weak Condition)} Both $k_1$ and $k_2$ are convex.
	
	$\bullet$ {\bf (Strong Condition)} The entire product kernel profile $K(r,s) = k_1(r)\cdot k_2(s)$ is convex.
	
\noindent Then, for any fixed bandwidths $\bm{h}$ and sample size $n$, we have that
	
	(a) Under {\bf (C1)} and {\bf (Weak Condition)}, the sequence of density estimates $\left\{\hat{f}_{\bm{h}}(\bm{z}^{(t)}) \right\}_{t=0}^{\infty}$ yielded by Version B is non-decreasing and thus converges.
	
	(b) Under {\bf (C1)} and {\bf (Strong Condition)}, the sequence of density estimates $\left\{\hat{f}_{\bm{h}}(\bm{z}^{(t)}) \right\}_{t=0}^{\infty}$ yielded by either Version A or B is non-decreasing and thus converges.
	
	(c) Under the assumptions in (a) or (b), we have that  $\lim\limits_{t\to\infty} \norm{\bm{z}^{(t+1)}-\bm{z}^{(t)}}_2=0$.
	
	(d) Assume the conditions in (a) or (b). If the local modes of $\hat{f}_{\bm{h}}$ are isolated, $\left\{\bm{z}^{(t)}\right\}_{t=0}^{\infty}$ converges to a local mode of $\hat{f}_{\bm{h}}$ when it is initialized within its small neighborhood.
\end{customthm}

\begin{proof}[Proof of \autoref{Thm:MS_conv}]
	We only present the proof for the most complex scenario, where one of the kernel profiles $k_1,k_2$ is directional $L$ and the other is linear $k$. One can slightly modify our argument to the linear-linear and directional-directional cases. Without the loss of generality, we assume that $k_1(r)=L(r)$ is directional and $k_2(s)=k(s)$ is linear.
	
	(a) Our condition (C1) guarantees that the sequence $\left\{\hat{f}_{\bm{h}}(\bm{x}^{(t)},\bm{y}^{(t)}) \right\}_{t=0}^{\infty}$ is bounded. Hence, it suffices to prove that it is non-decreasing. The convexity and differentiability of $L$ and $k$ imply that
	\begin{equation}
		\label{convex_diff}
		L(x_2) -L(x_1) \geq L'(x_1) \cdot (x_2-x_1) \quad \text{ and } \quad k(x_2) - k(x_1) \geq k'(x_1) \cdot (x_2 - x_1) 
	\end{equation}
	for any $x_1,x_2 \in [0,\infty)$. Notice that $\norm{\bm{x}^{(t)}}_2 = \norm{\bm{X}_i}_2=1$ for all $t=0,1,...$ and $i=1,...,n$. Given the inequalities in \eqref{convex_diff}, we calculate that
	\begin{align*}
		&\hat{f}_{\bm{h}}(\bm{x}^{(t+1)}, \bm{y}^{(t+1)}) - \hat{f}_{\bm{h}}(\bm{x}^{(t)}, \bm{y}^{(t)}) \\
		&= \frac{C_{L,D_1}(h_1) \cdot C_{k,D_2}}{nh_2^{D_2}}\sum_{i=1}^n \Bigg[ L\left(\frac{1-\bm{X}_i^T \bm{x}^{(t+1)}}{h_1^2} \right)  k\left(\norm{\frac{\bm{y}^{(t+1)}-\bm{Y}_i}{h_2}}_2^2 \right) \\
		&\hspace{35mm} - L\left(\frac{1-\bm{X}_i^T \bm{x}^{(t)}}{h_1^2} \right)  k\left(\norm{\frac{\bm{y}^{(t)}-\bm{Y}_i}{h_2}}_2^2 \right) \Bigg]\\
		&= \frac{C_{L,D_1}(h_1) \cdot C_{k,D_2}}{nh_2^{D_2}} \sum_{i=1}^n \Bigg[ L\left(\frac{1-\bm{X}_i^T \bm{x}^{(t+1)}}{h_1^2} \right)  k\left(\norm{\frac{\bm{y}^{(t+1)}-\bm{Y}_i}{h_2}}_2^2 \right) \\
		&\hspace{35mm} - L\left(\frac{1-\bm{X}_i^T \bm{x}^{(t+1)}}{h_1^2} \right)  k\left(\norm{\frac{\bm{y}^{(t)}-\bm{Y}_i}{h_2}}_2^2 \right) \\
		&\hspace{35mm} + L\left(\frac{1-\bm{X}_i^T \bm{x}^{(t+1)}}{h_1^2} \right)  k\left(\norm{\frac{\bm{y}^{(t)}-\bm{Y}_i}{h_2}}_2^2 \right) \\
		&\hspace{35mm} - L\left(\frac{1-\bm{X}_i^T \bm{x}^{(t)}}{h_1^2} \right)  k\left(\norm{\frac{\bm{y}^{(t)}-\bm{Y}_i}{h_2}}_2^2 \right) \Bigg]\\
		&\stackrel{\text{(i)}}{=} \frac{C_{L,D_1}(h_1) \cdot C_{k,D_2}}{nh_2^{D_2+2}} \sum_{i=1}^n L\left(\frac{1-\bm{X}_i^T \bm{x}^{(t+1)}}{h_1^2} \right)  k'\left(\norm{\frac{\bm{y}^{(t)}-\bm{Y}_i}{h_2}}_2^2 \right) \left[\norm{\bm{y}^{(t+1)} -\bm{Y}_i}_2^2 -\norm{\bm{y}^{(t)} -\bm{Y}_i}_2^2 \right]\\
		&\quad + \frac{C_{L,D_1}(h_1) \cdot C_{k,D_2}}{nh_1^2h_2^{D_2}} \sum_{i=1}^n L'\left(\frac{1-\bm{X}_i^T \bm{x}^{(t)}}{h_1^2} \right)  k\left(\norm{\frac{\bm{y}^{(t)}-\bm{Y}_i}{h_2}}_2^2 \right) \bm{X}_i^T \left(\bm{x}^{(t)} -\bm{x}^{(t+1)} \right)\\
		&\stackrel{\text{(ii)}}{=} \frac{C_{L,D_1}(h_1) \cdot C_{k,D_2}}{nh_2^{D_2+2}} \sum_{i=1}^n L\left(\frac{1-\bm{X}_i^T \bm{x}^{(t+1)}}{h_1^2} \right)  k'\left(\norm{\frac{\bm{y}^{(t)}-\bm{Y}_i}{h_2}}_2^2 \right)\\
		&\hspace{35mm} \times \left[\norm{\bm{y}^{(t+1)}}_2^2 -2\bm{Y}_i^T\bm{y}^{(t+1)} -\norm{\bm{y}^{(t)}}_2^2 + 2\bm{Y}_i^T\bm{y}^{(t)} \right]\\
		&\quad + \frac{C_{L,D_1}(h_1) \cdot C_{k,D_2}}{nh_1^2h_2^{D_2}} \left(\bm{x}^{(t+1)} \right)^T (\bm{x}^{(t+1)} -\bm{x}^{(t)}) \norm{\sum_{i=1}^n \bm{X}_i\cdot L'\left(\frac{1-\bm{X}_i^T \bm{x}^{(t)}}{h_1^2} \right)  k\left(\norm{\frac{\bm{y}^{(t)}-\bm{Y}_i}{h_2}}_2^2 \right)}_2\\
		&\stackrel{\text{(iii)}}{=} \frac{C_{L,D_1}(h_1) \cdot C_{k,D_2}}{nh_2^{D_2+2}} \sum_{i=1}^n L\left(\frac{1-\bm{X}_i^T \bm{x}^{(t+1)}}{h_1^2} \right)  k'\left(\norm{\frac{\bm{y}^{(t)}-\bm{Y}_i}{h_2}}_2^2 \right)\\
		&\hspace{40mm} \times \left[\norm{\bm{y}^{(t+1)}}_2^2 -2\norm{\bm{y}^{(t+1)}}_2^2 -\norm{\bm{y}^{(t)}}_2^2 + 2\left(\bm{y}^{(t+1)} \right)^T\bm{y}^{(t)} \right]\\
		&\quad + \frac{C_{L,D_1}(h_1) \cdot C_{k,D_2}}{2nh_1^2h_2^{D_2}} \norm{\bm{x}^{(t+1)} -\bm{x}^{(t)}}_2^2 \norm{\sum_{i=1}^n \bm{X}_i\cdot L'\left(\frac{1-\bm{X}_i^T \bm{x}^{(t)}}{h_1^2} \right)  k\left(\norm{\frac{\bm{y}^{(t)}-\bm{Y}_i}{h_2}}_2^2 \right)}_2\\
		&= -\frac{C_{L,D_1}(h_1) \cdot C_{k,D_2}}{nh_2^{D_2+2}} \sum_{i=1}^n L\left(\frac{1-\bm{X}_i^T \bm{x}^{(t+1)}}{h_1^2} \right)  k'\left(\norm{\frac{\bm{y}^{(t)}-\bm{Y}_i}{h_2}}_2^2 \right) \norm{\bm{y}^{(t)} -\bm{y}^{(t+1)}}_2^2\\
		&\quad + \frac{C_{L,D_1}(h_1) \cdot C_{k,D_2}}{nh_1^2h_2^{D_2}} \norm{\bm{x}^{(t+1)} -\bm{x}^{(t)}}_2^2 \norm{\sum_{i=1}^n \bm{X}_i\cdot L'\left(\frac{1-\bm{X}_i^T \bm{x}^{(t)}}{h_1^2} \right)  k\left(\norm{\frac{\bm{y}^{(t)}-\bm{Y}_i}{h_2}}_2^2 \right)}_2\\
		&\geq 0,
	\end{align*}
	where we leverage the inequalities \eqref{convex_diff} in (i), plug in the first-step iterative equation derived from \eqref{MS_iter_comp} of {\bf Version B} in the main paper as:
	$$\bm{x}^{(t+1)}=-\frac{\sum\limits_{i=1}^n \bm{X}_i\cdot L'\left(\frac{1-\bm{X}_i^T\bm{x}^{(t)}}{h_1^2} \right)  k\left(\norm{\frac{\bm{y}^{(t)}-\bm{Y}_i}{h_2}}_2^2 \right)}{\norm{\sum\limits_{i=1}^n \bm{X}_i \cdot L'\left(\frac{1-\bm{X}_i^T\bm{x}^{(t)}}{h_1^2} \right)  k\left(\norm{\frac{\bm{y}^{(t)}-\bm{Y}_i}{h_2}}_2^2 \right)}_2}$$ 
	to obtain (ii), as well as plug in the second-step iterative equation derived from \eqref{MS_iter_comp} of {\bf Version B} in the main paper as:
	$$\bm{y}^{(t+1)} = \frac{\sum\limits_{i=1}^n \bm{Y}_i \cdot L\left(\frac{1-\bm{X}_i^T\bm{x}^{(t+1)}}{h_1^2} \right) k'\left(\norm{\frac{\bm{y}^{(t)}-\bm{Y}_i}{h_2}}_2^2 \right)}{\sum\limits_{i=1}^n L\left(\frac{1-\bm{X}_i^T\bm{x}^{(t+1)}}{h_1^2} \right) k'\left(\norm{\frac{\bm{y}^{(t)}-\bm{Y}_i}{h_2}}_2^2 \right)}$$
	and use the fact that $\left(\bm{x}^{(t+1)}\right)^T \left(\bm{x}^{(t+1)} -\bm{x}^{(t)} \right) = \frac{1}{2}\norm{\bm{x}^{(t+1)} - \bm{x}^{(t)}}_2^2$ in (iii). Finally, we utilize the non-increasing property of $k$ to argue that $-k'(s)\geq 0$ for all $s\in [0,\infty)$ and the final summands are non-negative. The result follows.
	
	\noindent (b) By holding $r$ constant in $K(r,s) = L(r) \cdot k(s)$, it is easy to verify by definition that $k$ is convex given that the entire directional-linear kernel profile $K(r,s)$ is jointly convex. Similarly, we can demonstrate that $L$ is convex. By (a), the sequence $\left\{\hat{f}_{\bm{h}}(\bm{x}^{(t)},\bm{y}^{(t)}) \right\}_{t=0}^{\infty}$ of density estimates defined by {\bf Version B} in \eqref{MS_iter_comp} of the main paper is non-decreasing and thus converges. \\
	Now, we show that the sequence $\left\{\hat{f}_{\bm{h}}(\bm{x}^{(t)},\bm{y}^{(t)})\right\}_{t=0}^{\infty}$ defined by {\bf Version A} is also non-decreasing and thus converges. Notice that by the joint convexity and differentiability of $K(r,s)$,
	\begin{align}
		\label{convex_diff2}
		\begin{split}
			K(r_2,s_2) -K(r_1,s_1) &\geq \left[(r_2,s_2) - (r_1,s_1)\right] \nabla K(r_1,s_1) \\
			&= (r_2-r_1) \cdot \frac{\partial}{\partial r}K(r_1,s_1) + (s_2-s_1) \cdot \frac{\partial K}{\partial s}(r_1,s_1).
		\end{split}
	\end{align}
	Given this inequality and the fact that $\norm{\bm{x}^{(t)}}_2 = \norm{\bm{X}_i}_2=1$ for all $t=0,1,...$ and $i=1,...,n$, we compute that
	\begin{align*}
		&\hat{f}_{\bm{h}}(\bm{x}^{(t+1)}, \bm{y}^{(t+1)}) - \hat{f}_{\bm{h}}(\bm{x}^{(t)}, \bm{y}^{(t)}) \\
		&= \frac{C_{L,D_1}(h_1) \cdot C_{k,D_2}}{nh_2^{D_2}} \sum_{i=1}^n \Bigg[ L\left(\frac{1-\bm{X}_i^T \bm{x}^{(t+1)}}{h_1^2} \right)  k\left(\norm{\frac{\bm{y}^{(t+1)}-\bm{Y}_i}{h_2}}_2^2 \right) \\
		&\hspace{35mm}- L\left(\frac{1-\bm{X}_i^T \bm{x}^{(t)}}{h_1^2} \right)  k\left(\norm{\frac{\bm{y}^{(t)}-\bm{Y}_i}{h_2}}_2^2 \right) \Bigg]\\
		&\stackrel{\text{(i)}}{=} \frac{C_{L,D_1}(h_1) \cdot C_{k,D_2}}{nh_2^{D_2}} \sum_{i=1}^n \left(\frac{1-\bm{X}_i^T \bm{x}^{(t+1)}}{h_1^2} - \frac{1-\bm{X}_i^T\bm{x}^{(t)}}{h_1^2},\, \norm{\frac{\bm{y}^{(t+1)}- \bm{Y}_i}{h_2^2}}_2^2 -\norm{\frac{\bm{y}^{(t)} - \bm{Y}_i}{h_2^2}}_2^2 \right)\\
		&\hspace{30mm} \times \begin{pmatrix}
			L'\left(\frac{1-\bm{X}_i^T\bm{x}^{(t)}}{h_1^2} \right) k\left(\norm{\frac{\bm{y}^{(t)} -\bm{Y}_i}{h_2}}_2^2 \right)\\
			L\left(\frac{1-\bm{X}_i^T\bm{x}^{(t)}}{h_1^2} \right) k'\left(\norm{\frac{\bm{y}^{(t)} -\bm{Y}_i}{h_2}}_2^2 \right)
		\end{pmatrix}\\
		&= \frac{C_{L,D_1}(h_1) \cdot C_{k,D_2}}{nh_1^2h_2^{D_2}} \sum_{i=1}^n \bm{X}_i^T \left(\bm{x}^{(t)} -\bm{x}^{(t+1)} \right) L'\left(\frac{1-\bm{X}_i^T\bm{x}^{(t)}}{h_1^2} \right) k\left(\norm{\frac{\bm{y}^{(t)} -\bm{Y}_i}{h_2}}_2^2 \right)\\
		&\quad + \frac{C_{L,D_1}(h_1) \cdot C_{k,D_2}}{nh_2^{D_2+2}} \sum_{i=1}^n \left[\norm{\bm{y}^{(t+1)}}_2^2 -2\bm{Y}_i^T\bm{y}^{(t+1)} -\norm{\bm{y}^{(t)}}_2^2 + 2\bm{Y}_i^T \bm{y}^{(t)} \right] \\
		&\hspace{35mm} \times L\left(\frac{1-\bm{X}_i^T\bm{x}^{(t)}}{h_1^2} \right) k'\left(\norm{\frac{\bm{y}^{(t)} -\bm{Y}_i}{h_2}}_2^2 \right)\\
		&\stackrel{\text{(ii)}}{=} \frac{C_{L,D_1}(h_1) \cdot C_{k,D_2}}{nh_1^2h_2^{D_2}} \left(\bm{x}^{(t+1)} \right)^T \left(\bm{x}^{(t+1)} -\bm{x}^{(t)} \right) \norm{\sum_{i=1}^n \bm{X}_i\cdot L'\left(\frac{1-\bm{X}_i^T \bm{x}^{(t)}}{h_1^2} \right) k\left(\norm{\frac{\bm{y}^{(t)}-\bm{Y}_i}{h_2}}_2^2 \right)}_2\\
		&\quad + \frac{C_{L,D_1}(h_1) \cdot C_{k,D_2}}{nh_2^{D_2+2}} \sum_{i=1}^n L\left(\frac{1-\bm{X}_i^T \bm{x}^{(t)}}{h_1^2} \right)  k'\left(\norm{\frac{\bm{y}^{(t)}-\bm{Y}_i}{h_2}}_2^2 \right)\\
		&\hspace{30mm} \times \left[\norm{\bm{y}^{(t+1)}}_2^2 -2\norm{\bm{y}^{(t+1)}}_2^2 -\norm{\bm{y}^{(t)}}_2^2 + 2\left(\bm{y}^{(t+1)} \right)^T\bm{y}^{(t)} \right]\\
		&= \frac{C_{L,D_1}(h_1) \cdot C_{k,D_2}}{2nh_1^2h_2^{D_2}} \cdot \norm{\bm{x}^{(t+1)} -\bm{x}^{(t)}}_2^2 \norm{\sum_{i=1}^n \bm{X}_i\cdot L'\left(\frac{1-\bm{X}_i^T \bm{x}^{(t)}}{h_1^2} \right)  k\left(\norm{\frac{\bm{y}^{(t)}-\bm{Y}_i}{h_2}}_2^2 \right)}_2\\
		&\quad - \frac{C_{L,D_1}(h_1) \cdot C_{k,D_2}}{nh_2^{D_2+2}} \cdot \norm{\bm{y}^{(t+1)} -\bm{y}^{(t)}}_2^2 \sum_{i=1}^n L\left(\frac{1-\bm{X}_i^T \bm{x}^{(t)}}{h_1^2} \right)  k'\left(\norm{\frac{\bm{y}^{(t)}-\bm{Y}_i}{h_2}}_2^2 \right) \\
		&\geq 0,
	\end{align*}
	where we apply the inequality \eqref{convex_diff2} in (i), plug in the updating equation derived from \eqref{MS_iter_sim} of {\bf Version A} as:
	\begin{align*}
		\bm{x}^{(t+1)} &=-\frac{\sum\limits_{i=1}^n \bm{X}_i\cdot L'\left(\frac{1-\bm{X}_i^T\bm{x}^{(t)}}{h_1^2} \right)  k\left(\norm{\frac{\bm{y}^{(t)}-\bm{Y}_i}{h_2}}_2^2 \right)}{\norm{\sum\limits_{i=1}^n \bm{X}_i \cdot L'\left(\frac{1-\bm{X}_i^T\bm{x}^{(t)}}{h_1^2} \right)  k\left(\norm{\frac{\bm{y}^{(t)}-\bm{Y}_i}{h_2}}_2^2 \right)}_2},\\
		\bm{y}^{(t+1)} &= \frac{\sum\limits_{i=1}^n \bm{Y}_i \cdot L\left(\frac{1-\bm{X}_i^T\bm{x}^{(t)}}{h_1^2} \right) k'\left(\norm{\frac{\bm{y}^{(t)}-\bm{Y}_i}{h_2}}_2^2 \right)}{\sum\limits_{i=1}^n L\left(\frac{1-\bm{X}_i^T\bm{x}^{(t)}}{h_1^2} \right) k'\left(\norm{\frac{\bm{y}^{(t)}-\bm{Y}_i}{h_2}}_2^2 \right)}
	\end{align*}
	in (ii), and leverage the non-increasing property of $k$ to argue the last inequality. The results thus follow.
	
	(c) Our previous calculations in (a) and (b) suggest that
	\begin{align}
		\label{den_diff_seq}
		\begin{split}
			&\hat{f}_{\bm{h}}(\bm{x}^{(t+1)}, \bm{y}^{(t+1)}) - \hat{f}_{\bm{h}}(\bm{x}^{(t)}, \bm{y}^{(t)}) \\
			&=\frac{C_{L,D_1}(h_1) \cdot C_{k,D_2}}{2nh_1^2h_2^{D_2}} \norm{\bm{x}^{(t+1)} -\bm{x}^{(t)}}_2^2 \norm{\sum\limits_{i=1}^n \bm{X}_i\cdot L'\left(\frac{1-\bm{X}_i^T \bm{x}^{(t)}}{h_1^2} \right) k\left(\norm{\frac{\bm{y}^{(t)}-\bm{Y}_i}{h_2}}_2^2 \right)}_2 \\
			&\quad - \begin{cases}
				\frac{C_{L,D_1}(h_1) \cdot C_{k,D_2}}{nh_2^{D_2+2}} \norm{\bm{y}^{(t+1)} -\bm{y}^{(t)}}_2^2 \sum\limits_{i=1}^n L\left(\frac{1-\bm{X}_i^T \bm{x}^{(t)}}{h_1^2} \right)  k'\left(\norm{\frac{\bm{y}^{(t)}-\bm{Y}_i}{h_2}}_2^2 \right) \quad \text{ for Version A}, \\
				\frac{C_{L,D_1}(h_1) \cdot C_{k,D_2}}{nh_2^{D_2+2}} \norm{\bm{y}^{(t)} -\bm{y}^{(t+1)}}_2^2 \sum\limits_{i=1}^n L\left(\frac{1-\bm{X}_i^T \bm{x}^{(t+1)}}{h_1^2} \right)  k'\left(\norm{\frac{\bm{y}^{(t)}-\bm{Y}_i}{h_2}}_2^2 \right)  \quad \text{ for Version B}.
			\end{cases}
		\end{split}
	\end{align}
	Under the conditions that $k$ and $L$ are strictly decreasing on $[0,\infty)$ and the data-generating density $f$ has a compact support on $\mathcal{S}_1\times \mathcal{S}_2$, all the factors in front of $\norm{\bm{x}^{(t+1)} - \bm{x}^{(t)}}_2^2$ and $\norm{\bm{y}^{(t+1)} - \bm{y}^{(t)}}_2^2$ in \eqref{den_diff_seq} are strictly positive. Because we have shown that 
	$$\hat{f}_{\bm{h}}(\bm{x}^{(t+1)}, \bm{y}^{(t+1)}) - \hat{f}_{\bm{h}}(\bm{x}^{(t)}, \bm{y}^{(t)}) \to 0$$ 
	as $t\to\infty$ in (a) and (b), it implies that $\norm{(\bm{x}^{(t+1)},\bm{y}^{(t+1)}) - (\bm{x}^{(t)},\bm{y}^{(t)})}_2 = \norm{\bm{z}^{(t+1)} - \bm{z}^{(t)}}_2 \to 0$ as $t\to\infty$.
	
	(d) The proof of (d) follows from (c) together with the arguments of Theorem 2 in \cite{MS2007_pf}, Theorem 1 in \cite{MS2015_Gaussian}, and Theorem 11 in \cite{DirMS2020}. We only outline the key steps. First, given the isolation of local modes of $\hat{f}_{\bm{h}}$, one can derive that 
	$$\grad \hat{f}_{\bm{h}}(\bm{z}) > 0 \quad \text{ for any } \bm{z} \in B\left(\hat{\bm{m}}, \rho \right)\setminus \left\{\hat{\bm{m}} \right\},$$
	where $\hat{\bm{m}} \in \hat{\mathcal{M}}$ is a local mode of $\hat{f}_{\bm{h}}$ and $B\left(\hat{\bm{m}}, \rho \right)$ is the $\rho$-neighborhood of $\hat{\bm{m}}$ in $\mathcal{S}_1\times \mathcal{S}_2$. Then, as $\norm{\bm{z}^{(t+1)} - \bm{z}^{(t)}}_2 \to 0$ when $t\to\infty$, one can argue that
	$$\grad \hat{f}_{\bm{h}}(\bm{z}^{(t)}) \to 0 \quad \text{ and } \quad \bm{z}^{(t)} \in B\left(\hat{\bm{m}}, \rho \right) \quad \text{ as } t \to \infty.$$
	Lastly, as $\hat{\bm{m}}$ is the only point satisfying $\grad \hat{f}_{\bm{h}}(\hat{\bm{m}})=0$ within $B\left(\hat{\bm{m}}, \rho \right)$, it follows that $\bm{z}^{(t)} \to \hat{\bm{m}}$ as $t\to\infty$.
\end{proof}
\vspace{5mm}

We restate the full version of Proposition~\ref{Thm:trans_grad_conv} here.
\begin{proposition}
	\label{Thm:trans_grad_conv_full}
	Assume conditions (A1-2). For any fixed $\bm{z}\in \mathcal{S}_1\times \mathcal{S}_2$, we have that 
	\begin{align*}
			\mathcal{P}_{\bm{z}} \bm{H}^{-1} \tilde{D}(\bm{z})^{-1} \nabla \hat{f}_{\bm{h}}(\bm{z}) &= \tilde{D}(\bm{z})^{-1} \bm{H}^{-1} \grad \hat{f}_{\bm{h}}(\bm{z})
		\end{align*}
	under any given $h_1,h_2,n$ and 
	\begin{align*}
			\left[\bm{H}\tilde{D}(\bm{z})\right]^{-1} \grad \hat{f}_{\bm{h}}(\bm{z}) -\tilde{F}(\bm{z})^{-1}\grad f(\bm{z}) &= O(h^2) + O_P\left(\sqrt{\frac{1}{nh^{D_1+D_2+2}}} \right)
		\end{align*}
	as $h_1,h_2\lesssim h \to 0$ and $nh^{D_1+D_2+2} \to \infty$, where $\tilde{F}(\bm{z}) = \begin{pmatrix}
				\tilde{F}_1 \cdot \bm{I}_{D_1+\mathbbm{1}_{\{\mathcal{S}_1=\Omega_{D_1}\}}}& \bm{0}\\
				\bm{0} & \tilde{F}_2 \cdot \bm{I}_{D_2+\mathbbm{1}_{\{\mathcal{S}_2=\Omega_{D_2}\}}}
			\end{pmatrix}$ with 
	\begin{align*}
		&\tilde{F}_1=
			\begin{cases}
					-C_{k,D_1} \int_{\mathbb{R}^{D_1}} k'\left(\norm{\bm{x}}_2^2/2 \right) d\bm{x} \cdot f(\bm{z})  & \text{ if } \mathcal{S}_1=\mathbb{R}^{D_1},\\
					-\frac{\int_0^{\infty} L(r) r^{\frac{D_1}{2}-1} dr}{\int_0^{\infty} L'(r) r^{\frac{D_1}{2}-1} dr} \cdot f(\bm{z}) & \text{ if } \mathcal{S}_1=\Omega_{D_1},
				\end{cases} \\ 
			&\text{ and } \quad \tilde{F}_2=
			\begin{cases}
					-C_{k,D_2} \int_{\mathbb{R}^{D_2}} k'\left(\norm{\bm{y}}_2^2/2 \right) d\bm{y} \cdot f(\bm{z}) & \text{ if } \mathcal{S}_2=\mathbb{R}^{D_2},\\
					-\frac{\int_0^{\infty} L(r) r^{\frac{D_2}{2}-1} dr}{\int_0^{\infty} L'(r) r^{\frac{D_2}{2}-1} dr} \cdot f(\bm{z}) & \text{ if } \mathcal{S}_2=\Omega_{D_2}.
				\end{cases}
		\end{align*}
\end{proposition}

\begin{proof}[Proof of Proposition~\ref{Thm:trans_grad_conv_full}]
	Based on the definitions of the projection matrix $\mathcal{P}_{\bm{z}}$ in \eqref{Riem_grad}, bandwidth matrix $\bm{H}$ in \eqref{KDE_Exp}, and $\tilde{D}(\bm{z})$ in \eqref{trans_tot_grad_mat} in the main paper, we know that $\bm{H},\tilde{D}(\bm{z})$ are diagonal and $\mathcal{P}_{\bm{z}}$ is block diagonal. It is intuitive that $\mathcal{P}_{\bm{z}}, \bm{H}^{-1}, \tilde{D}(\bm{z})^{-1}$ are mutually commutative. Hence, for any given $h_1,h_2,n$,
	$\mathcal{P}_{\bm{z}} \bm{H}^{-1} \tilde{D}(\bm{z})^{-1} \nabla \hat{f}_{\bm{h}}(\bm{z}) = \tilde{D}(\bm{z})^{-1} \bm{H}^{-1} \grad \hat{f}_{\bm{h}}(\bm{z}).$
	
	As for the pointwise convergence of $\tilde{D}(\bm{z})^{-1} \bm{H}^{-1} \grad \hat{f}_{\bm{h}}(\bm{z})$, we know from Lemma~\ref{pw_rate_KDE} that for any fixed $\bm{z}\in \mathcal{S}_1\times \mathcal{S}_2$,
	\begin{equation}
		\label{pw_grad_conv}
		\grad \hat{f}_{\bm{h}}(\bm{z}) - \grad f(\bm{z}) = O(h^2) +O_P\left(\sqrt{\frac{1}{nh^{D_1+D_2+2}}} \right)
	\end{equation}
	as $h_1,h_2 \lesssim h \to 0$ and $nh^{D_1+D_2+2} \to \infty$. It remains to derive the pointwise convergence rate of $\tilde{D}(\bm{z})^{-1} \bm{H}^{-1}= \left[\bm{H}\tilde{D}(\bm{z})\right]^{-1}$. Recall that 
	$$\bm{H}\tilde{D}(\bm{z})= \Diag\left(h_1^2G_{\bm{x}} \bm{I}_{D_1+\mathbbm{1}_{\{\mathcal{S}_1= \Omega_{D_1}\}}}, h_2^2G_{\bm{y}} \bm{I}_{D_2+\mathbbm{1}_{\{\mathcal{S}_2= \Omega_{D_2}\}}} \right),$$
	where 
	\begin{align*}
		h_1^2G_{\bm{x}} &= -\frac{2 C(\bm{H})}{n} \sum_{i=1}^n k_1'\left(\norm{\frac{\bm{x}-\bm{X}_i}{h_1}}_2^2\right)K_2\left(\frac{\bm{y}-\bm{Y}_i}{h_2}\right) \\
		&= -\frac{2 \prod_{j=1}^2 C_{k_j,D_j}(h_j)}{n} \sum_{i=1}^n k_1'\left(\norm{\frac{\bm{x}-\bm{X}_i}{h_1}}_2^2\right)k_2\left(\norm{\frac{\bm{y}-\bm{Y}_i}{h_2}}_2^2\right), \\ 
		h_2^2G_{\bm{y}} &= -\frac{2 C(\bm{H})}{n} \sum_{i=1}^n K_1\left(\frac{\bm{x}-\bm{X}_i}{h_1}\right)k_2'\left(\norm{\frac{\bm{y}-\bm{Y}_i}{h_2}}_2^2\right) \\
		&= -\frac{2 \prod_{j=1}^2 C_{k_j,D_j}(h_j)}{n} \sum_{i=1}^n k_1\left(\norm{\frac{\bm{x}-\bm{X}_i}{h_1}}_2^2\right)k_2'\left(\norm{\frac{\bm{y}-\bm{Y}_i}{h_2}}_2^2\right).
	\end{align*}
	It suggests that $h_1^2G_{\bm{x}}, h_2^2G_{\bm{y}}$ can be viewed as ``KDEs'' with kernel profiles $-k_1'(\cdot) k_2(\cdot)$ and $-k_1(\cdot) k_2'(\cdot)$, respectively. With regards to the pointwise convergence rate of $h_1^2G_{\bm{x}}$, we consider the following two cases of the kernel profile $k_1$.
	
	$\bullet$ {\bf Case 1: $k_1$ is a linear kernel profile}. In this case, $\mathcal{S}_1=\mathbb{R}^{D_1}$ and 
	$C_{k_1,D_1}(h_1)\cdot k_1\left(\norm{\bm{u}}_2^2 \right)=\frac{C_{k,D_1}}{h_1^{D_1}} \cdot k\left(\frac{\norm{\bm{u}}_2^2}{2} \right).$
	Then,
	$$h_1^2G_{\bm{x}} = -\frac{C_{k,D_1} \cdot C_{k_2,D_2}(h_2)}{nh_1^{D_1}} \sum_{i=1}^n k'\left(\frac{1}{2}\norm{\frac{\bm{x}-\bm{X}_i}{h_1}}_2^2\right)k_2\left(\norm{\frac{\bm{y}-\bm{Y}_i}{h_2}}_2^2\right).$$
	One can follow the arguments in Theorems 1 and 2 of \cite{Asymp_deri_KDE2011} or Lemma 3.2 of \cite{DirSCMS2021} to show that
	\begin{equation}
		\label{G_x_conv_rate1}
		h_1^2 G_{\bm{x}} - \left(-C_{k,D_1} \int_{\mathbb{R}^{D_1}} k'\left(\norm{\bm{x}}_2^2/2 \right) d\bm{x} \cdot f(\bm{z}) \right) = O(h^2) + O_P\left(\sqrt{\frac{1}{nh^{D_1+D_2}}} \right)
	\end{equation}
	as $h_1,h_2\lesssim h \to 0$ and $nh^{D_1+D_2} \to \infty$.
	
	$\bullet$ {\bf Case 2: $k_1$ is a directional kernel profile}. In this case, $\mathcal{S}_1=\Omega_{D_1}$ and $C_{k_1,D_1}(h_1)\cdot k_1\left(\norm{\bm{u}}_2^2 \right)=C_{L,D_1}(h_1) \cdot L\left(\frac{\norm{\bm{u}}_2^2}{2} \right).$
	Then, using the fact that $\norm{\bm{x}}_2=\norm{\bm{X}_i}_2=1$ for any $i=1,...,n$, 
	$$h_1^2G_{\bm{x}} = -\frac{C_{L,D_1}(h_1) \cdot C_{k_2,D_2}(h_2)}{n} \sum_{i=1}^n L'\left(\frac{1-\bm{x}^{\top}\bm{X}_i}{h_1^2}\right)k_2\left(\norm{\frac{\bm{y}-\bm{Y}_i}{h_2}}_2^2\right).$$
	One can follow the arguments in Proposition 1 of \cite{Dir_Linear2013} and Theorem 2 of \cite{DirMS2020} to show that
	\begin{equation}
		\label{G_x_conv_rate2}
		h_1^2 G_{\bm{x}} - \left(- \frac{\int_0^{\infty} L(r) r^{\frac{D_1}{2}-1} dr}{\int_0^{\infty} L'(r) r^{\frac{D_1}{2}-1} dr} \cdot f(\bm{z})\right) = O(h^2) + O_P\left(\sqrt{\frac{1}{nh^{D_1+D_2}}} \right)
	\end{equation}
	as $h_1,h_2\lesssim h \to 0$ and $nh^{D_1+D_2} \to \infty$.
	
	Combining \eqref{pw_grad_conv}, \eqref{G_x_conv_rate1}, and \eqref{G_x_conv_rate2}, we obtain by Taylor's theorem that
	$\frac{\grad \hat{f}_{\bm{h}}(\bm{z})}{h_1^2G_{\bm{x}}} - \frac{\grad f(\bm{z})}{\tilde{F}_1} = O(h^2) + O_P\left(\sqrt{\frac{1}{nh^{D_1+D_2+2}}} \right)$
	as $h_1,h_2\lesssim h \to 0$ and $nh^{D_1+D_2+2} \to \infty$. A similar argument demonstrates that
	$\frac{\grad \hat{f}_{\bm{h}}(\bm{z})}{h_2^2G_{\bm{y}}} - \frac{\grad f(\bm{z})}{\tilde{F}_2} = O(h^2) + O_P\left(\sqrt{\frac{1}{nh^{D_1+D_2+2}}} \right)$
	as $h_1,h_2\lesssim h \to 0$ and $nh^{D_1+D_2+2} \to \infty$. The results thus follow.
\end{proof}

With an additional condition (A3), one can extend Proposition~\ref{Thm:trans_grad_conv} to an uniform rate of convergence. This stronger result will not provide more insights into the inconsistency of the naive SCMS algorithm, so we omit its presentation.

\section{Proof of Proposition~\ref{prop:bw_LSCV}}
\label{Sec:auxi_pf}

\begin{customprop}{A.1}[Explicit LSCV loss under von Mises and/or Gaussian kernels]
Let $\hat{f}_{\bm{h}}(\bm{x},\bm{y})$ in \eqref{KDE_prod} be defined on the product space $\mathcal{S}_1\times \mathcal{S}_2$ with $\bm{h}=(h_1,h_2)$. If the von Mises kernel profile $L(r)=e^{-r}$ is used for directional components and Gaussian kernel profile $k(s) = e^{-s/2}$ is applied to Euclidean components, then we have the following results.

(a) When $\mathcal{S}_1\times \mathcal{S}_2 = \mathbb{R}^{D_1}\times \mathbb{R}^{D_2}$,
\begin{align*}
	\mathrm{LSCV}(\bm{h}) &= \frac{1}{2^{D_1+D_2} \pi^{\frac{D_1+D_2}{2}} n h^{D_1}h^{D_2}}\left[1+ \frac{2}{n}\sum\limits_{i=1}^{n-1} \sum\limits_{j=1,j>i}^n \exp\left(-\frac{\norm{\bm{X}_i-\bm{X}_j}_2^2}{4h_1^2} - \frac{\norm{\bm{Y}_i-\bm{Y}_j}_2^2}{4h_2^2}\right) \right] \\
	&\quad - \frac{4}{(2\pi)^{\frac{D_1+D_2}{2}} n(n-1)h_1^{D_1} h_2^{D_2}}\sum\limits_{i=1}^{n-1} \sum\limits_{j=1,j>i}^n \exp\left(-\frac{\norm{\bm{X}_i-\bm{X}_j}_2^2}{2h_1^2} - \frac{\norm{\bm{Y}_i-\bm{Y}_j}_2^2}{2h_2^2}\right).
\end{align*}

(b) When $\mathcal{S}_1\times \mathcal{S}_2 = \Omega_{D_1}\times \Omega_{D_2}$,
\begin{align*}
	\mathrm{LSCV}(\bm{h}) &= \frac{C_{D_1}\left(\frac{1}{h_1^2}\right)^2\cdot C_{D_2}\left(\frac{1}{h_2^2}\right)^2}{n\cdot  C_{D_1}\left(\frac{2}{h_1^2}\right)\cdot C_{D_2}\left(\frac{2}{h_2^2}\right)} + \frac{2}{n^2}\sum\limits_{i=1}^{n-1} \sum\limits_{j=1,j>i}^n \frac{C_{D_1}\left(\frac{1}{h_1^2}\right)^2\cdot C_{D_2}\left(\frac{1}{h_2^2}\right)^2}{C_{D_1}\left(\frac{\norm{\bm{X}_i + \bm{X}_j}_2}{h_1^2}\right)\cdot C_{D_2}\left(\frac{\norm{\bm{Y}_i + \bm{Y}_j}_2}{h_2^2}\right)} \\
	&\quad - \frac{4 C_{D_1}\left(\frac{1}{h_1^2}\right)\cdot C_{D_2}\left(\frac{1}{h_2^2}\right)}{n(n-1)} \sum\limits_{i=1}^{n-1} \sum\limits_{j=1,j>i}^n \exp\left(\frac{\bm{X}_i^T\bm{X}_j}{h_1^2} + \frac{\bm{Y}_i^T\bm{Y}_i}{h_2^2}\right),
\end{align*}
where $C_q(\kappa)>0$ is the normalizing constant of the vMF distribution.

(c) When $\mathcal{S}_1\times \mathcal{S}_2 = \Omega_{D_1}\times \mathbb{R}^{D_2}$,
\begin{align*}
	\mathrm{LSCV}(\bm{h}) &= \frac{C_{D_1}\left(\frac{1}{h_1^2}\right)^2}{2^{D_2} \pi^{\frac{D_2}{2}} n \cdot C_{D_1}\left(\frac{2}{h_1^2}\right)\cdot h^{D_2}} + \frac{C_{D_1}\left(\frac{1}{h_1^2}\right)^2}{2^{D_2-1}\pi^{\frac{D_2}{2}}n^2} \sum\limits_{i=1}^{n-1} \sum\limits_{j=1,j>i}^n \frac{\exp\left(-\frac{\norm{\bm{Y}_i-\bm{Y}_j}_2^2}{4h_2^2}\right)}{C_{D_1}\left(\frac{\norm{\bm{X}_i+\bm{X}_j}_2}{h_1^2}\right)}\\
	&\quad - \frac{4C_{D_1}\left(\frac{1}{h_1^2}\right)}{(2\pi)^{\frac{D_2}{2}} n(n-1) h_2^{D_2}}\sum\limits_{i=1}^{n-1} \sum\limits_{j=1,j>i}^n \exp\left(\frac{\bm{X}_i^T\bm{X}_j}{h_1^2} - \frac{\norm{\bm{Y}_i-\bm{Y}_j}_2^2}{2h_2^2}\right).
\end{align*}
\end{customprop}

\begin{proof}[Proof of Proposition~\ref{prop:bw_LSCV}]
Recall from \eqref{KDE_prod} that 
$$\hat{f}_{\bm{h}}(\bm{x},\bm{y}) = \frac{1}{n} \sum_{i=1}^n K_1\left(\frac{\bm{x}-\bm{X}_i}{h_1} \right) K_2\left(\frac{\bm{y}-\bm{Y}_i}{h_2} \right)$$
for two general kernels $K_1,K_2$. Then, we know that
\begin{align*}
&\mathrm{LSCV}(\bm{h}) \\
&= \int_{\mathcal{S}_1\times \mathcal{S}_2} \hat{f}_{\bm{h}}(\bm{x},\bm{y})^2 \omega(d\bm{x},d\bm{y}) - \frac{2}{n} \sum_{i=1}^n \hat{f}_{\bm{h},-i}(\bm{X}_i,\bm{Y}_i) \\
&= \frac{1}{n^2} \int_{\mathcal{S}_1\times \mathcal{S}_2}\Bigg[\sum_{i=1}^n K_1\left(\frac{\bm{x}-\bm{X}_i}{h_1} \right)^2 K_2\left(\frac{\bm{y}-\bm{Y}_i}{h_2} \right)^2 \\
&\quad \quad\quad + 2\sum_{i=1}^n \sum_{j>i} K_1\left(\frac{\bm{x}-\bm{X}_i}{h_1} \right) K_2\left(\frac{\bm{y}-\bm{Y}_i}{h_2} \right) K_1\left(\frac{\bm{x}-\bm{X}_j}{h_1} \right) K_2\left(\frac{\bm{y}-\bm{Y}_j}{h_2} \right)\Bigg] \omega(d\bm{x},d\bm{y}) \\
&\quad - \frac{4}{n(n-1)} \sum_{i=1}^n \sum_{j>i} K_1\left(\frac{\bm{X}_i-\bm{X}_j}{h_1} \right) K_2\left(\frac{\bm{Y}_i-\bm{Y}_j}{h_2} \right).
\end{align*}
Now, we special $K_1,K_2$ to the von Mises and/or Gaussian kernels and consider three cases as follows.
	
(a) When $\mathcal{S}_1\times \mathcal{S}_2 = \mathbb{R}^{D_1}\times \mathbb{R}^{D_2}$, we have that
\begin{align*}
&\mathrm{LSCV}(\bm{h}) \\
&= \frac{1}{(2\pi)^{D_1+D_2}n h_1^{2D_1}h_2^{2D_2}}\int_{\mathbb{R}^{D_1}\times \mathbb{R}^{D_2}} \exp\left(-\frac{\norm{\bm{x}-\bm{X}_i}_2^2}{h_1^2} - \frac{\norm{\bm{y}-\bm{Y}_i}_2^2}{h_2^2}\right) d\bm{x} d\bm{y}\\
&\quad +\frac{2}{(2\pi)^{D_1+D_2}n^2 h_1^{2D_1}h_2^{2D_2}} \sum_{i=1}^n \sum_{j>i} \\
&\quad \quad \quad \int_{\mathbb{R}^{D_1}\times \mathbb{R}^{D_2}}\exp\left(-\frac{\norm{\bm{x}-\bm{X}_i}_2^2}{2h_1^2} - \frac{\norm{\bm{x}-\bm{X}_j}_2^2}{2h_1^2}- \frac{\norm{\bm{y}-\bm{Y}_i}_2^2}{2h_2^2} - \frac{\norm{\bm{y}-\bm{Y}_j}_2^2}{2h_2^2}\right) d\bm{x} d\bm{y}\\
&- \frac{4}{(2\pi)^{\frac{D_1+D_2}{2}} n(n-1)h_1^{D_1} h_2^{D_2}}\sum\limits_{i=1}^n\sum\limits_{j>i} \exp\left(-\frac{\norm{\bm{X}_i-\bm{X}_j}_2^2}{2h_1^2} - \frac{\norm{\bm{Y}_i-\bm{Y}_j}_2^2}{2h_2^2}\right)\\
&= \frac{1}{2^{D_1+D_2} \pi^{\frac{D_1+D_2}{2}} n h^{D_1}h^{D_2}}\left[1+ \frac{2}{n}\sum_{i=1}^{n-1} \sum_{j=1,j>i}^n \exp\left(-\frac{\norm{\bm{X}_i-\bm{X}_j}_2^2}{4h_1^2} - \frac{\norm{\bm{Y}_i-\bm{Y}_j}_2^2}{4h_2^2}\right) \right] \\
&\quad - \frac{4}{(2\pi)^{\frac{D_1+D_2}{2}} n(n-1)h_1^{D_1} h_2^{D_2}} \sum_{i=1}^{n-1} \sum_{j=1,j>i}^n \exp\left(-\frac{\norm{\bm{X}_i-\bm{X}_j}_2^2}{2h_1^2} - \frac{\norm{\bm{Y}_i-\bm{Y}_j}_2^2}{2h_2^2}\right).
\end{align*}

(b) When $\mathcal{S}_1\times \mathcal{S}_2 = \Omega_{D_1}\times \Omega_{D_2}$, we have that
\begin{align*}
&\mathrm{LSCV}(\bm{h}) \\
&= \frac{C_{D_1}\left(\frac{1}{h_1^2}\right)^2\cdot C_{D_2}\left(\frac{1}{h_2^2}\right)^2}{n} \int_{\Omega_{D_1}\times \Omega_{D_2}} \exp\left(\frac{2\bm{x}^T\bm{X}_i}{h_1^2} + \frac{2\bm{y}^T\bm{Y}_i}{h_2^2}\right)\omega(d\bm{x}) \omega(d\bm{y}) \\
&\quad + \frac{2C_{D_1}\left(\frac{1}{h_1^2}\right)^2\cdot C_{D_2}\left(\frac{1}{h_2^2}\right)^2}{n^2} \sum\limits_{i=1}^n\sum\limits_{j>i} \int_{\Omega_{D_1}\times \Omega_{D_2}} \exp\left[\frac{\bm{x}^T(\bm{X}_i+\bm{X}_j)}{h_1^2} + \frac{\bm{y}^T(\bm{Y}_i+\bm{Y}_j)}{h_2^2}\right]\omega(d\bm{x}) \omega(d\bm{y})\\
&\quad - \frac{4 C_{D_1}\left(\frac{1}{h_1^2}\right)\cdot C_{D_2}\left(\frac{1}{h_2^2}\right)}{n(n-1)} \sum_{i=1}^n \sum_{j>i} \exp\left(\frac{\bm{X}_i^T\bm{X}_j}{h_1^2} + \frac{\bm{Y}_i^T\bm{Y}_i}{h_2^2}\right) \\
&= \frac{C_{D_1}\left(\frac{1}{h_1^2}\right)^2\cdot C_{D_2}\left(\frac{1}{h_2^2}\right)^2}{n\cdot  C_{D_1}\left(\frac{2}{h_1^2}\right)\cdot C_{D_2}\left(\frac{2}{h_2^2}\right)} + \frac{2}{n^2}\sum_{i=1}^{n-1} \sum_{j=1,j>i}^n \frac{C_{D_1}\left(\frac{1}{h_1^2}\right)^2\cdot C_{D_2}\left(\frac{1}{h_2^2}\right)^2}{C_{D_1}\left(\frac{\norm{\bm{X}_i + \bm{X}_j}_2}{h_1^2}\right)\cdot C_{D_2}\left(\frac{\norm{\bm{Y}_i + \bm{Y}_j}_2}{h_2^2}\right)} \\
&\quad - \frac{4 C_{D_1}\left(\frac{1}{h_1^2}\right)\cdot C_{D_2}\left(\frac{1}{h_2^2}\right)}{n(n-1)} \sum_{i=1}^{n-1} \sum_{j=1,j>i}^n \exp\left(\frac{\bm{X}_i^T\bm{X}_j}{h_1^2} + \frac{\bm{Y}_i^T\bm{Y}_i}{h_2^2}\right),
\end{align*}
where the last equality uses the fact that 
\begin{align*}
\int_{\Omega_{D_1}} \exp\left[\frac{\bm{x}^T(\bm{X}_i+\bm{X}_j)}{h_1^2} \right] \omega(d\bm{x}) &= \int_{\Omega_{D_1}} \exp\left[\frac{\bm{x}^T(\bm{X}_i+\bm{X}_j)}{\norm{\bm{X}_i+\bm{X}_j}_2} \cdot \frac{\norm{\bm{X}_i+\bm{X}_j}_2}{h_1^2} \right] \omega(d\bm{x}) \\
&= C_{D_1}\left(\frac{\norm{\bm{X}_i+\bm{X}_j}_2}{h_1^2}\right).
\end{align*}

(c) When $\mathcal{S}_1\times \mathcal{S}_2 = \Omega_{D_1}\times \mathbb{R}^{D_2}$, we can follow similar arguments above to derive that
\begin{align*}
&\mathrm{LSCV}(\bm{h}) \\
&= \frac{C_{D_1}\left(\frac{1}{h_1^2}\right)^2}{(2\pi)^{D_2}n h_2^{2D_2}}\int_{\Omega_{D_1}\times \mathbb{R}^{D_2}} \exp\left(\frac{2\bm{x}^T\bm{X}_i}{h_1^2} -\frac{\norm{\bm{y}-\bm{Y}_i}_2^2}{h_2^2}\right) \omega(d\bm{x}) d\bm{y}\\
&\quad + \frac{2C_{D_1}\left(\frac{1}{h_1^2}\right)^2}{(2\pi)^{D_2}n^2 h_2^{2D_2}} \sum_{i=1}^n \sum_{j>i} \int_{\mathbb{R}^{D_1}\times \mathbb{R}^{D_2}}\exp\left(\frac{\bm{x}^T(\bm{X}_i+\bm{X}_i)}{h_1^2} - \frac{\norm{\bm{y}-\bm{Y}_i}_2^2}{2h_2^2} - \frac{\norm{\bm{y}-\bm{Y}_j}_2^2}{2h_2^2}\right) \omega(d\bm{x}) d\bm{y}\\
&\quad - \frac{4C_{D_1}\left(\frac{1}{h_1^2}\right)^2}{(2\pi)^{\frac{D_2}{2}} n(n-1) h_2^{D_2}}\sum_{i=1}^n\sum_{j>i} \exp\left(\frac{\bm{X}_i^T\bm{X}_j}{h_1^2} - \frac{\norm{\bm{Y}_i-\bm{Y}_j}_2^2}{2h_2^2}\right)\\
&= \frac{C_{D_1}\left(\frac{1}{h_1^2}\right)^2}{2^{D_2} \pi^{\frac{D_2}{2}} n \cdot C_{D_1}\left(\frac{2}{h_1^2}\right)\cdot h^{D_2}} + \frac{C_{D_1}\left(\frac{1}{h_1^2}\right)^2}{2^{D_2-1}\pi^{\frac{D_2}{2}}n^2} \sum_{i=1}^{n-1} \sum_{j=1,j>i}^n \frac{\exp\left(-\frac{\norm{\bm{Y}_i-\bm{Y}_j}_2^2}{4h_2^2}\right)}{C_{D_1}\left(\frac{\norm{\bm{X}_i+\bm{X}_j}_2}{h_1^2}\right)}\\
&\quad - \frac{4C_{D_1}\left(\frac{1}{h_1^2}\right)}{(2\pi)^{\frac{D_2}{2}} n(n-1) h_2^{D_2}}\sum_{i=1}^{n-1} \sum_{j=1,j>i}^n \exp\left(\frac{\bm{X}_i^T\bm{X}_j}{h_1^2} - \frac{\norm{\bm{Y}_i-\bm{Y}_j}_2^2}{2h_2^2}\right).
\end{align*}
The results thus follow.
\end{proof}

\begin{comment}

\bibliography{Bib_DLSCMS}

\end{document}